\PassOptionsToPackage{prologue,usenames,dvipsnames}{xcolor}
\documentclass[sigconf]{acmart}
\acmSubmissionID{726}
\usepackage{tikz}
\usetikzlibrary{spy,calc}
\usepackage{booktabs} 

\usepackage{lipsum} 
\usepackage{comment}
\usepackage{array}
\usepackage{multicol}
\usepackage{amsmath}
\usepackage{amsfonts}
\usepackage[ruled]{algorithm2e} 

\SetAlFnt{\small}
\SetAlCapFnt{\small}
\SetAlCapNameFnt{\small}
\SetAlCapHSkip{0pt}
\newif\ifblackandwhitecycle
\gdef\patternnumber{0}

\pgfkeys{/tikz/.cd,
    zoombox paths/.style={
        draw=orange,
        very thick
    },
    black and white/.is choice,
    black and white/.default=static,
    black and white/static/.style={
        draw=white,
        zoombox paths/.append style={
            draw=white,
            postaction={
                draw=black,
                loosely dashed
            }
        }
    },
    black and white/static/.code={
        \gdef\patternnumber{1}
    },
    black and white/cycle/.code={
        \blackandwhitecycletrue
        \gdef\patternnumber{1}
    },
    black and white pattern/.is choice,
    black and white pattern/0/.style={},
    black and white pattern/1/.style={
            draw=white,
            postaction={
                draw=black,
                dash pattern=on 2pt off 2pt
            }
    },
    black and white pattern/2/.style={
            draw=white,
            postaction={
                draw=black,
                dash pattern=on 4pt off 4pt
            }
    },
    black and white pattern/3/.style={
            draw=white,
            postaction={
                draw=black,
                dash pattern=on 4pt off 4pt on 1pt off 4pt
            }
    },
    black and white pattern/4/.style={
            draw=white,
            postaction={
                draw=black,
                dash pattern=on 4pt off 2pt on 2 pt off 2pt on 2 pt off 2pt
            }
    },
    zoomboxarray inner gap/.initial=5pt,
    zoomboxarray columns/.initial=2,
    zoomboxarray rows/.initial=1,
    zoomboxarray heightmultiplier/.initial=0.5,
    subfigurename/.initial={},
    figurename/.initial={zoombox},
    zoomboxarray/.style={
        execute at begin picture={
            \begin{scope}[
                spy using outlines={%
                    zoombox paths,
                    width=\imagewidth / \pgfkeysvalueof{/tikz/zoomboxarray columns} - (\pgfkeysvalueof{/tikz/zoomboxarray columns} - 1) / \pgfkeysvalueof{/tikz/zoomboxarray columns} * \pgfkeysvalueof{/tikz/zoomboxarray inner gap} -\pgflinewidth,
                    height=\pgfkeysvalueof{/tikz/zoomboxarray heightmultiplier} * (\imageheight / \pgfkeysvalueof{/tikz/zoomboxarray rows} - (\pgfkeysvalueof{/tikz/zoomboxarray rows} - 1) / \pgfkeysvalueof{/tikz/zoomboxarray rows} * \pgfkeysvalueof{/tikz/zoomboxarray inner gap}-\pgflinewidth),
                    magnification=3,
                    every spy on node/.style={
                        zoombox paths
                    },
                    every spy in node/.style={
                        zoombox paths
                    }
                }
            ]
        },
        execute at end picture={
            \end{scope}
     \gdef\patternnumber{0}
        },
        spymargin/.initial=0.2em, 
        zoomboxes xshift/.initial=1,
        zoomboxes right/.code=\pgfkeys{/tikz/zoomboxes xshift=1},
        zoomboxes left/.code=\pgfkeys{/tikz/zoomboxes xshift=-1},
        zoomboxes yshift/.initial=0,
        zoomboxes above/.code={
            \pgfkeys{/tikz/zoomboxes yshift=1},
            \pgfkeys{/tikz/zoomboxes xshift=0}
        },
        zoomboxes below/.code={
            \pgfkeys{/tikz/zoomboxes yshift=-1},
            \pgfkeys{/tikz/zoomboxes xshift=0}
        },
        caption margin/.initial=0ex, %
    },
    adjust caption spacing/.code={},
    image container/.style={
        inner sep=0pt,
        at=(image.north),
        anchor=north,
        adjust caption spacing
    },
    zoomboxes container/.style={
        inner sep=0pt,
        at=(image.north),
        anchor=north,
        name=zoomboxes container,
        xshift=\pgfkeysvalueof{/tikz/zoomboxes xshift}*(\imagewidth+\pgfkeysvalueof{/tikz/spymargin}),
        yshift=\pgfkeysvalueof{/tikz/zoomboxes yshift}*(\imageheight+\pgfkeysvalueof{/tikz/spymargin}+\pgfkeysvalueof{/tikz/caption margin}),
        adjust caption spacing
    },
    calculate dimensions/.code={
        \pgfpointdiff{\pgfpointanchor{image}{south west} }{\pgfpointanchor{image}{north east} }
        \pgfgetlastxy{\imagewidth}{\imageheight}
        \global\let\imagewidth=\imagewidth
        \global\let\imageheight=\imageheight
        \gdef\columncount{1}
        \gdef\rowcount{1}
        
    },
    image node/.style={
        inner sep=0pt,
        name=image,
        anchor=south west,
        append after command={
            [calculate dimensions]
            node [image container,subfigurename=\pgfkeysvalueof{/tikz/figurename}-image] {\phantomimage}
            node [zoomboxes container,subfigurename=\pgfkeysvalueof{/tikz/figurename}-zoom] {\phantomimage}
        }
    },
    color code/.style={
        zoombox paths/.append style={draw=#1}
    },
    connect zoomboxes/.style={
    spy connection path={\draw[draw=none,zoombox paths] (tikzspyonnode) -- (tikzspyinnode);}
    },
    help grid code/.code={
        \begin{scope}[
                x={(image.south east)},
                y={(image.north west)},
                font=\footnotesize,
                help lines,
                overlay
            ]
            \foreach \x in {0,1,...,9} {
                \draw(\x/10,0) -- (\x/10,1);
                \node [anchor=north] at (\x/10,0) {0.\x};
            }
            \foreach \y in {0,1,...,9} {
                \draw(0,\y/10) -- (1,\y/10);                        \node [anchor=east] at (0,\y/10) {0.\y};
            }
        \end{scope}
    },
    help grid/.style={
        append after command={
            [help grid code]
        }
    },
}

\newcommand\phantomimage{%
    \phantom{%
        \rule{\imagewidth}{\imageheight}%
    }%
}
\newcommand\zoombox[2][]{
    \begin{scope}[zoombox paths]
        \pgfmathsetmacro\xpos{
            (\columncount-1)*(\imagewidth / \pgfkeysvalueof{/tikz/zoomboxarray columns} + \pgfkeysvalueof{/tikz/zoomboxarray inner gap} / \pgfkeysvalueof{/tikz/zoomboxarray columns} ) + \pgflinewidth
        }
        \pgfmathsetmacro\ypos{
            (\rowcount-1) * (\imageheight / \pgfkeysvalueof{/tikz/zoomboxarray rows} + \pgfkeysvalueof{/tikz/zoomboxarray inner gap} / \pgfkeysvalueof{/tikz/zoomboxarray rows} ) + 0.5*\pgflinewidth
        }
        \edef\dospy{\noexpand\spy [
            #1,
            zoombox paths/.append style={
                black and white pattern=\patternnumber
            },
            every spy on node/.append style={#1},
            x=\imagewidth,
            y=\imageheight
        ] on (#2) in node [anchor=north west] at ($(zoomboxes container.north west)+(\xpos pt,-\ypos pt)$);}
        \dospy
        \pgfmathtruncatemacro\pgfmathresult{ifthenelse(\columncount==\pgfkeysvalueof{/tikz/zoomboxarray columns},\rowcount+1,\rowcount)}
        \global\let\rowcount=\pgfmathresult
        \pgfmathtruncatemacro\pgfmathresult{ifthenelse(\columncount==\pgfkeysvalueof{/tikz/zoomboxarray columns},1,\columncount+1)}
        \global\let\columncount=\pgfmathresult
        \ifblackandwhitecycle
            \pgfmathtruncatemacro{\newpatternnumber}{\patternnumber+1}
            \global\edef\patternnumber{\newpatternnumber}
        \fi
    \end{scope}
}

\newcommand{\new}[1]{{\color{black}{#1}}}

\newcommand{\methodshort}{{4DRotorGS}}
\newcommand{\method}{{4D-Rotor Gaussian Splatting}}
\newcommand{\methodgaussiansplattingshort}{{3DGS}}
\newcommand{\methodgaussiantrackingshort}{{Dynamic3DGS}}
\newcommand{\methodiclrshort}{{RealTime4DGS}}

\definecolor{wildwatermelon}{rgb}{0.99, 0.42, 0.52}
\definecolor{cottoncandy}{rgb}{1.0, 0.74, 0.85}
\definecolor{aqua}{rgb}{0.0, 1.0, 1.0}
\definecolor{columbiablue}{rgb}{0.61, 0.87, 1.0}
\definecolor{skyblue}{rgb}{0.53, 0.81, 0.92}
\definecolor{teagreen}{rgb}{0.82, 0.94, 0.75}
\definecolor{rose}{rgb}{1.0, 0.0, 0.5}
\definecolor{radicalred}{rgb}{1.0, 0.21, 0.37}
\definecolor{princetonorange}{rgb}{1.0, 0.56, 0.0}
\definecolor{pistachio}{rgb}{0.58, 0.77, 0.45}
\definecolor{pear}{rgb}{0.82, 0.89, 0.19}
\definecolor{deepskyblue}{rgb}{0.0, 0.75, 1.0}
\makeatletter
\DeclareRobustCommand\onedot{\futurelet\@let@token\@onedot}
\def\@onedot{\ifx\@let@token.\else.\null\fi\xspace}

\def\eg{\emph{e.g}\onedot} 
\def\ie{\emph{i.e}\onedot} 
 
 \def\vs{\emph{vs}\onedot}

\makeatother

\newcommand{\point}{\mathbf{x}}
\newcommand{\threeDPosition}{\point}

\newcommand{\threeDGaussian}{{G}}
\newcommand{\centerPos}{\boldsymbol{\mu}}
\newcommand{\centerPosScalar}{{\mu}}
\newcommand{\threeDCov}{\mathbf{\Sigma}}
\newcommand{\threeDScaling}{\mathbf{S}}
\newcommand{\threeDRotation}{\mathbf{R}}
\newcommand{\opacity}{o}
\newcommand{\twoDCov}{\mathbf{\Sigma'}}

\newcommand{\Jacobian}{\mathbf{J}}
\newcommand{\extrinsic}{\mathbf{V}}
\newcommand{\twoDGaussianFromThreeDG}{G'}
\newcommand{\pixelColor}{C}
\newcommand{\weightBlending}{\alpha}

\newcommand{\fourDGaussian}{\threeDGaussian_{4D}}
\newcommand{\fourDCenter}{\centerPos_{4D}}
\newcommand{\fourDCov}{\threeDCov_{4D}}
\newcommand{\fourDScaling}{\threeDScaling_{4D}}
\newcommand{\fourDRotation}{\threeDRotation_{4D}}

\newcommand{\rotorFour}{ \mathbf{r}}

\newcommand{\s}{s}
\newcommand{\bi}{b}
\newcommand{\p}{p}
\newcommand{\e}{\mathbf{e}}

\newcommand{\rotorNormFunc}{\mathcal{F}_{norm}}
\newcommand{\rotorConvertFunc}{\mathcal{F}_{map}}

\newcommand{\deltA}{\delta}
\newcommand{\length}{l}
\newcommand{\varepsiloN}{\varepsilon}
\newcommand{\DeltA}{\Delta}
\newcommand{\uSrcFourD}{u}
\newcommand{\uTargetFourD}{u'}

\newcommand{\threeDGaussianFromFourDG}{\threeDGaussian_{3D}}
\newcommand{\threeDCovFromfourDCov}{\threeDCov_{3D}}
\newcommand{\bigU}{\mathbf{U}}
\newcommand{\V}{\mathbf{V}}
\newcommand{\W}{\mathbf{W}}
\newcommand{\A}{\mathbf{A}}
\newcommand{\M}{\mathbf{M}}
\newcommand{\Z}{\mathbf{Z}}
\newcommand{\lambdA}{\lambda}
\newcommand{\timeT}{t}
\newcommand{\speed}{\mathbf{s}}

\newcommand{\loss}{L}

\newcommand{\ita}{\textit{a}}
\newcommand{\itb}{\textit{b}}
\newcommand{\itc}{\textit{c}}
\newcommand{\itd}{\textit{d}}

\newcommand{\itf}{\textit{f}}
\newcommand{\itg}{\textit{g}}
\newcommand{\ith}{\textit{h}}

\newcommand{\B}{B}
\newcommand{\h}{h}
\newcommand{\g}{g}

\copyrightyear{2024}
\acmYear{2024}
\setcopyright{acmlicensed}\acmConference[SIGGRAPH Conference Papers '24]{Special Interest Group on Computer Graphics and Interactive Techniques Conference Conference Papers '24}{July 27-August 1, 2024}{Denver, CO, USA}
\acmBooktitle{Special Interest Group on Computer Graphics and Interactive Techniques Conference Conference Papers '24 (SIGGRAPH Conference Papers '24), July 27-August 1, 2024, Denver, CO, USA}
\acmDOI{10.1145/3641519.3657463}
\acmISBN{979-8-4007-0525-0/24/07}

\begin{document}
\title{4D-Rotor Gaussian Splatting:
Towards Efficient Novel View Synthesis for Dynamic Scenes}

\author{Yuanxing Duan}
\authornote{Equal contribution}
\orcid{0009-0001-9877-5757}
\affiliation{%
 \institution{Peking University}
 \country{China}}
\email{mjdyx@pku.edu.cn}
\author{Fangyin Wei}
\authornotemark[1]
\orcid{0009-0008-4004-6659}
\affiliation{%
 \institution{Princeton University}
 \country{USA}}
\email{fwei@princeton.edu}
\author{Qiyu Dai}
\orcid{0000-0003-4659-2438}
\affiliation{%
 \institution{Peking University} 
 \country{China}}
 \affiliation{%
 	\institution{State Key Laboratory of General AI} 
 \country{China}}
\email{qiyudai@pku.edu.cn}
\author{Yuhang He}
\orcid{0009-0009-7254-2996}
\affiliation{%
 \institution{Peking University}
 \country{China}}
\email{2100014725@stu.pku.edu.cn}
\author{Wenzheng Chen}
\authornote{Co-corresponding authors}
\orcid{0009-0008-5623-1963}
\affiliation{%
 \institution{Peking University}
 \country{China}}
  \affiliation{%
 	\institution{NVIDIA}
 	\country{Canada}}
\email{wenzhengchen@pku.edu.cn}
\author{Baoquan Chen}
\authornotemark[2]
\orcid{0000-0003-4702-036X}
\affiliation{%
 \institution{Peking University} 
 \country{China}}
  \affiliation{%
 	\institution{State Key Laboratory of General AI} 
 \country{China}}
\email{baoquan@pku.edu.cn}

\begin{abstract}

We consider the problem of novel-view synthesis (NVS) for dynamic scenes. 
Recent neural approaches have accomplished exceptional NVS results for static 3D scenes, but extensions to 4D time-varying scenes remain non-trivial. 
Prior efforts often encode dynamics by learning a canonical space plus implicit or explicit deformation fields, which struggle in challenging scenarios like sudden movements or generating high-fidelity renderings.
In this paper, we introduce 4D Gaussian Splatting ({\methodshort}), a novel method that represents dynamic scenes with anisotropic 4D $XYZT$ Gaussians, inspired by the success of 3D Gaussian Splatting in static scenes~\cite{kerbl20233d}.
We model dynamics at each timestamp by temporally slicing the 4D Gaussians, which naturally compose dynamic 3D Gaussians and can be seamlessly projected into images.
As an explicit spatial-temporal representation, {\methodshort}  demonstrates powerful capabilities for modeling complicated dynamics and fine details—especially for scenes with abrupt motions.
We further implement our temporal slicing and splatting techniques in a highly optimized CUDA acceleration framework, achieving real-time inference rendering speeds of up to 277 FPS on an RTX 3090 GPU and 583 FPS on an RTX 4090 GPU. 
Rigorous evaluations on scenes with diverse motions showcase the superior efficiency and effectiveness of {\methodshort}, which consistently outperforms existing methods both quantitatively and qualitatively.

\end{abstract}

%


\begin{teaserfigure}
	\centering
	\includegraphics[page=1,trim={1.8cm 15.5cm 2cm 8.5cm},clip]{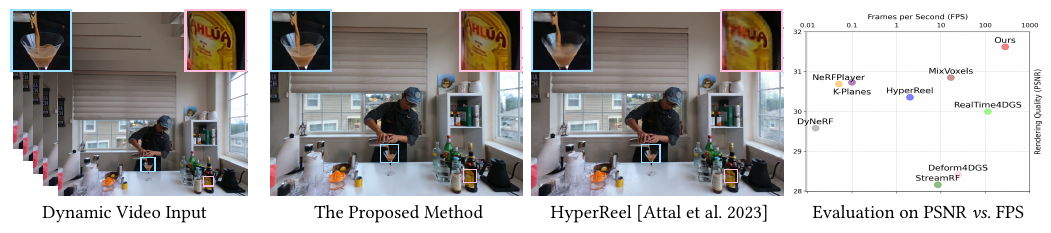}
	\captionof{figure}{We present 4D Rotor Gaussian Splatting, a novel approach that takes videos as input and synthesizes 
		dynamic novel-view images with fine details. The proposed method outperforms prior arts in rendering quality 
		and achieves 277 FPS inference speed on an RTX 3090 GPU at Plenoptic Dataset~\cite{li2022neural}. 
	}
\end{teaserfigure}

\maketitle

\section{Introduction}
\label{sec:intro}

Reconstructing 3D scenes from 2D images and synthesizing their appearance from novel views has been a long-standing goal in computer vision and graphics.
This task is pivotal in numerous industrial applications including film, gaming, and VR/AR, where there is a substantial demand for \emph{high-speed, photo-realistic} rendering effects.
%
The task falls into two different scene types: static scenes where objects are still across all images~\cite{mildenhall2020nerf,knapitsch2017tanks,hedman2018deep,barron2022mip} and dynamic scenes where scene contents exhibit temporal variations~\cite{pumarola2021d,park2021hypernerf,li2022neural,wu2020multi,cheng2023dna}.
%
While the former has witnessed significant progress recently,  efficient and accurate NVS for dynamic scenes remains challenging due to the complexities introduced by the temporal dimension and diverse motion patterns.  

A variety of methods have been proposed to tackle the challenges in dynamic NVS.
Some methods model the 3D scene and its dynamics jointly~\cite{du2021neural,gao2021dynamic}. 
However, these methods often fall short in preserving fine details in the NVS renderings due to the complexity caused by the highly entangled spatial and temporal dimensions. 
%
Alternatively, many existing techniques~\cite{park2021nerfies,park2021hypernerf,tretschk2021non, pumarola2021d,fang2022fast} decouple dynamic scenes by learning a static canonical space and then predicting a deformation field to account for the temporal variations. Nonetheless, this paradigm struggles in capturing complex dynamics such as objects appearing or disappearing suddenly.
More importantly, prevailing methods on dynamic NVS mostly build upon volumetric rendering~\cite{mildenhall2020nerf} which requires dense sampling on millions of rays. As a consequence, these methods typically cannot support real-time rendering speed even for static scenes~\cite{li2022neural, song2023nerfplayer}.

Recently, 3D Gaussian Splatting (\methodgaussiansplattingshort)~\cite{kerbl20233d} has emerged as a powerful tool for efficient  NVS of static scenes.
By explicitly modeling the scene with 3D Gaussian ellipsoids and employing fast rasterization technique, it achieves photo-realistic NVS in real time.  Inspired by this, 
we propose to lift Gaussians from 3D to 4D  and provide a novel spatial-temporal representation that enables NVS for more challenging dynamic scenes. 

\begin{figure}
	\centering
	\includegraphics[width=.85\linewidth,trim={0cm 0cm 0cm 0cm},clip]{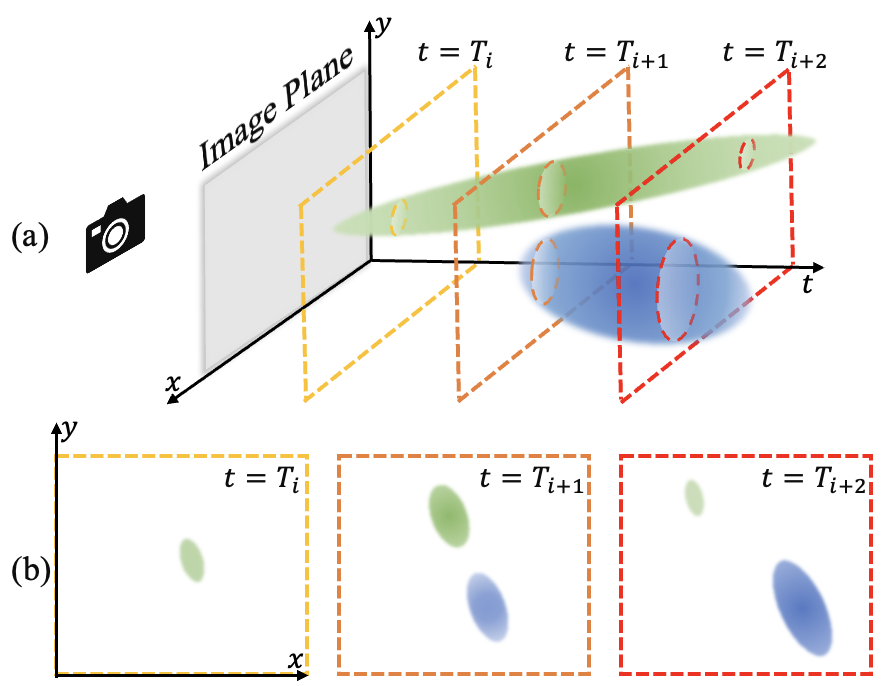}
	\caption{\label{fig:2dmotion} \textbf{A Simplified 2D Illustration of the Proposed Temporal Slicing.} \textbf{(a)} We model 2D dynamics with 3D $XYT$ ellipsoids and slice them with different time queries.  \textbf{(b)} The sliced 3D ellipsoids form 2D  dynamic ellipses at each timestamp. }
\end{figure}

Our key observation is that 3D scene dynamics at each timestamp can be viewed as 4D spatial-temporal Gaussian ellipsoids sliced with different time queries.  
Fig.~\ref{fig:2dmotion} illustrates a simplified $XYT$  case: the dynamics in 2D $XY$ space at time $T_i$ is equivalent to building 3D $XYT$ Gaussians sliced by the $t=T_i$ plane. 
Analogously, we extend 3D Gaussians to 4D $XYZT$ space to model dynamic 3D scenes. The temporally sliced 4D Gaussians compose 3D Gaussians that can be projected to 2D screens via fast rasterization, 
inheriting both exquisite rendering effects and high rendering speed from {\methodgaussiansplattingshort}.
%
%
Moreover, extending the prune-split mechanism in the temporal dimension makes 4D Gaussians particularly suitable for representing complex dynamics, including abrupt appearances or disappearances.

It is non-trivial to lift 3D Gaussians into 4D space, where tremendous  challenges exist in the design of the 4D rotation, slicing, as well as the joint spatial-temporal optimization scheme. 
%
We draw inspiration from geometric algebra and carefully choose 4D rotor~\cite{bosch2020n} to represent 4D rotation, which is a spatial-temporal separable rotation representation. 
%
%
Notably, rotor representation accommodates both 3D and 4D rotation: when its temporal dimensions are set to zero, it becomes equivalent to a quaternion and can represent 3D spatial rotation as well.
%
Such adaptability grants our method 
the flexibility to model both dynamic and static scenes. In other words, {\methodshort} is a generalizable form of {\methodgaussiansplattingshort}: when closing the temporal dimension, our {\methodshort} reduces to {\methodgaussiansplattingshort}.

%
%
%
We enhance the optimization strategies in  {\methodgaussiansplattingshort} and 
introduce two new regularization terms to stabilize and improve the dynamic reconstruction. 
We first propose an entropy loss that pushes the opacity of Gaussians towards either one or zero, which proves effective to remove ``floaters'' in our experiments. 
We further introduce a novel 4D consistency loss to regularize the motion of Gaussians  and yield more consistent dynamics reconstruction.
%
%
%
Experiments show that both terms notably  improve the rendering quality. 
%

%

While existing Gaussian-based methods~\cite{luiten2023dynamic,yang2023deformable,wu20234d,yang2023real} are mostly based on PyTorch~\cite{paszke2019pytorch}, we further develop a highly optimized CUDA framework  with careful engineering designs for fast training and inference speed. 
Our framework supports rendering 1352$\times$1014 videos at an unprecedented 583 FPS on an RTX 4090 GPU and 277 FPS on an RTX 3090 GPU.  
%
%
We conduct extensive experiments on two datasets spanning a wide range of settings and motion patterns, including synthetic monocular videos~\cite{pumarola2021d} and realistic multi-camera videos~\cite{li2022neural}.
Quantitative and qualitative evaluations demonstrate the distinct advantages over preceding methods, including the new state-of-the-art rendering quality and speed. Code is available at \href{https://github.com/weify627/4D-Rotor-Gaussians}{https://github.com/weify627/4D-Rotor-Gaussians}.

\section{Related Work}
\label{sec:formatting}


In this section, we mainly review optimization-based novel-view synthesis (NVS) methods, where the input is a set of posed images and the output is new appearance from a novel viewpoint. 
We first describe NVS for static scenes, then talk about its dynamic extensions. Lastly, we discuss recent Gaussian-based NVS methods.


\paragraph{Static NVS}
Prior work formalizes light-field~\cite{levoy1996light} or lumigraph~\cite{gortler1996lumigraph,buehler2001unstructured} that generate novel-view images by interpolating existing views, which require densely captured images to acquire realistic renderings.
Other classical methods exploit 
geometric proxies such as mesh and volume to reproject and blend contents from source images onto novel views.
Mesh-based methods~\cite{debec1996modeling,riegler2020free,thies2019deferred,waechter2014let,wood2023surface} represent the scenes with surfaces that support efficient rendering but are hard to optimize for complex geometries. 
%
Volume-based methods use voxel grids~\cite{kutulakos2000theory,penner2017soft,seitz1999photorealistic} or multi-plane images~\cite{flynn2019deepview,mildenhall2019local,srinivasan2019pushing,zhou2018stereo}, which provide delicate rendering effects but are memory-inefficient or limited to small view changes. 
Recent trend of NVS is spearheaded by Neural Radiance Fields (NeRFs)~\cite{mildenhall2020nerf} that 
achieves photo-realistic rendering.
Since then, a series of efforts have emerged to accelerate training~\cite{muller2022instant,fridovich2022plenoxels,chen2022tensorf}, enhance rendering quality~\cite{barron2021mip,verbin2022ref}, or extend to more challenging scenarios such as reflection and refraction~\cite{kopanas2022neural,bemana2022eikonal,Yan_2023_CVPR}.
Still, most of these methods rely on volume rendering that requires sampling millions of rays and hinders real-time rendering~\cite{mildenhall2020nerf,barron2021mip,chen2022tensorf}.

\begin{figure*}
	\centering
	\includegraphics[width=\linewidth]{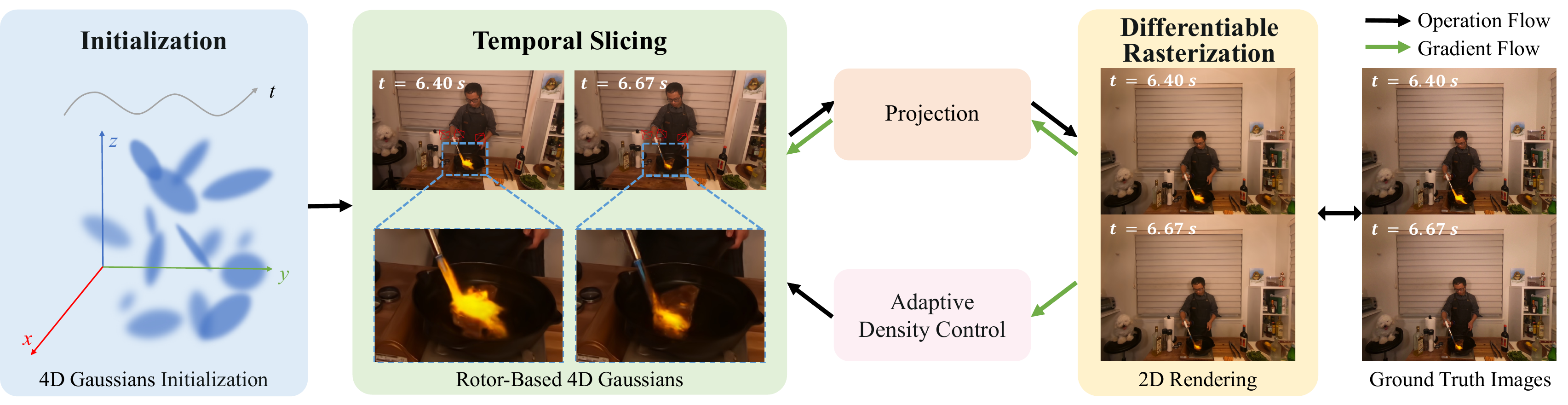}
 \caption{\label{fig:framework}\textbf{Framework Overview.} After initialization, we first temporally slice the 4D Gaussians whose spatio-temporal movements are modeled with rotors. The dynamics such as the flickering flames naturally evolve through time, even within a short period of 0.27 second. The sliced 3D Gaussians are then projected onto 2D using differentiable rasterization. The gradients from image loss are back-propagated and guide the adaptive density control of 4D Gaussians. 
 }
\end{figure*}

\paragraph{Dynamic NVS} 
New challenges are posed by dynamic NVS due to the temporal variations in the input images.
%
Traditional methods estimate varying geometries such as surfaces~\cite{li2012temporally,collet2015high} and depth maps~\cite{kanade1997virtualized,zitnick2004high} to account for dynamics.
%
Inspired by NeRFs, recent work typically models dynamic scenes with neural representations.
A branch of methods implicitly models the dynamics by adding a temporal input or a latent code~\cite{du2021neural,gao2021dynamic}. 
Another line of works~\cite{park2021nerfies,park2021hypernerf,tretschk2021non, pumarola2021d,fang2022fast} explicitly models deformation fields that map 3D points at arbitrary timestamp into a canonical space. 
Other techniques decompose a scene into static and dynamic components~\cite{song2023nerfplayer}, using key frames to reduce redundancy~\cite{li2022neural,attal2023hyperreel}, estimating a flow field~\cite{li2021nsff,guo2023forward,tian2023mononerf}, or exploiting 4D grid-based representations~\cite{li2022streaming, fridovich2023k,cao2023hexplane,wang2023mixed,fang2022fast,gan2023v4d,shao2023tensor4d}.
%
%
The common issues of dynamic scene modeling are the complexities from the spatial-temporal entanglement, as well as the additional memory and time cost for the temporal dimension. 
%

\paragraph{Gaussian-Based NVS}
The recent seminal work~\cite{zhang2022differentiable, xu2022point,abou2022particlenerf,kerbl20233d} models static scenes with 3D Gaussians whose positions and appearance are learned through a differentiable splatting-based renderer.
%
Particularly, 3D Gaussian Splatting ({\methodgaussiansplattingshort})~\cite{kerbl20233d} achieves impressive fast training and real-time rendering thanks to its Gaussian split/clone operations and the fast splatting-based rendering technique. 
Our work draws inspiration from {\methodgaussiansplattingshort} but lifts 3D Gaussians into 4D space and focuses on dynamic scenes. 
Several concurrent works also extend  {\methodgaussiansplattingshort}  to model dynamics. 
{\methodgaussiantrackingshort}~\cite{luiten2023dynamic} directly learns the temporal motion and rotation of each 3D Gaussian along time, 
which makes it suitable for  dynamic tracking applications~\cite{wang2023tracking}.
Similarly,~\cite{yang2023deformable,wu20234d} utilize MLPs to predict the temporal movement. However, it is challenging for these methods to represent dynamic contents that suddenly appear or disappear.
%
%
%
%
%
%
%
%
Similar to us, {\methodiclrshort}~\cite{yang2023real} also leverages 4D Gaussian representation to model 3D dynamics.  
They choose a dual-quaternion based 4D rotation formulation that is less interpretable and lacks the spatial-temporal separable property compared to rotor-base representation. 
\new{Moreover, we further introduce new regularization terms which prove effective in improving rendering quality.}

\section{Method}
In this section, we first review the 3D Gaussian Splatting ({\methodgaussiansplattingshort}) method~\cite{kerbl20233d} in Sec.~\ref{sec:3dgs}. 
We then describe our 4G Gaussian Splatting algorithm in Sec.~\ref{sec:ourmethod}, where we present rotor-based 4D Gaussian representation in Sec.~\ref{sec:rotor} and discuss the temporal slicing technique for differentiable real-time rasterization in Sec.~\ref{sec:rendering}.
Finally, we introduce our dynamic optimization strategies in Sec.~\ref{sec:optim}.

\subsection{Preliminary of 3D Gaussian Splatting}
\label{sec:3dgs}
3D Gaussian Splatting ({\methodgaussiansplattingshort})~\cite{kerbl20233d} has demonstrated real-time, state-of-the-art rendering quality on static scenes. 
It models a scene with a dense cluster of $N$ anisotropic 3D Gaussian ellipsoids. 
Each Gaussian is represented by a full 3D covariance matrix $\threeDCov$ and its center position $\centerPos$:
\begin{equation}
\label{eq:3DGS}\threeDGaussian(\point) = e^{-\frac{1}{2} (\point-\centerPos)^T \threeDCov^{-1} (\point-\centerPos)}.
\end{equation}
To ensure a valid positive semi-definite covariance matrix during the optimization, $\threeDCov$ is decomposed into the scaling matrix $\threeDScaling$ and the rotation matrix $\threeDRotation$ to characterize the geometry of a 3D Gaussian:
\begin{equation}
\threeDCov = \threeDRotation \threeDScaling \threeDScaling^T \threeDRotation^T,
\end{equation}
where $\threeDScaling = \text{diag}(s_x, s_y, s_z) \in \mathbb{R}^3$ and $\threeDRotation \in SO(3)$ are stored as a 3D vector and quaternion, respectively.
Beyond $\centerPos$, $\threeDScaling$ and $\threeDRotation$, each Gaussian maintains additional learnable parameters including 
opacity $\opacity \in (0, 1)$ and spherical harmonic (SH) coefficients in $\mathbb{R}^k$ representing view-dependent colors ($k$ is related to SH order). During optimization, 3DGS adaptively controls the Gaussian distribution 
by splitting and cloning Gaussians in regions with large view-space positional gradients, as well as the culling of Gaussians that exhibit near-transparency.

Efficient rendering and parameter optimization in {\methodgaussiansplattingshort} are powered by the differentiable tile-based rasterizer. Firstly, 3D Gaussians are projected to 2D space by computing the camera space covariance matrix $\twoDCov= \Jacobian \extrinsic \threeDCov \extrinsic^T \Jacobian^T$,
where $\Jacobian$ is the Jacobian matrix of the affine approximation of the projection transformation, and $\extrinsic$ is the extrinsic camera matrix. Then, 
the color of each pixel on the image plane is calculated by blending Gaussians sorted by their depths:
\begin{equation}
\pixelColor = \sum_{i=1}^N c_i \weightBlending_i \prod_{j=1}^{i-1} (1-\weightBlending_j),
\end{equation}
where $c_i$ is the color of the $i$-th 3D Gaussian $\threeDGaussian_i$, $\weightBlending_i = \opacity_i \twoDGaussianFromThreeDG_i$ with $\opacity_i$ and $\twoDGaussianFromThreeDG_i$ being the opacity and 2D projection of $\threeDGaussian_i$, respectively. 
Please refer to {\methodgaussiansplattingshort}~\cite{kerbl20233d} for more details.

\subsection{4D Gaussian Splatting }
\label{sec:ourmethod}

We now discuss our {\method} ({\methodshort}) algorithm, as illustrated in Fig.~\ref{fig:framework}. 
Specifically, we model the 4D Gaussian with rotor-based rotation representation (Sec.~\ref{sec:rotor}) and slice the time dimension to generate dynamic 3D Gaussians at each timestamp (Sec.~\ref{sec:rendering}).
The 3D Gaussian ellipsoids sliced at each timestamp can be efficiently rasterized onto a 2D image plane for real-time and  high-fidelity rendering of dynamic scenes.

\subsubsection{Rotor-Based 4D Gaussian Representation}
\label{sec:rotor}

Analogous to 3D Gaussians, a 4D Gaussian can be expressed with a 4D center position $\fourDCenter = (\centerPosScalar_{x}, \centerPosScalar_{y}, \centerPosScalar_{z}, \centerPosScalar_t)$ and a 4D covariance matrix $\fourDCov$ as
\begin{equation}
\fourDGaussian(\point) = e^{-\frac{1}{2} (\point-\fourDCenter)^T \fourDCov^{-1} (\point-\fourDCenter)}.
\end{equation}
The covariance $\fourDCov$ can be further factorized into the 4D scaling $\fourDScaling$ and the 4D rotation $\fourDRotation$ as
\begin{equation}
\fourDCov = \fourDRotation \fourDScaling \fourDScaling^T \fourDRotation^T.
\end{equation}
While modeling $\fourDScaling = \text{diag}(s_x, s_y, s_z, s_t)$ is straightforward, seeking a proper 4D rotation representation for $\fourDRotation$ is challenging.
In geometric algebra, rotations of high-dimensional vectors can be described using rotors~\cite{bosch2020n}. Motivated by this, we introduce 4D rotors to characterize the 4D rotations. A 4D rotor $\rotorFour$ is composed of 8 components based on a set of basis:
\begin{align}
    \rotorFour &= \s + \bi_{01} \e_{01} + \bi_{02} \e_{02}  + \bi_{12} \e_{12} + \bi_{03} \e_{03} \nonumber \\
       \label{eq:rotor4} &\quad + \bi_{13} \e_{13} + \bi_{23} \e_{23} + \p \e_{0123}, 
\end{align}
where $\e_{0123}=\e_0 \wedge \e_1 \wedge\e_2 \wedge \e_3$, and $\e_{ij}=\e_i \wedge \e_j$ is the exterior product between 4D axis $\e_i$ ($i\in\{0,1,2,3\}$) corresponding to $x,y,z,t$ that form the orthonormal basis in 4D Euclidean space. Therefore, a 4D rotation can be determined by 8 coefficients 
$\left(\s, \bi_{01}, \bi_{02}, \bi_{12}, \bi_{03}, \bi_{13}, \bi_{23}, \p\right)$.

 Analogous to quaternion, a rotor $\rotorFour$ can also be converted into a 4D rotation matrix  $\fourDRotation$ with proper normalization function $\rotorNormFunc$ and rotor-matrix mapping function $\rotorConvertFunc$. We carefully derive a numerically stable normalization method for rotor transformation and provide details of the two functions in Supplementary Material:
\begin{equation}
	\fourDRotation  =  \rotorConvertFunc (\rotorNormFunc(\rotorFour)).
\end{equation}

\begin{figure}
	\centering
	\includegraphics[width=\linewidth,trim={0cm 0cm 0cm 0cm},clip]{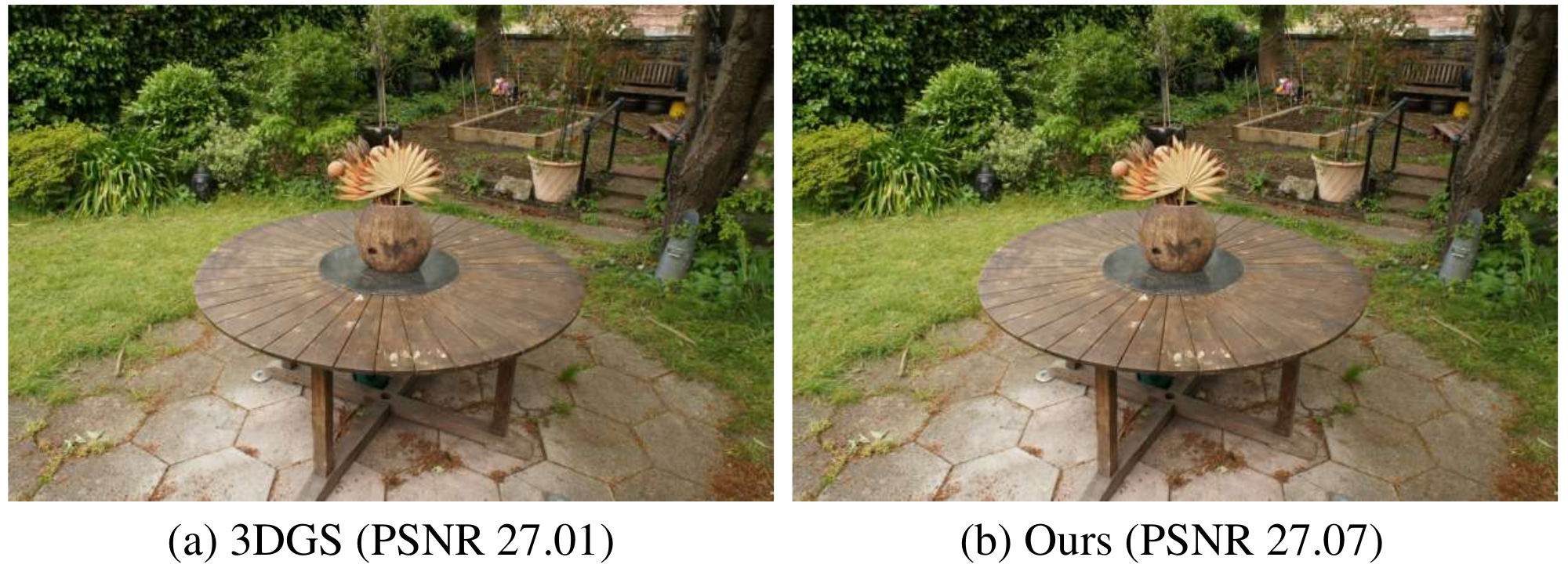}
	\caption{\label{fig:3dcomp} \textbf{Modeling 3D Static Scenes.} Our rotor-based representation enables both 3D static and 4D dynamic scene modeling, matching the results of 3DGS on 3D scenes.}
\end{figure}

Our rotor-based 4D Gaussian offers a well-defined, interpretable rotation representation: the first four components encode the 3D spatial rotation, and the last four components define spatio-temporal rotation, \ie, spatial translation. 
In particular, setting the last four components to zero effectively reduces $\rotorFour$ to a quaternion for 3D spatial rotation, thereby enabling our framework to model both static and dynamic scenes. 
Fig.~\ref{fig:3dcomp} presents an example where our result on a static 3D scene matches that of 3DGS~\cite{kerbl20233d}.

Alternatively, concurrent work~\cite{yang2023real} also models dynamic scenes with 4D Gaussian. However, they represent 4D rotation with two entangled isoclinic quaternions~\cite{cayley1894collected}.
As a result, their spatial and temporal rotations are tightly coupled, and it is unclear how to constrain and normalize this alternative rotation representation during optimization to model static 3D scenes.

\subsubsection{Temporally-Sliced 4D Gaussians Splatting}
\label{sec:rendering}

We now describe how to slice 4D Gaussians into 3D space.
Given that $\fourDCov$ and its inverse $\fourDCov^{-1}$ are both symmetric matrices, we define
\begin{equation}
\fourDCov = \left(\begin{matrix}
        \bigU   & \V \\
        \V^T & \W \\
\end{matrix}\right)
\text{~and~}
\fourDCov^{-1} = \left(\begin{matrix}
        \A   & \M \\
        \M^T & \Z      \\
    \end{matrix}\right),
\end{equation}
where both $\bigU$ and $\A$ are $3\times3$ matrices. Then, given a time $\timeT$, the projected 3D Gaussian is obtained as (detailed derivation in the Supplementary Material):
\begin{equation}
\label{eq:projected3dgs_main}
	\threeDGaussianFromFourDG(\threeDPosition, \timeT) = \text{e}^{-\frac{1}{2}\lambdA (\timeT - \centerPosScalar_t)^2}\text{e}^{-\frac{1}{2}\left[\threeDPosition - \centerPos(\timeT)\right]^T\threeDCovFromfourDCov^{-1}\left[\threeDPosition - \centerPos(\timeT)\right]},
\end{equation}
where
\begin{equation}
\begin{aligned}
\lambdA &= \W^{-1}, \\
\threeDCovFromfourDCov &= \A^{-1} = \bigU - \dfrac{\V \V^T}{\W}, \\ 
\centerPos(\timeT) &= (\centerPosScalar_x, \centerPosScalar_y, \centerPosScalar_z)^T + (\timeT - \centerPosScalar_t)\dfrac{\V}{\W}.
	\label{eq:center-motion}
\end{aligned}
\end{equation}

Compared with Eq.~\ref{eq:3DGS} in the original 3DGS~\cite{kerbl20233d}, the sliced 3D Gaussian in Eq.~\ref{eq:projected3dgs_main} contains a temporal decay term $\text{e}^{-\frac{1}{2}\lambdA(\timeT-\centerPosScalar_t)^2}$.  
As $\timeT$ goes by, a Gaussian point first appears once $\timeT$ is sufficiently close to its temporal position $\centerPosScalar_t$ and starts to grow. It reaches the peak opacity when $\timeT=\centerPosScalar_t$. After that, the 3D Gaussian gradually shrinks in density until vanishing when $\timeT$ is sufficiently far from $\centerPosScalar_t$. By controlling the temporal position and scaling factor, 4D Gaussian can represent challenging dynamics,~\eg, motions that suddenly appear or disappear.  
During rendering, we filter points that are too far from current time, where the threshold for visibility $\lambdA(\timeT-\centerPosScalar_t)^2$ is empirically set to 16. 

Moreover, the sliced 3D Gaussian exhibits a new motion term of $(\timeT - \centerPosScalar_t){\V}/{\W}$ added to the center position $(\centerPosScalar_x, \centerPosScalar_y, \centerPosScalar_z)^T$. 
In theory, linear movement of a 3D Gaussian emerges from our 4D slicing operation. 
We assume in a small time interval, all motions can be approximated by linear motions, and more complex non-linear cases can be represented by combination of multiple Gaussians. Further, ${\V}/{\W}$ indicates the motion speed in the current timestamp. Therefore, by modeling a scene with our framework, we can acquire speed field for free. We visualize the optical flow in Fig.~\ref{fig:flow}. 

Finally, following {\methodgaussiansplattingshort}~\cite{kerbl20233d}, we project the sliced 3D Gaussians to the 2D image plane in depth order and perform the fast differentiable rasterization to obtain the final image. We implement rotor representation and slicing in a high-performance CUDA framework and achieve  much higher rendering speed compared to PyTorch implementation.

\subsection{Optimization Schema}
\label{sec:optim}

When lifting 3D Gaussians into 4D space, the increased dimension extends the freedom of Gaussian points.  Therefore, we introduce two regularization terms to help stabilize the training process: \emph{entropy loss} and \emph{4D consistency loss}.

\subsubsection{Entropy Loss}  

Similar to NeRFs, each Gaussian point has a learnable opacity term $\opacity_i$ and the volume rendering formula is applied to composite the final color.  Ideally, a Gaussian point should be close to the object surface, and in most cases its opacity should be close to one. Therefore, we encourage the opacity to be close to one or close to zero by adding an  entropy loss, and by default Gaussians with near-zero opacity will be pruned during training:
\begin{align}
	\loss_{\text{entropy}} = & \frac{1}{N} \sum_{i=1}^N -\opacity_i log(\opacity_i).
\end{align}

We find that $\loss_{\text{entropy}}$ helps condense Gaussian points and filter noisy floaters, which is very useful when training with sparse views. 

\subsubsection{4D Consistency Loss} 

 Intuitively, nearby Gaussians in 4D space should have similar motions. We further regularize the 4D Gaussian points by adding the 4D spatial-temporal consistency loss. Recall when slicing a 4D Gaussian at a given time $\timeT$, a speed term $\speed$ is derived. Thus, given the $i$-th Gaussian point, we gather $K$ nearest 4D points from its neighbour space $\Omega_i$ and regularize their motions to be consistent:
\begin{align}
	\loss_{\text{consistent4D}} = & \frac{1}{N} \sum_{i=1}^N \left|\left|  \speed_i -   \frac{1}{K} \sum_{j \in \Omega_i} \speed_j   \right|\right |_1.
\end{align}
For 4D Gaussians, 4D distance is a better metric for point similarity than 3D distance, because points that are neighbors in 3D do not necessarily follow the same motions,~\eg, when they belong to two objects with different motions. Note that calculating 4D nearest neighbors is uniquely and naturally enabled in our 4D representation, which cannot be exploited by deformation-based methods~\cite{wu20234d}. The ``nearest'' is defined with Euclidean distance. We balance the different scales of each dimension by dividing with the corresponding spatial and temporal scene scales.

\subsubsection{ Total Loss} 

We follow original {\methodgaussiansplattingshort}~\cite{kerbl20233d} and add $L_1$ and SSIM losses between the rendered images and ground truth images.  Our final loss is defined as:
\begin{align}
	\loss = & (1 - \lambda_1) L_1  + \lambda_1 L_{ssim} + \lambda_2 	\loss_{\text{entropy}} +\lambda_3 	\loss_{\text{consistent4D}}.
\end{align}

\subsubsection{Optimization Framework}
\label{sec:framework}
We implement two versions of our method: one using PyTorch for fast development and one highly-optimized equivalent in C++ and CUDA for fast training and inference.
Compared to the PyTorch version, our CUDA acceleration allows to render at an unprecedented 583 FPS at 1352$\times$1014 resolution on a single NVIDIA RTX 4090 GPU.
Further, our CUDA framework also accelerates training by 16.6x. For benchmarking with baselines, we also test our framework on an RTX 3090 GPU and achieve 277 FPS, which significantly outperforms current state of the art (114 FPS~\cite{yang2023real}).

\definecolor{col1}{HTML}{e0d291}
\definecolor{col2}{HTML}{d66079}
\newcommand\videozoomed[7]{
    \begin{tikzpicture}[
    zoomboxarray,
    zoomboxes below,
    connect zoomboxes,
    zoombox paths/.append style={thick}]
        \node[image node]{\includegraphics[width=0.19\textwidth]{fig/ple-origin/#1.png}};
        \zoombox[magnification=#2,color code=cottoncandy]{#3,#4}  
        \zoombox[magnification=#5,color code=columbiablue]{#6,#7} %
    \end{tikzpicture}
}

\begin{figure*}[t]\centering
    \def\arraystretch{1}
    \setlength{\tabcolsep}{3pt}
    \begin{tabular}{@{}c@{}c@{}c@{}c@{}c@{}}
    \videozoomed{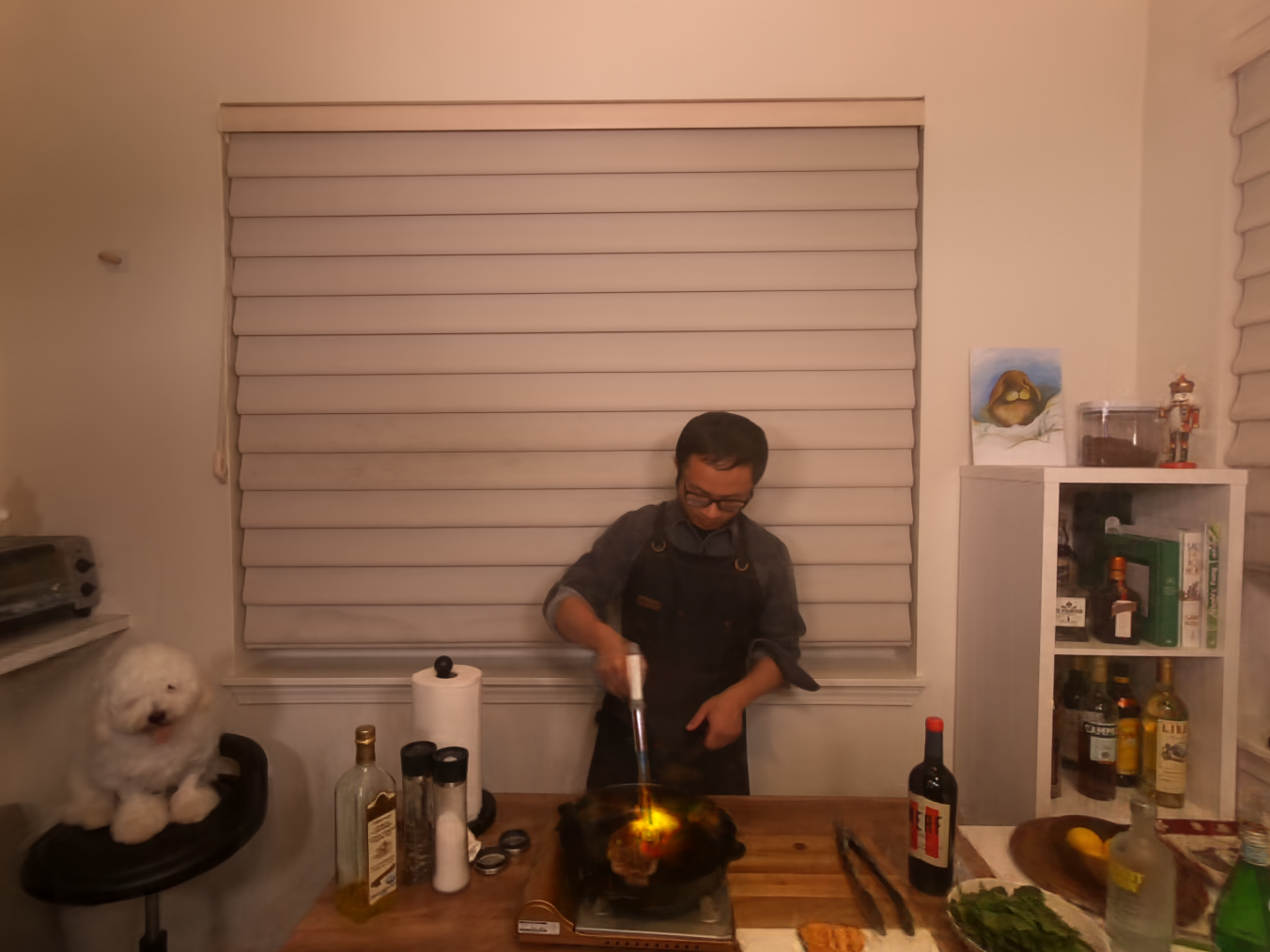}{2.5}{0.13}{0.23}{3.5}{0.5}{0.14} &
    \videozoomed{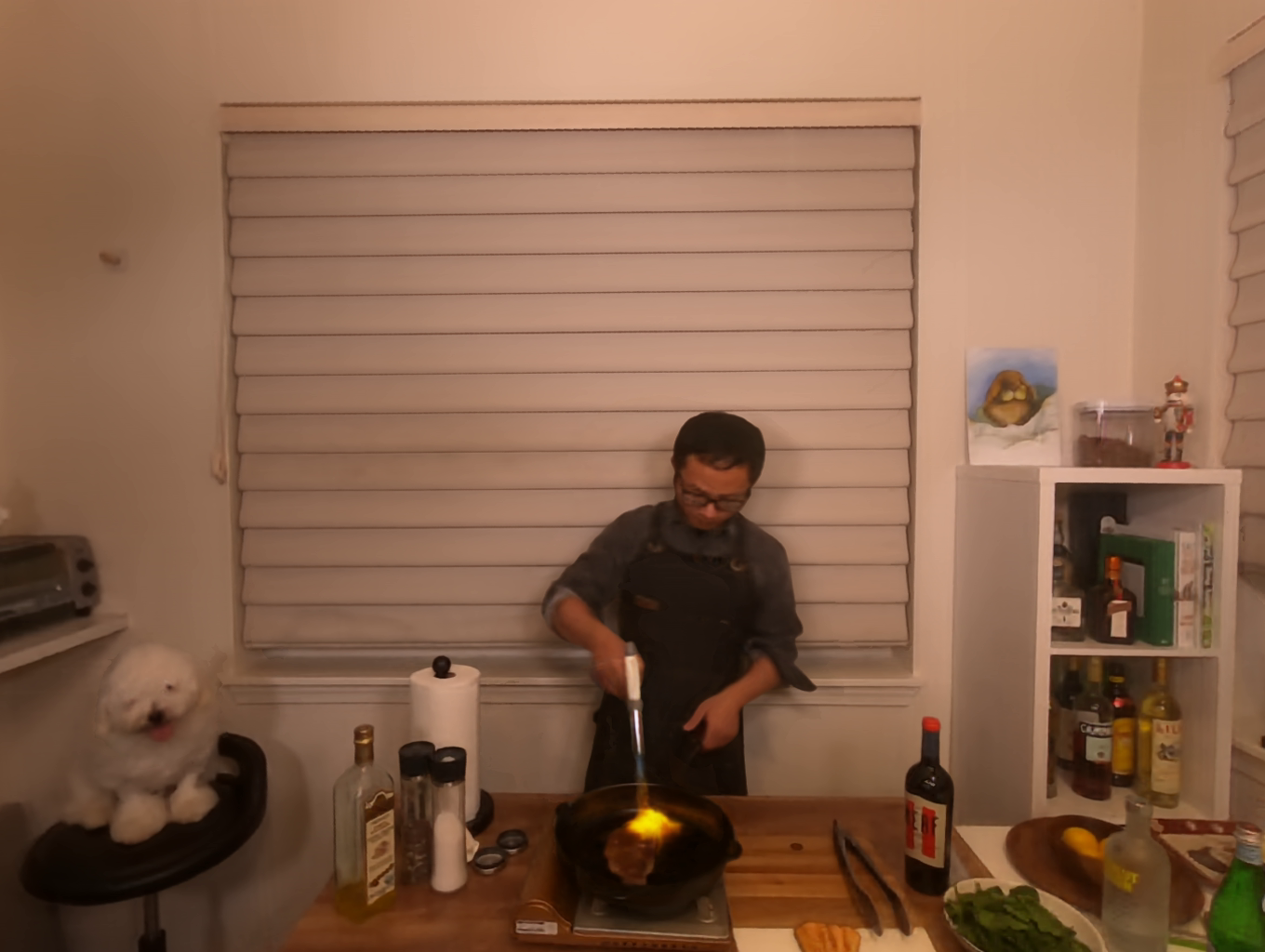}{2.5}{0.13}{0.23}{3.5}{0.5}{0.14} &
    \videozoomed{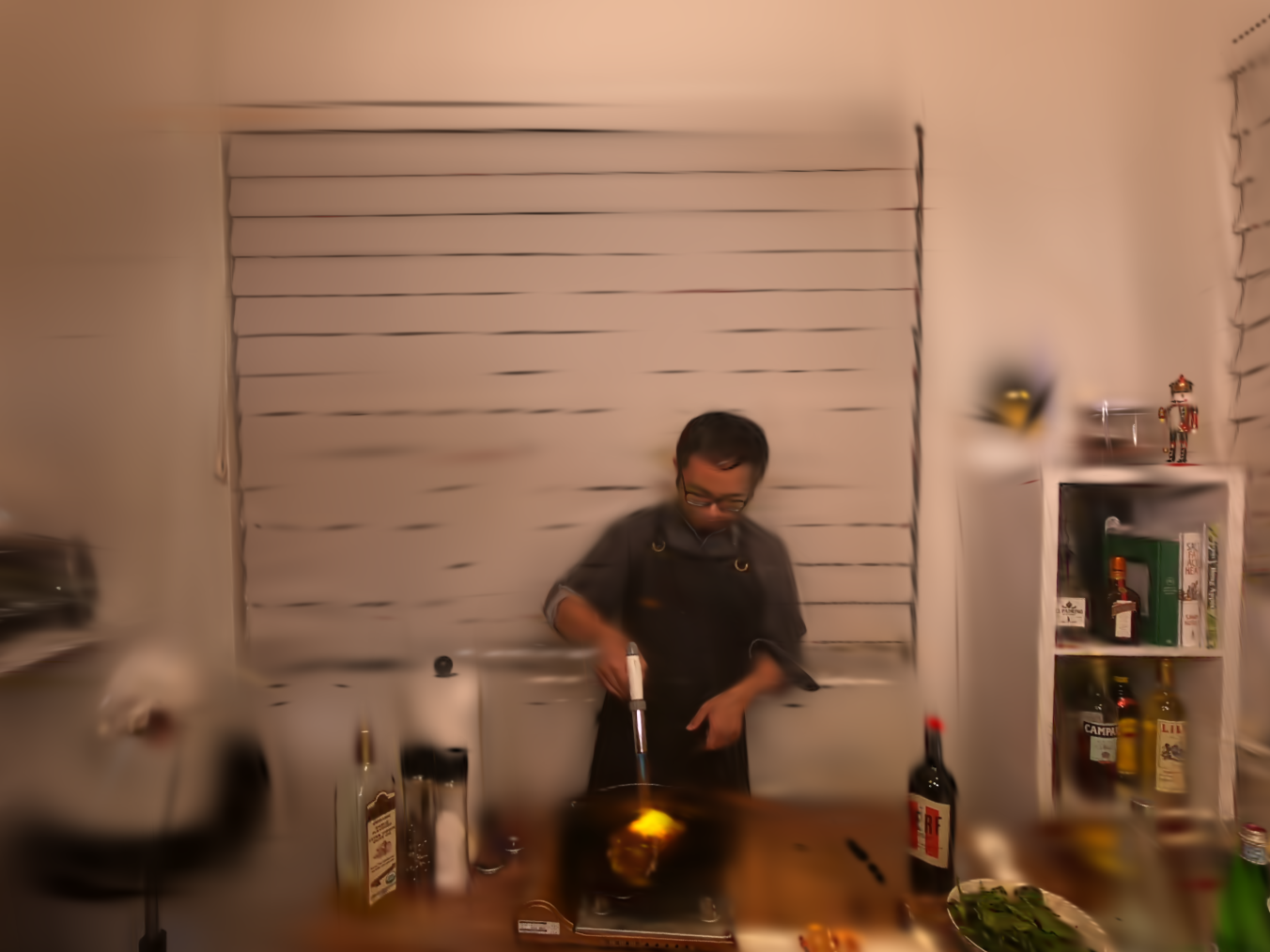}{2.5}{0.13}{0.23}{3.5}{0.5}{0.14} &
    \videozoomed{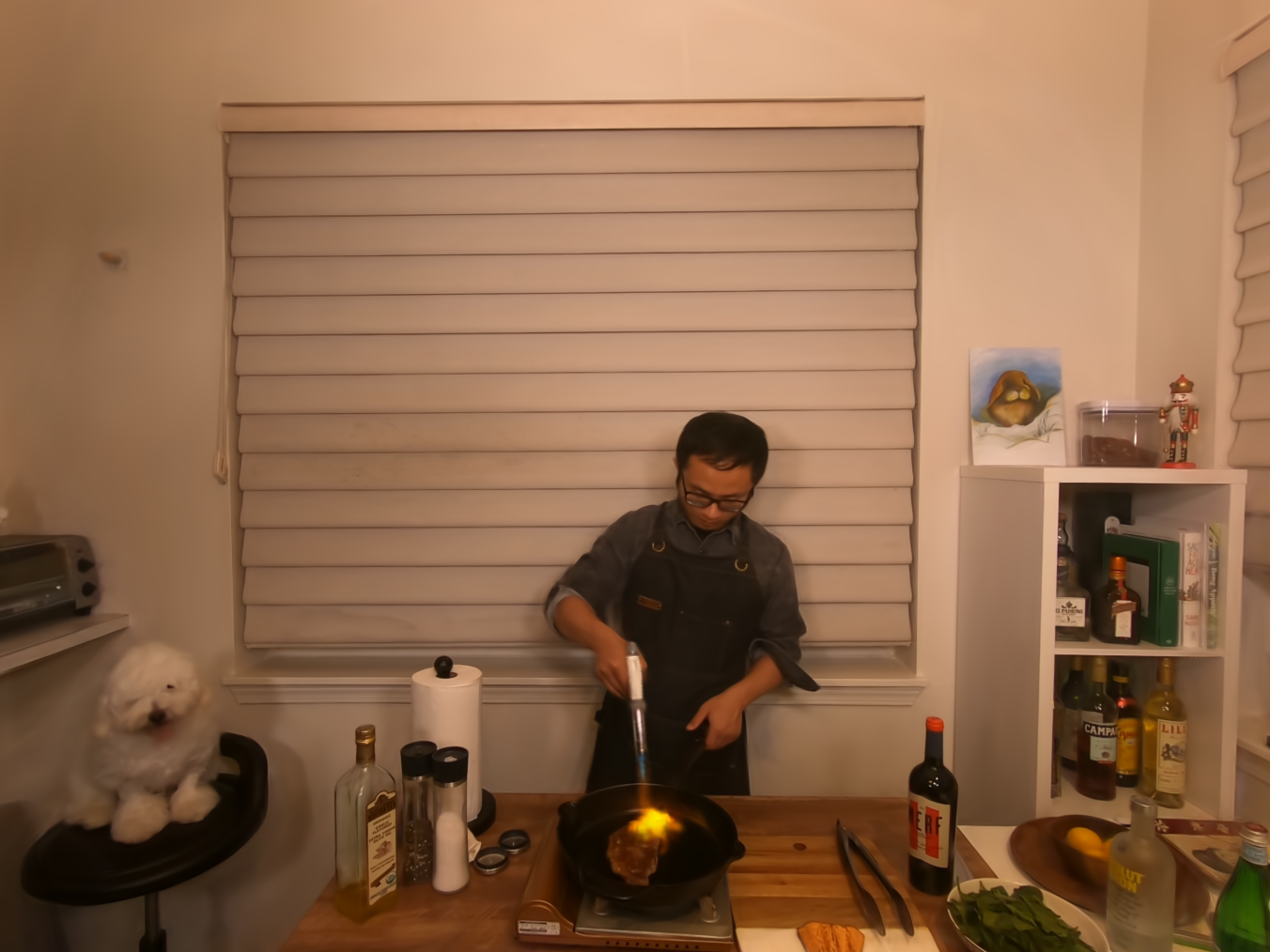}{2.5}{0.13}{0.23}{3.5}{0.5}{0.14} &
    \videozoomed{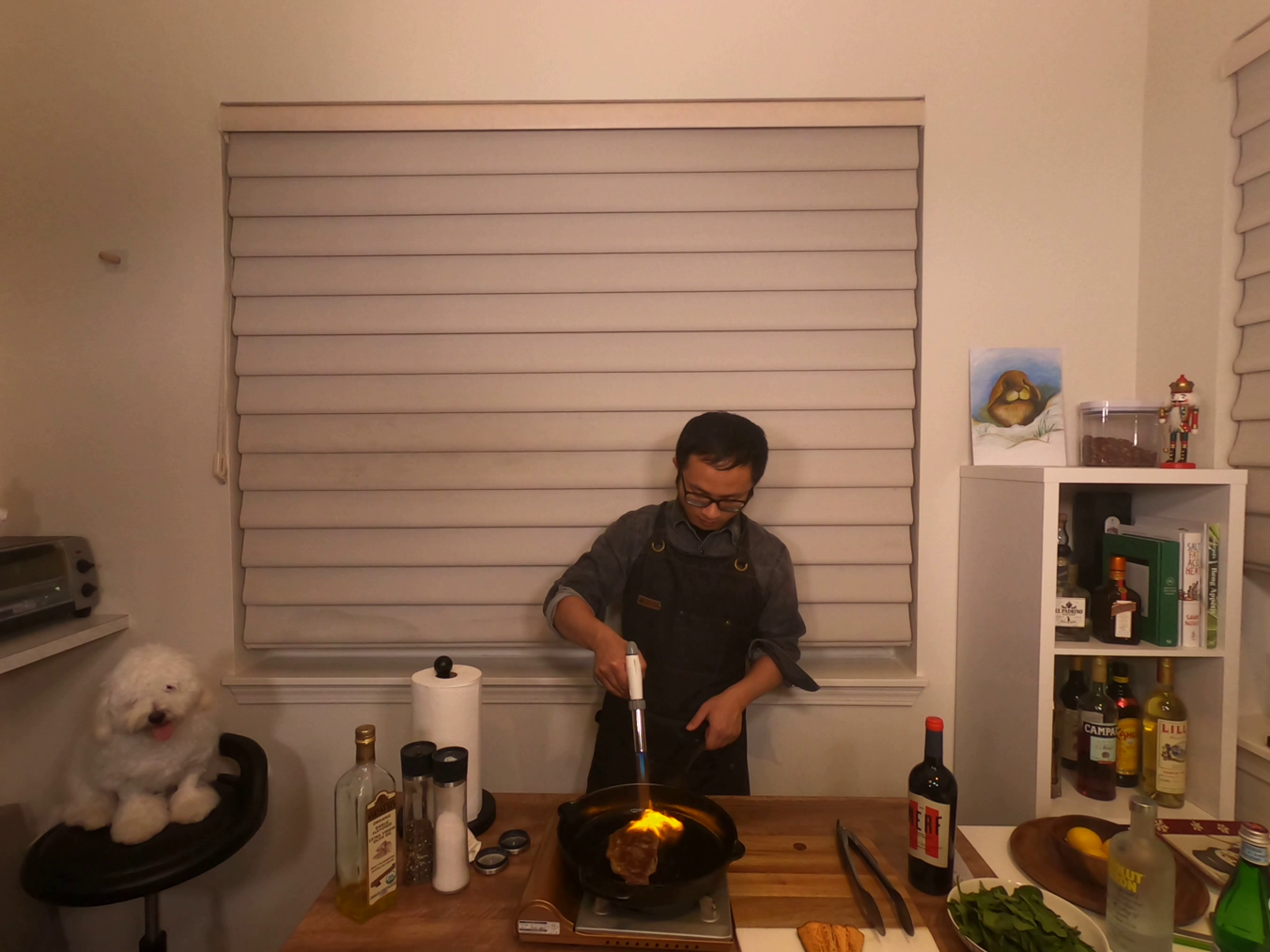}{2.5}{0.13}{0.23}{3.5}{0.5}{0.14} 
    \\ [-13.5mm]
    \videozoomed{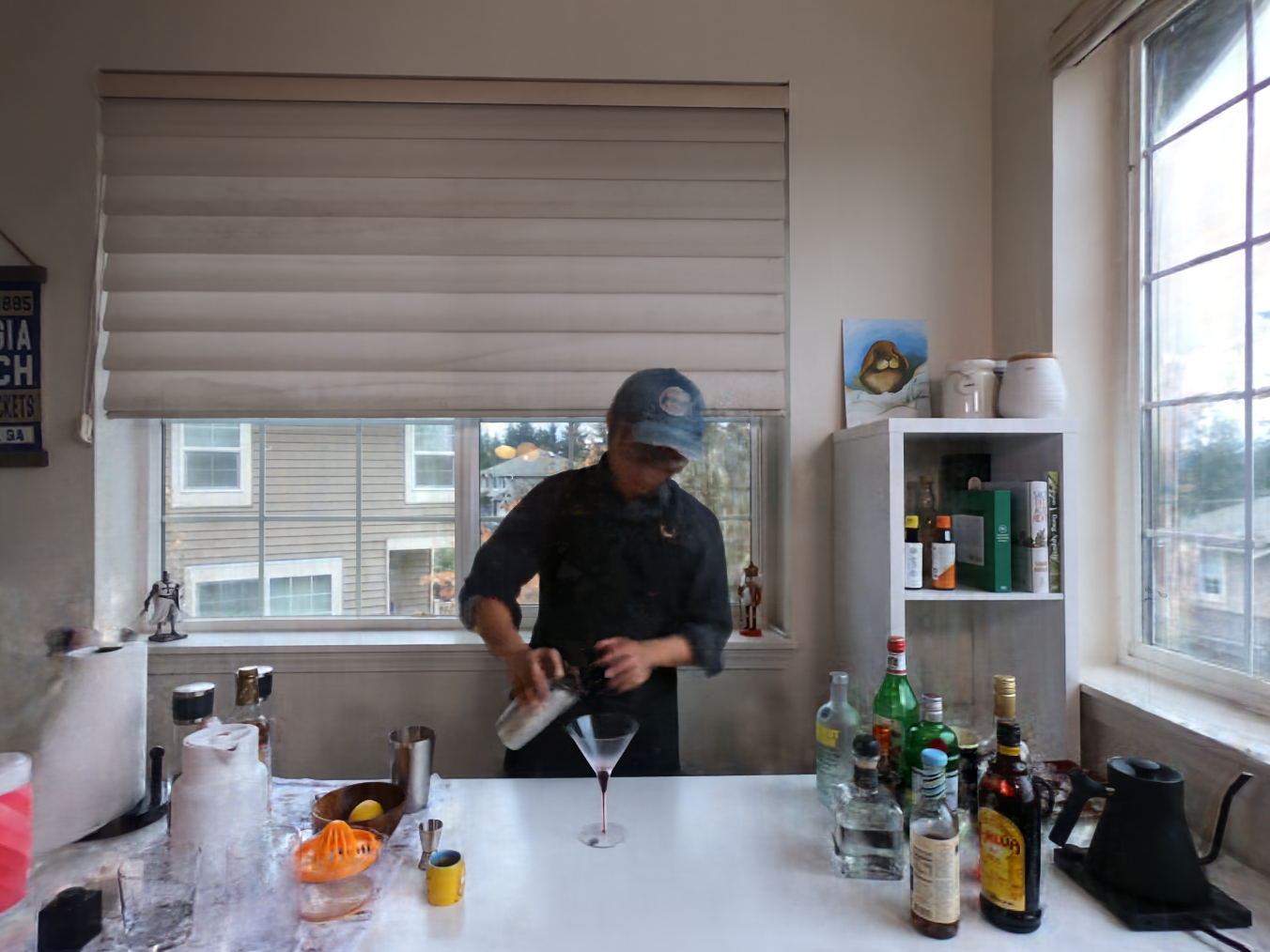}{4}{0.415}{0.312}{3.6}{0.691}{0.17} &
    \videozoomed{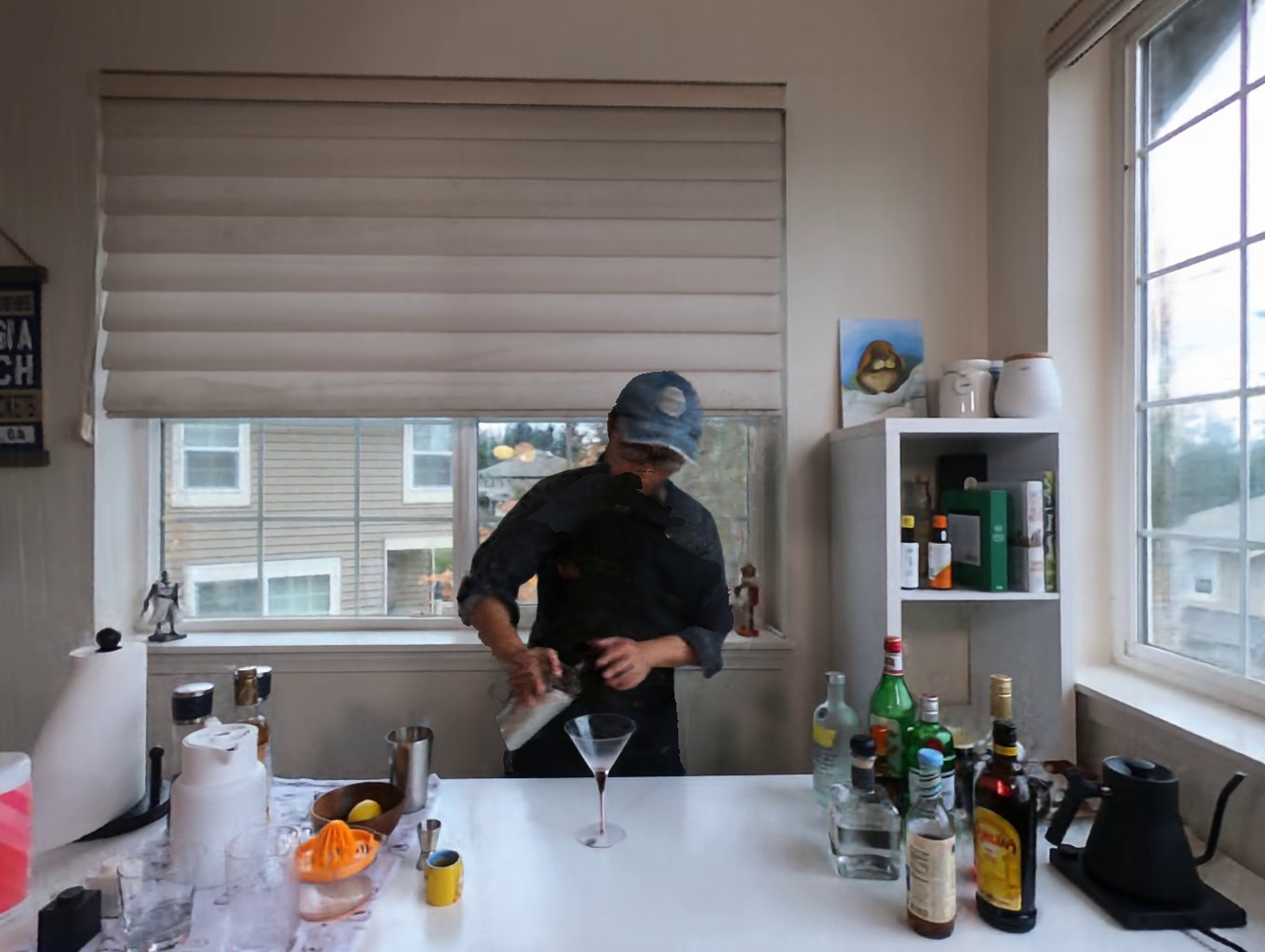}{4}{0.415}{0.312}{3.6}{0.691}{0.17} &
    \videozoomed{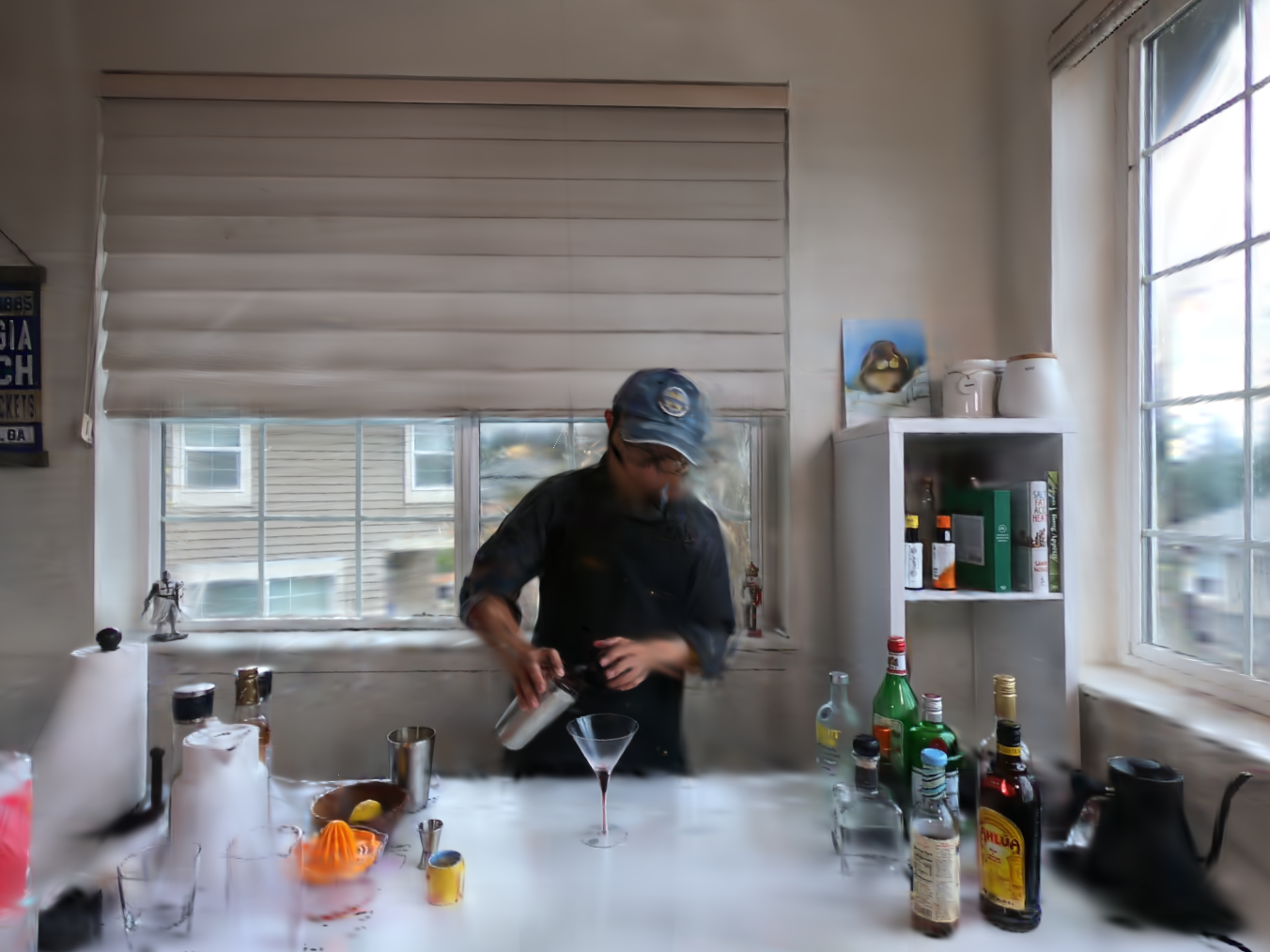}{4}{0.415}{0.312}{3.6}{0.691}{0.17} &
    \videozoomed{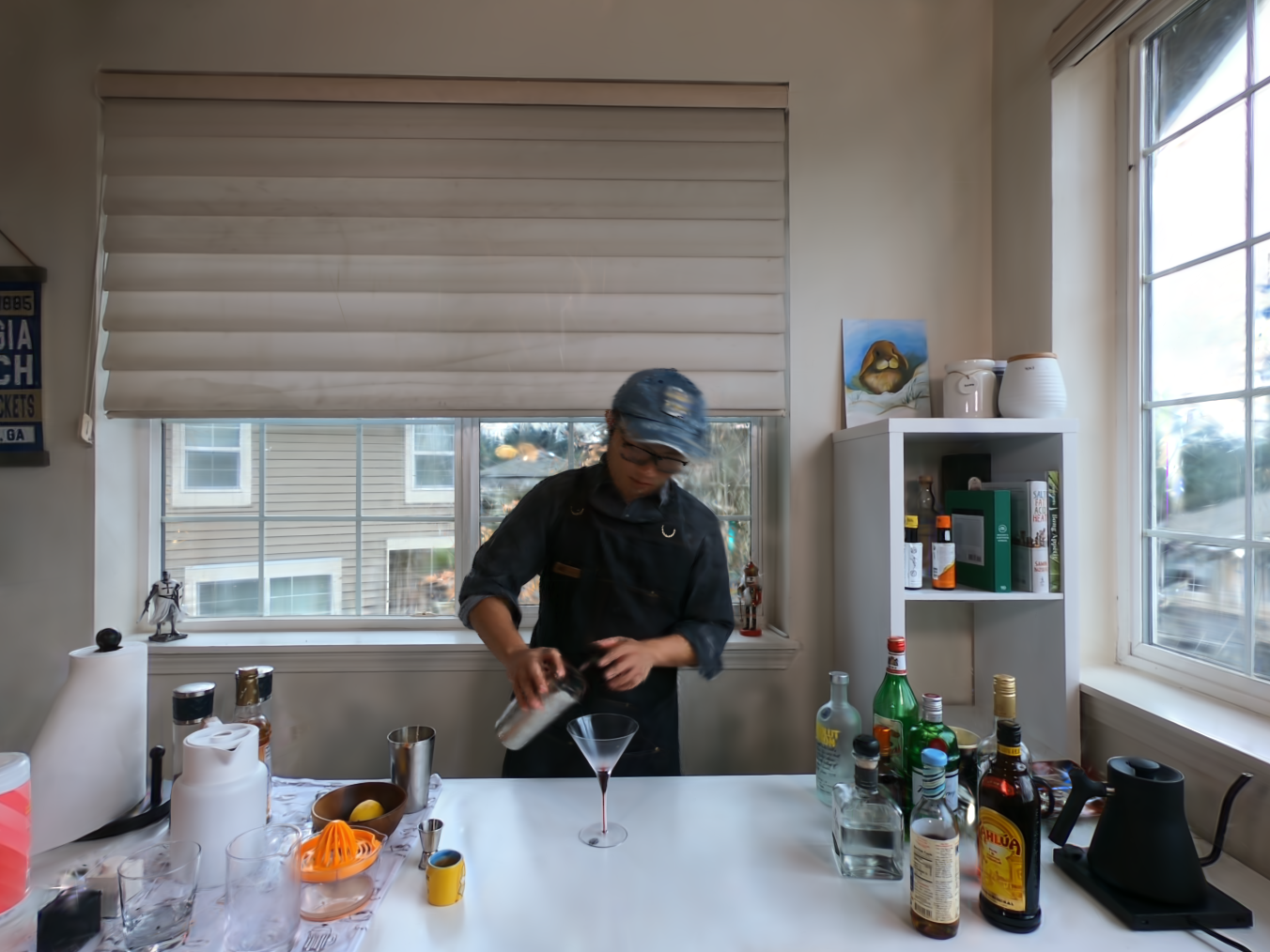}{4}{0.415}{0.312}{3.6}{0.691}{0.17} &
    \videozoomed{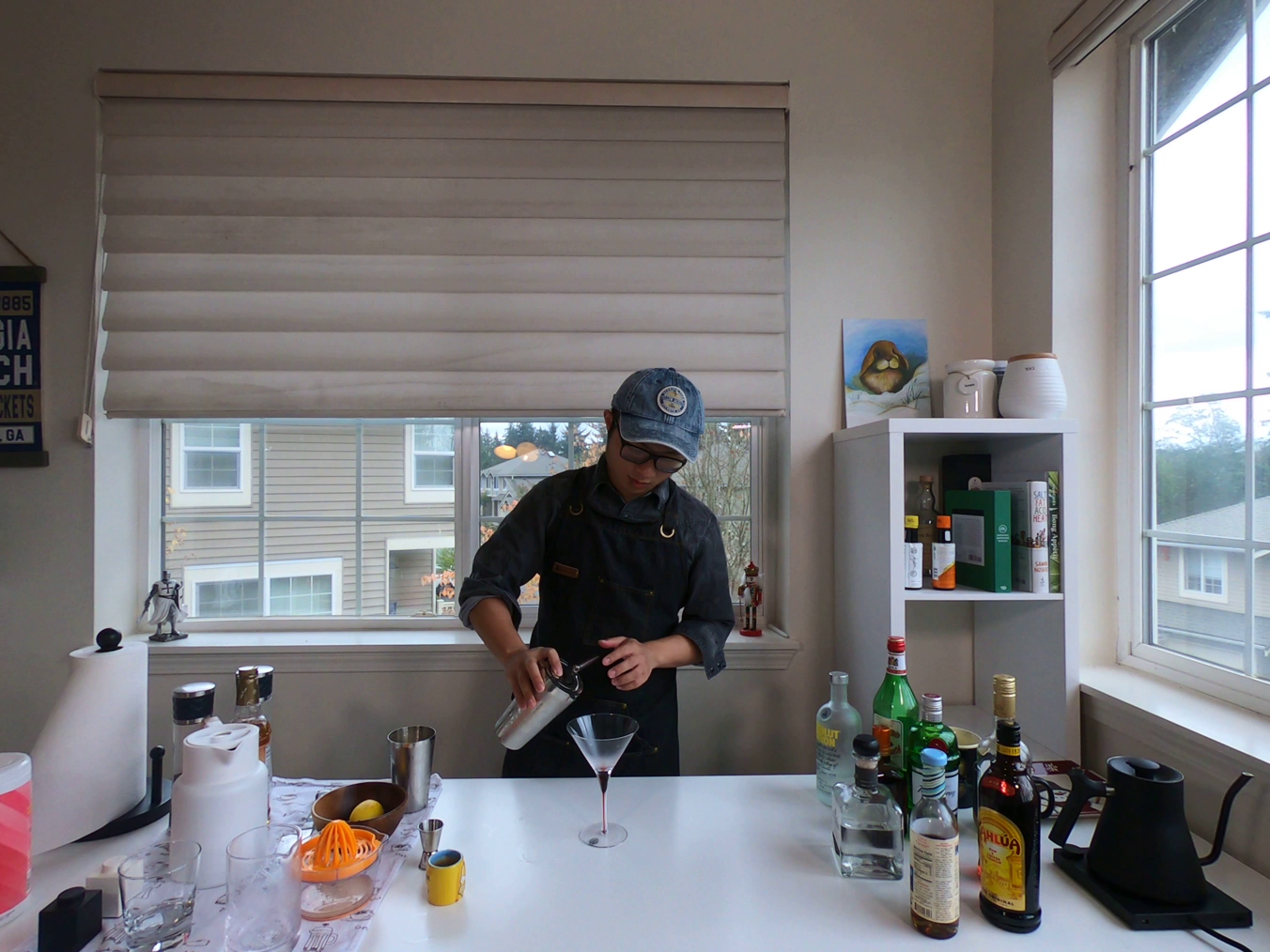}{4}{0.415}{0.312}{3.6}{0.691}{0.17} 
    \\ [-13.5mm]
    \videozoomed{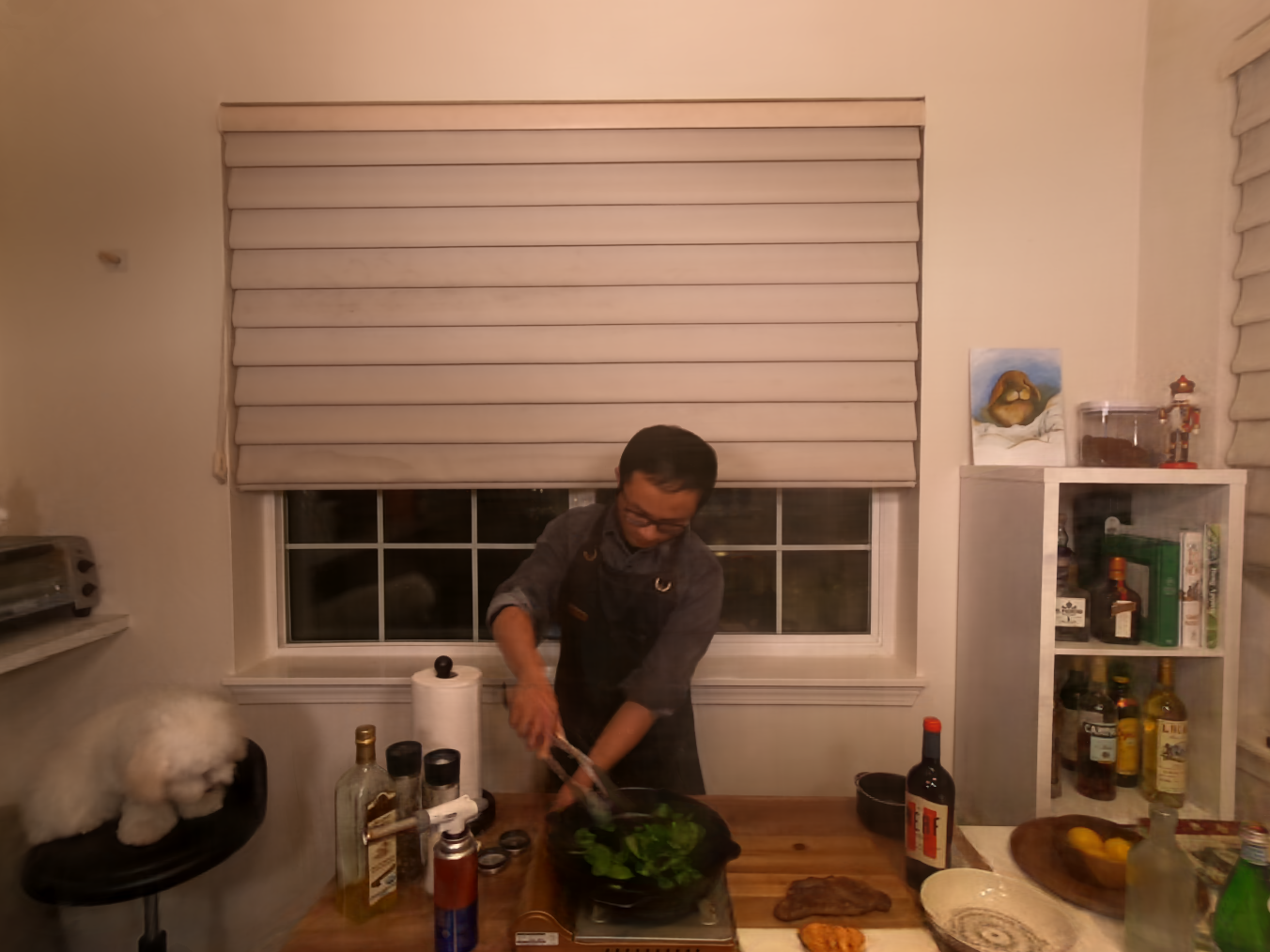}{5.5}{0.091}{0.726}{4}{0.524}{0.487} &
    \videozoomed{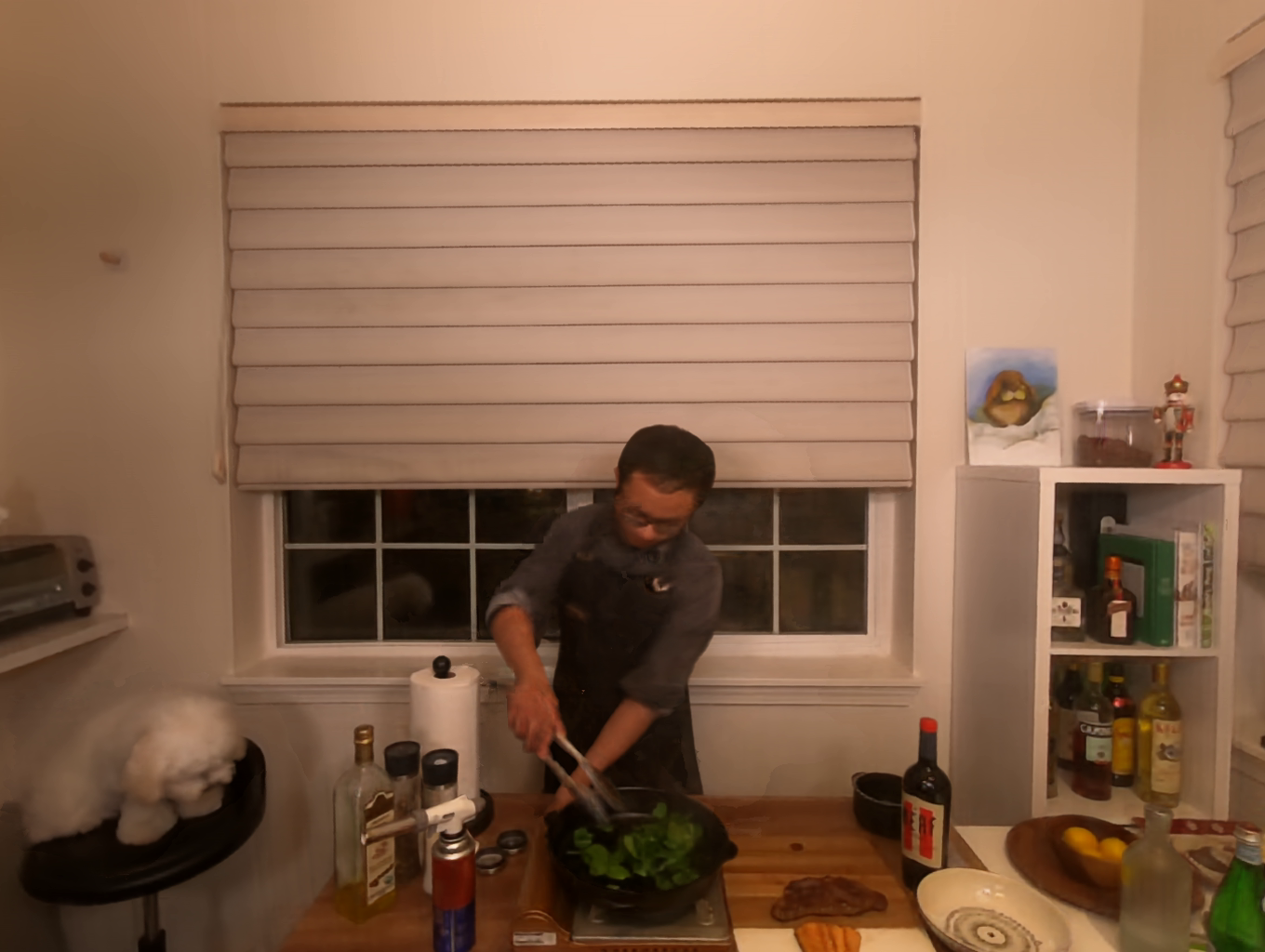}{5.5}{0.091}{0.726}{4}{0.524}{0.487} & 
    \videozoomed{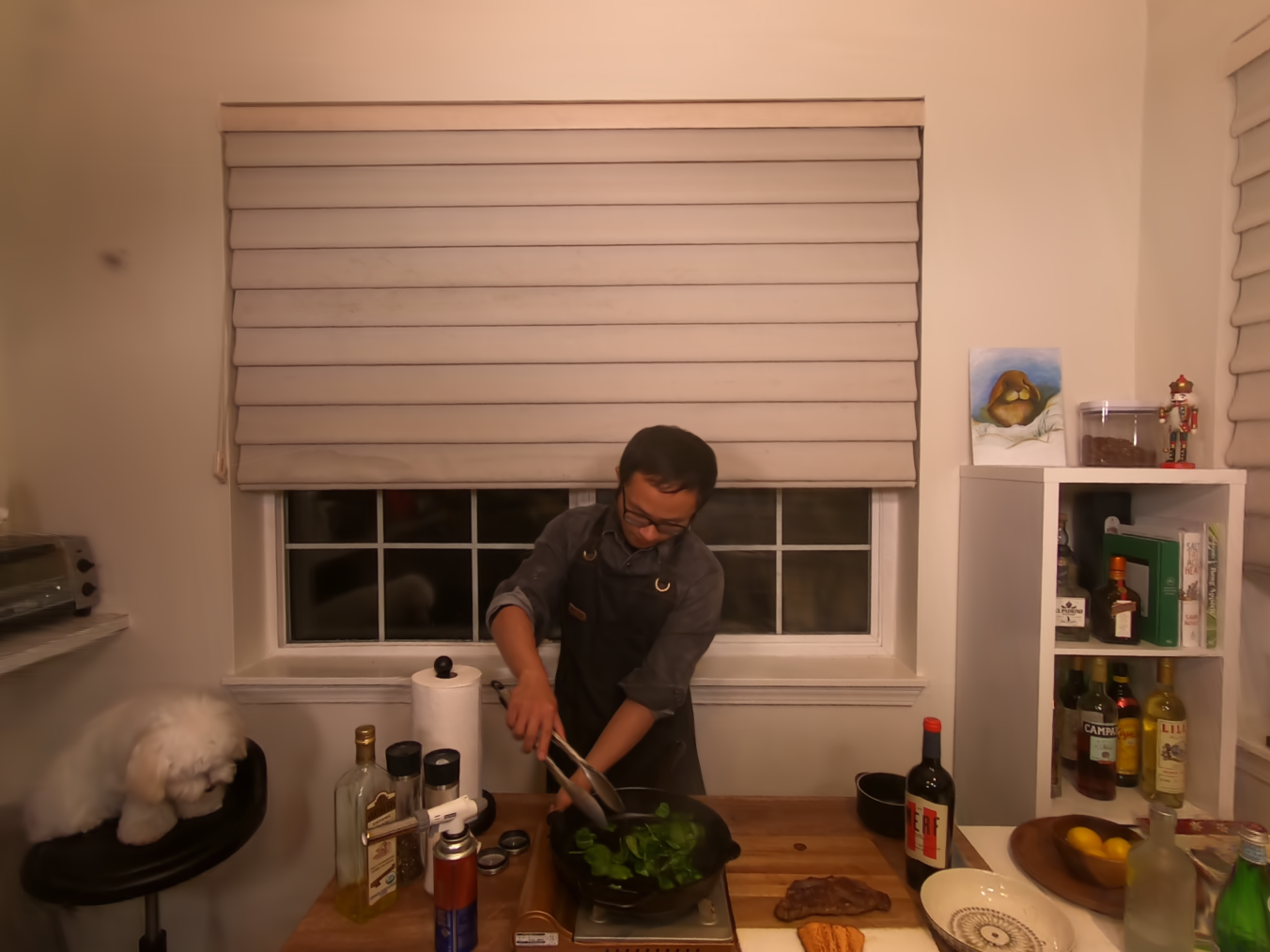}{5.5}{0.091}{0.726}{4}{0.524}{0.487} &
    \videozoomed{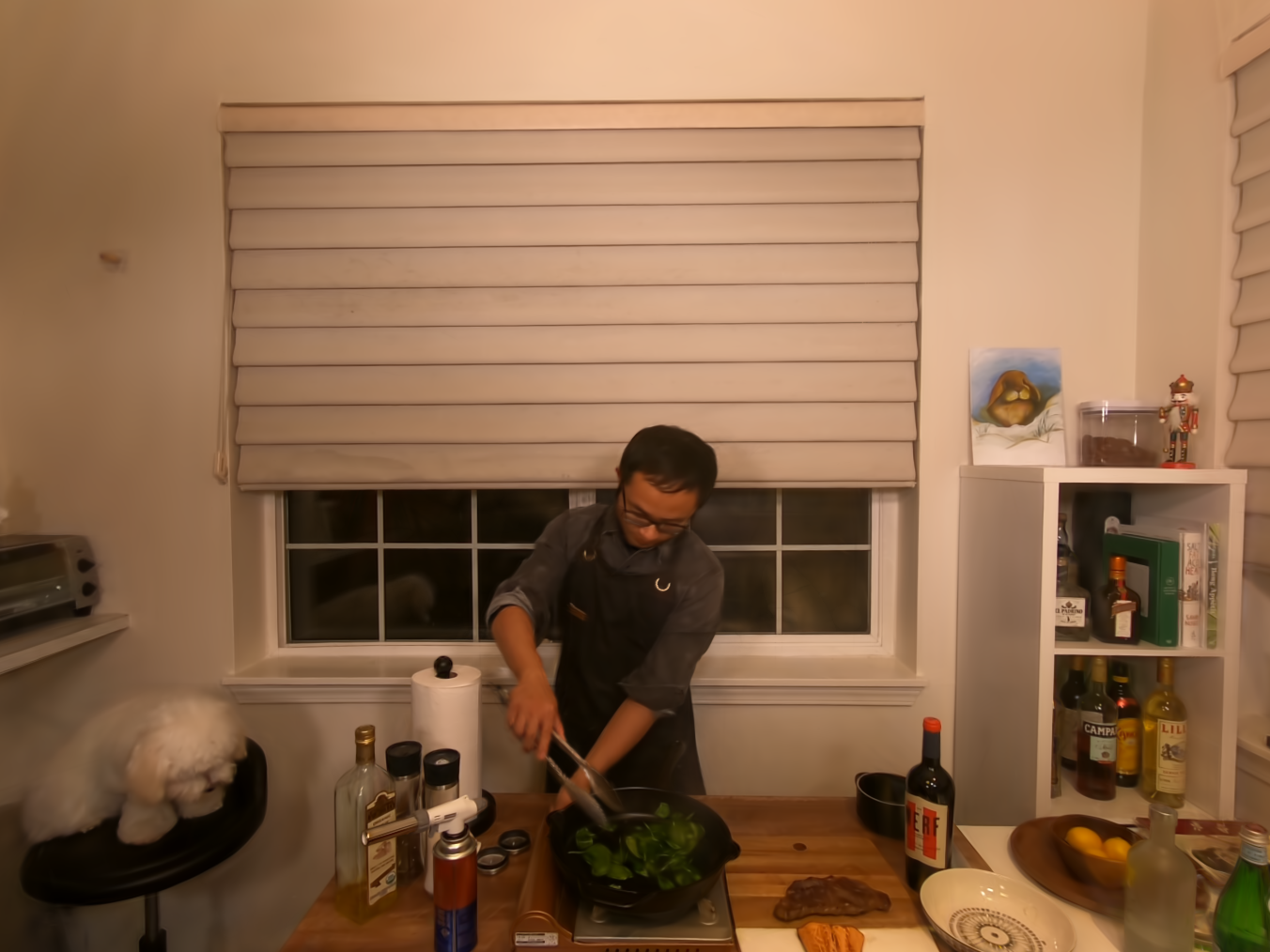}{5.5}{0.091}{0.726}{4}{0.524}{0.487} &
    \videozoomed{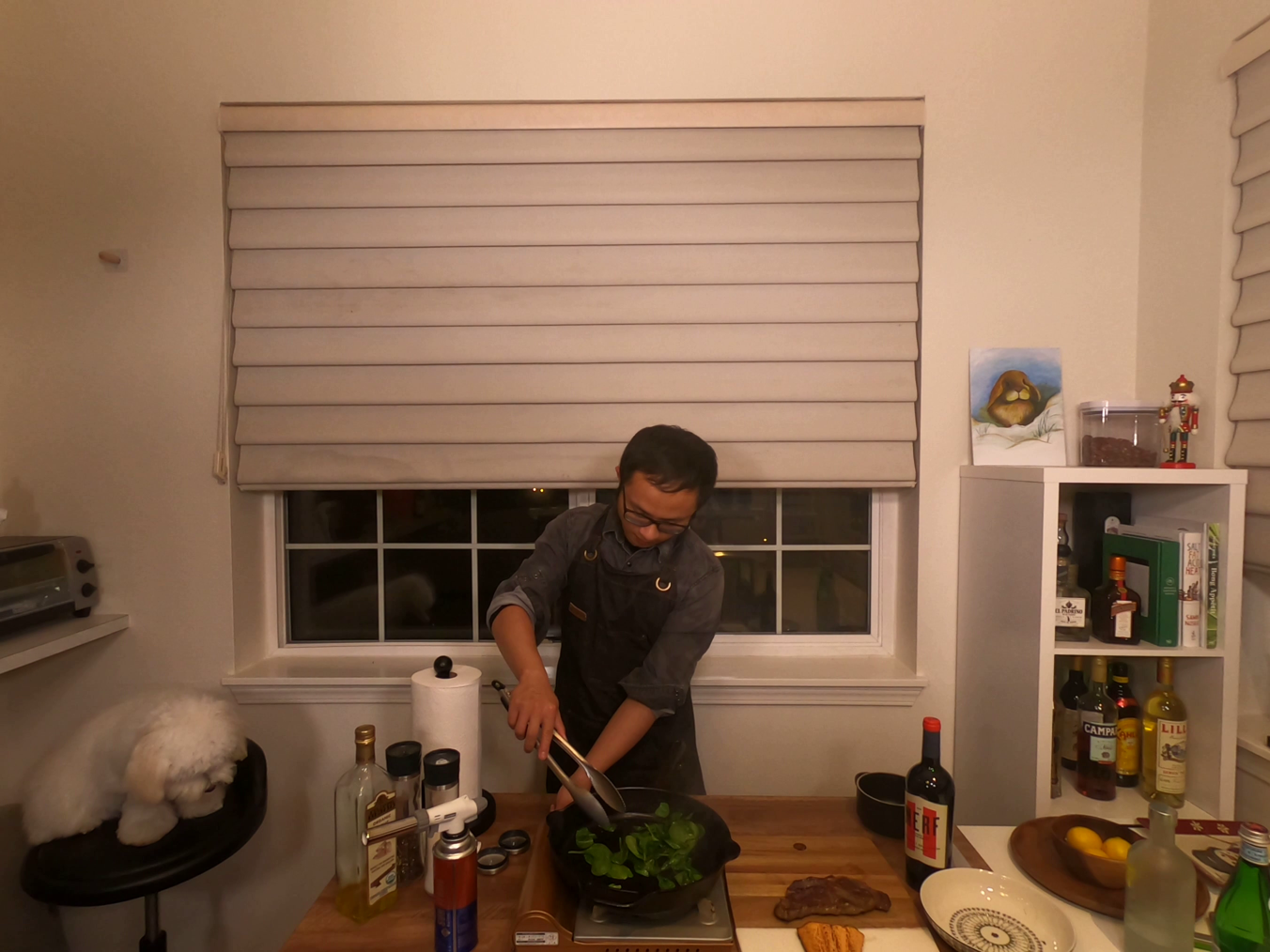}{5.5}{0.091}{0.726}{4}{0.524}{0.487}
    \\ [-13.5mm]
    \videozoomed{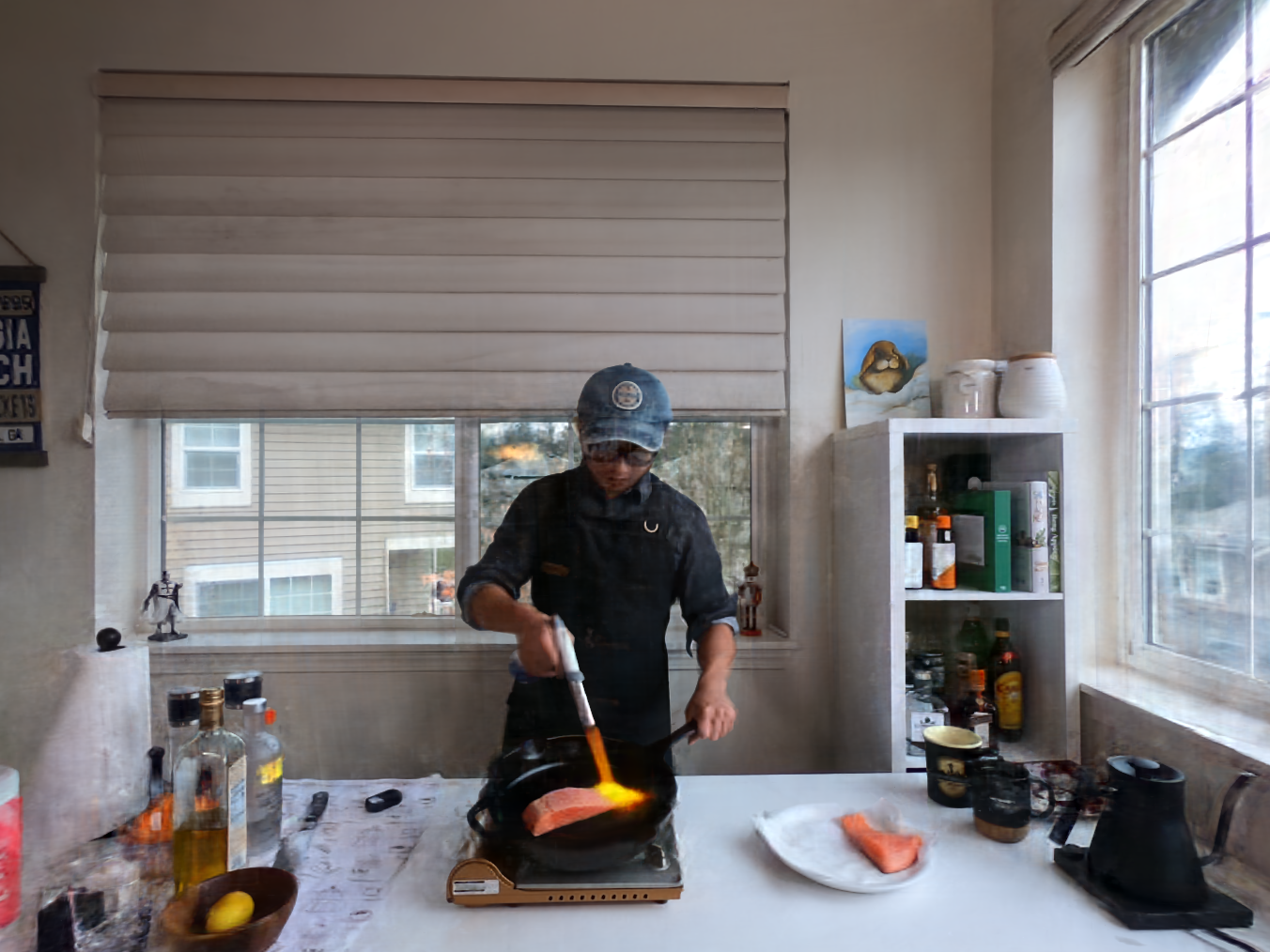}{3}{0.2}{0.100}{5}{0.483}{0.183} &
    \videozoomed{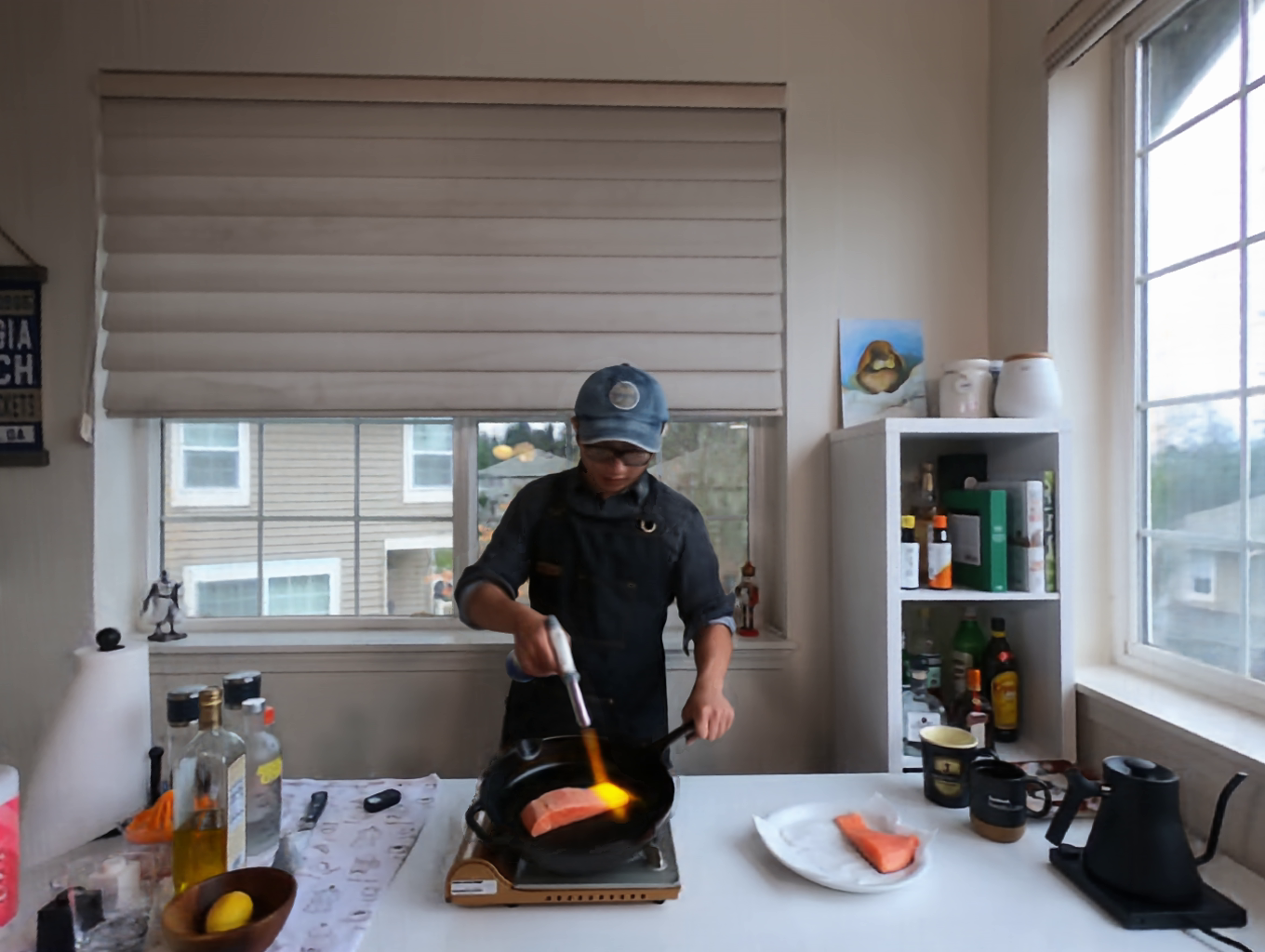}{3}{0.2}{0.100}{5}{0.483}{0.183} &
    \videozoomed{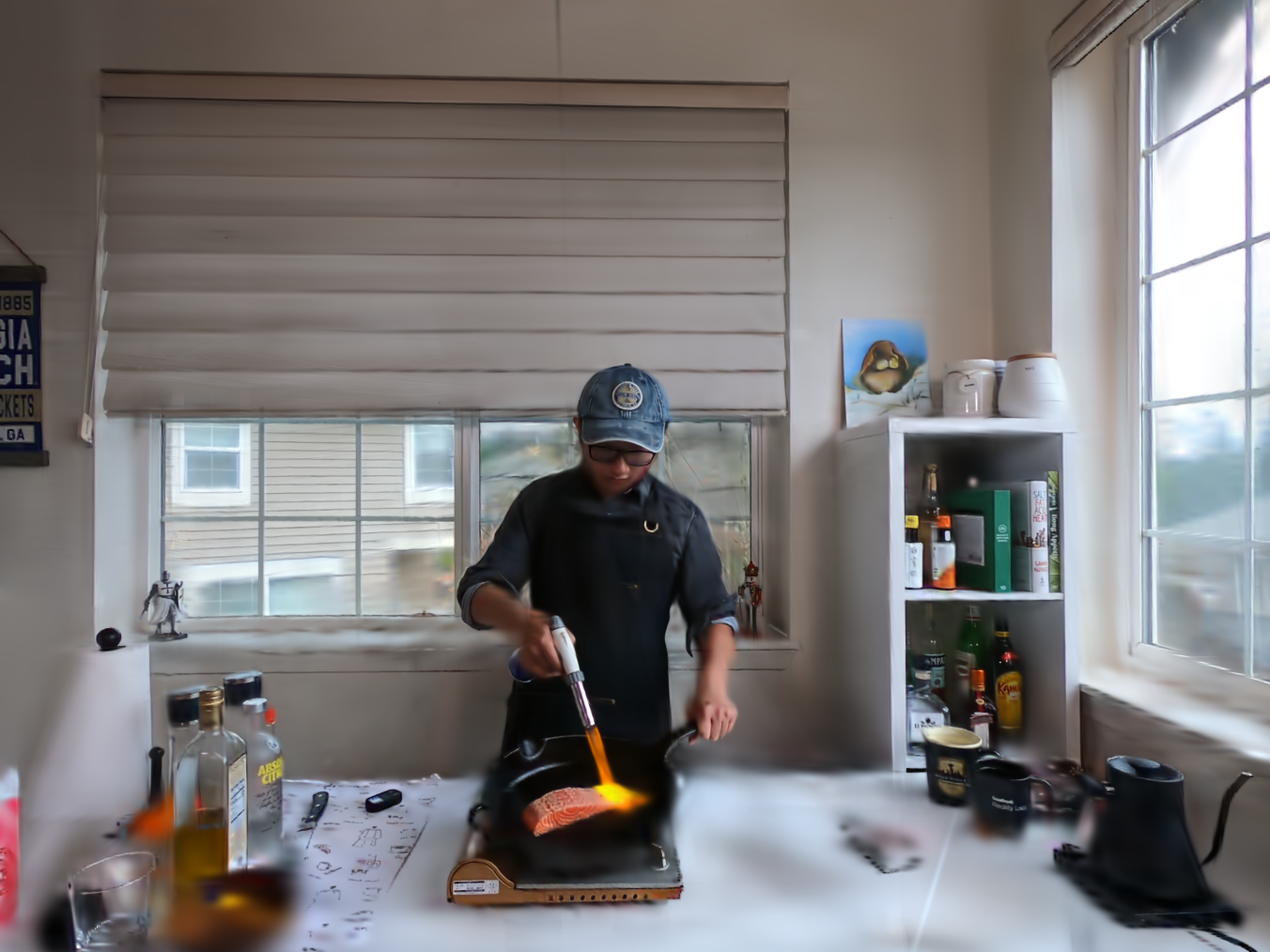}{3}{0.2}{0.100}{5}{0.483}{0.183} &
    \videozoomed{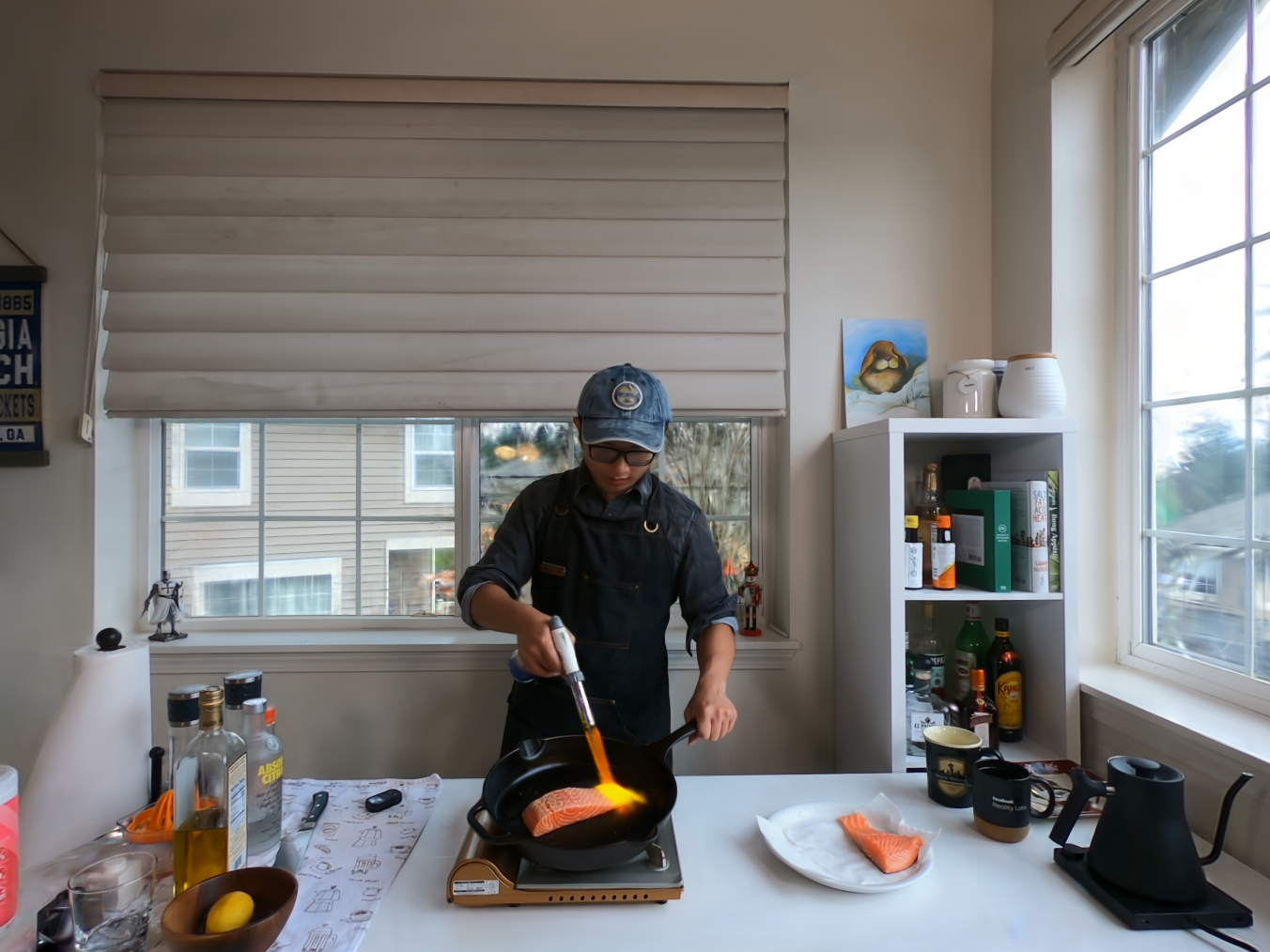}{3}{0.2}{0.100}{5}{0.483}{0.183} &
    \videozoomed{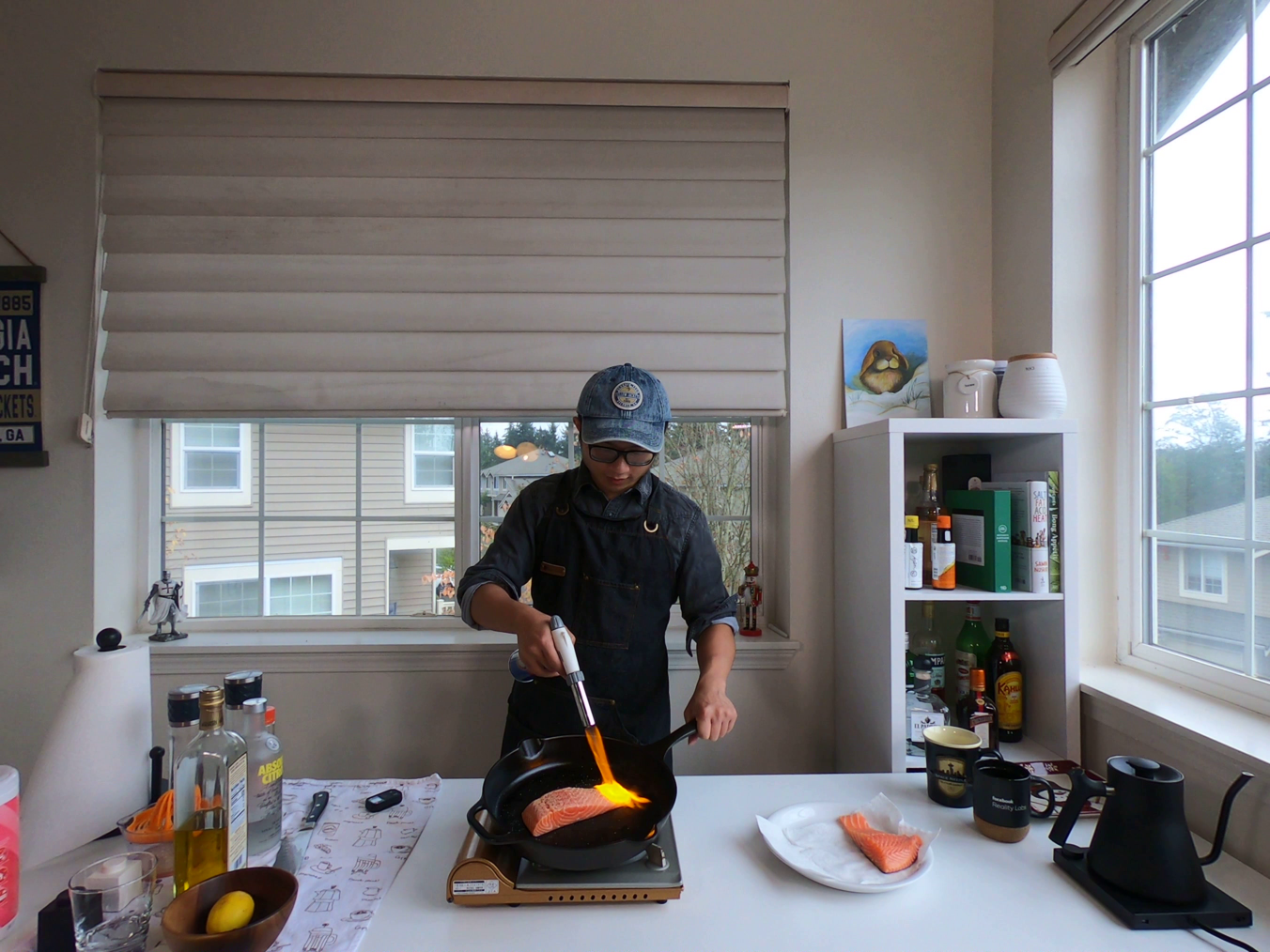}{3}{0.2}{0.100}{5}{0.483}{0.183}
    \\ [-13.5mm]
    \videozoomed{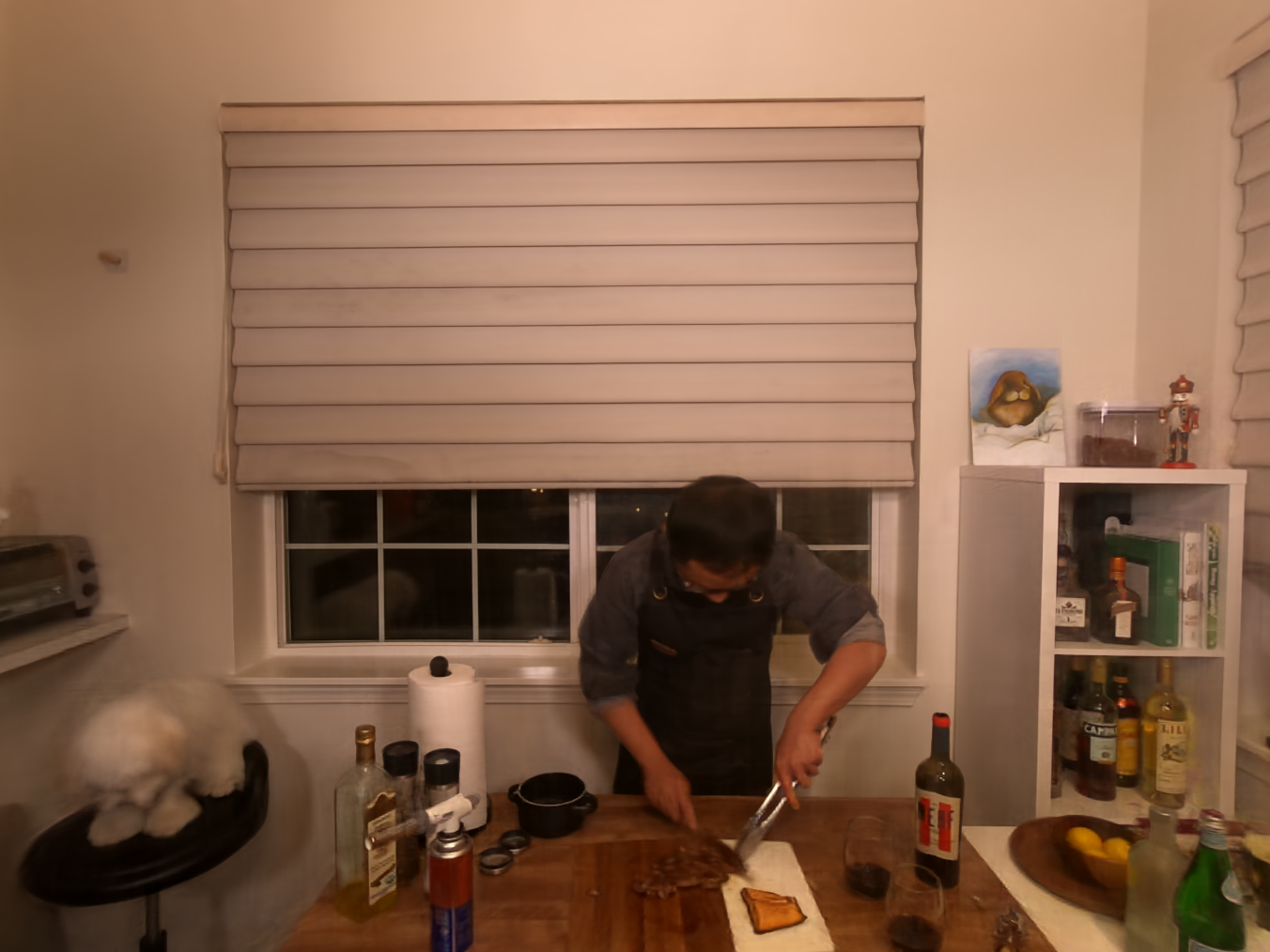}{3.8}{0.555}{0.14}{4}{0.92}{0.112}   &
    \videozoomed{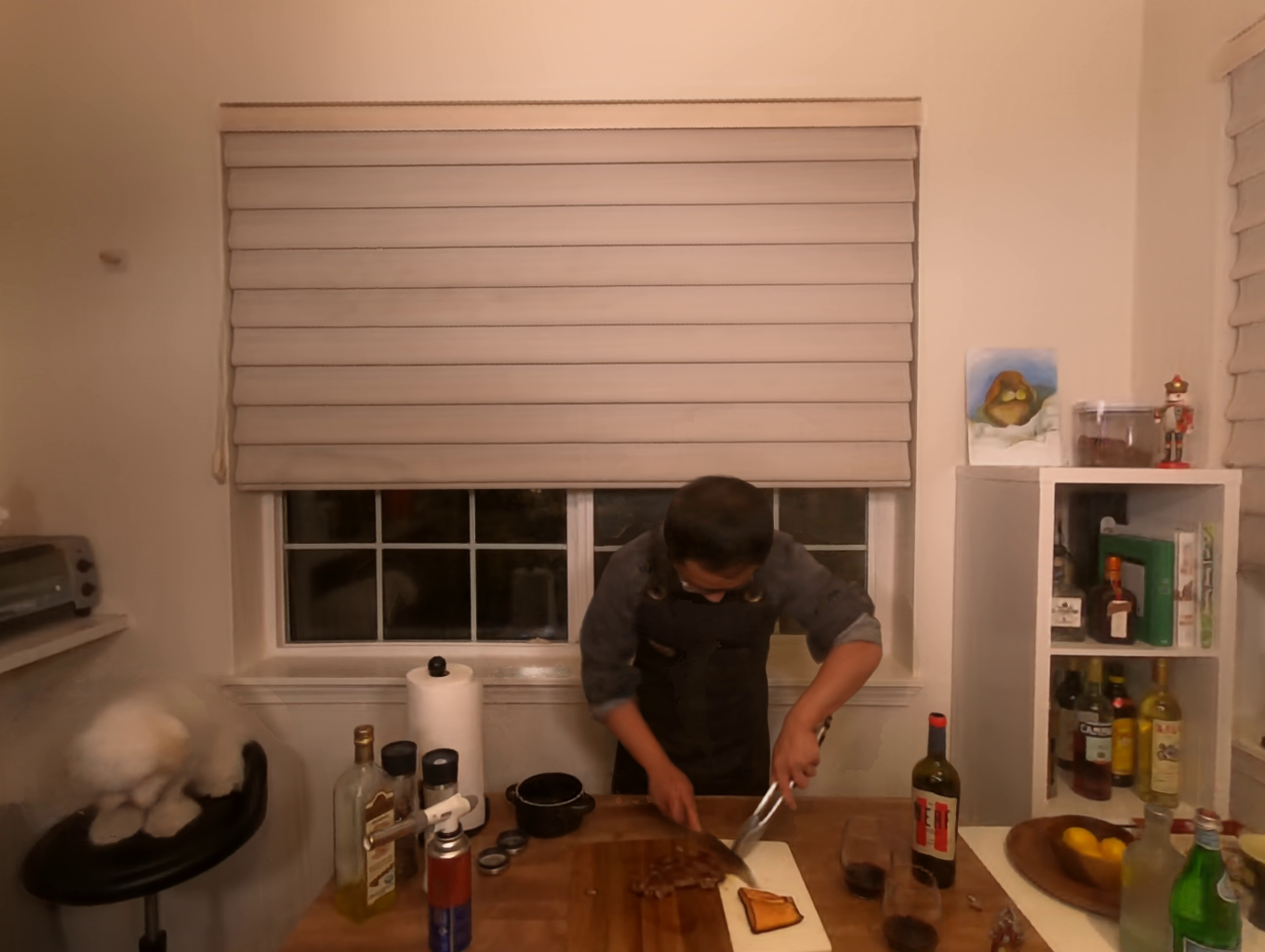}{3.5}{0.55}{0.14}{4}{0.92}{0.112}  &
    \videozoomed{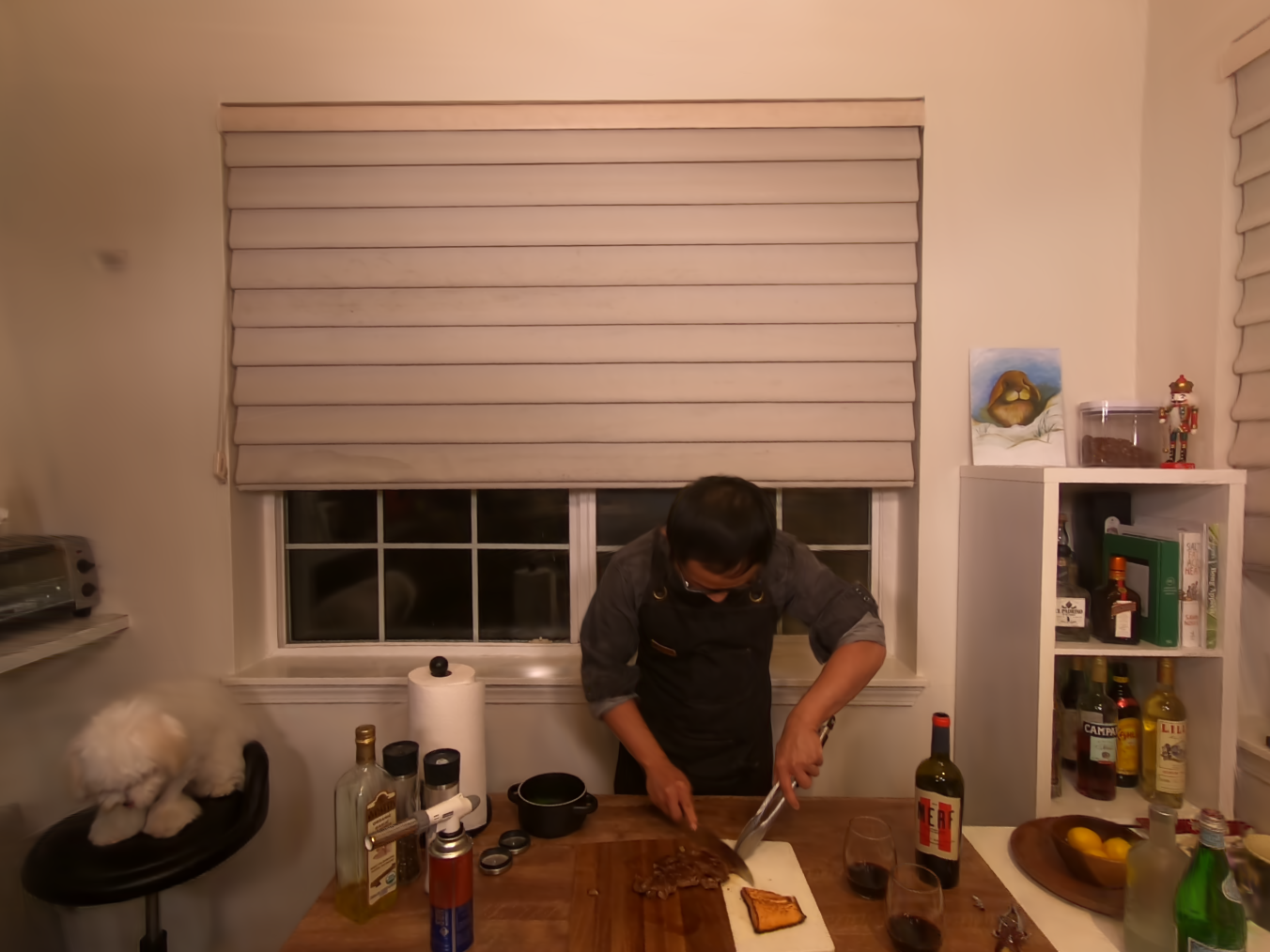}{3.5}{0.55}{0.14}{4}{0.92}{0.112}   &
    \videozoomed{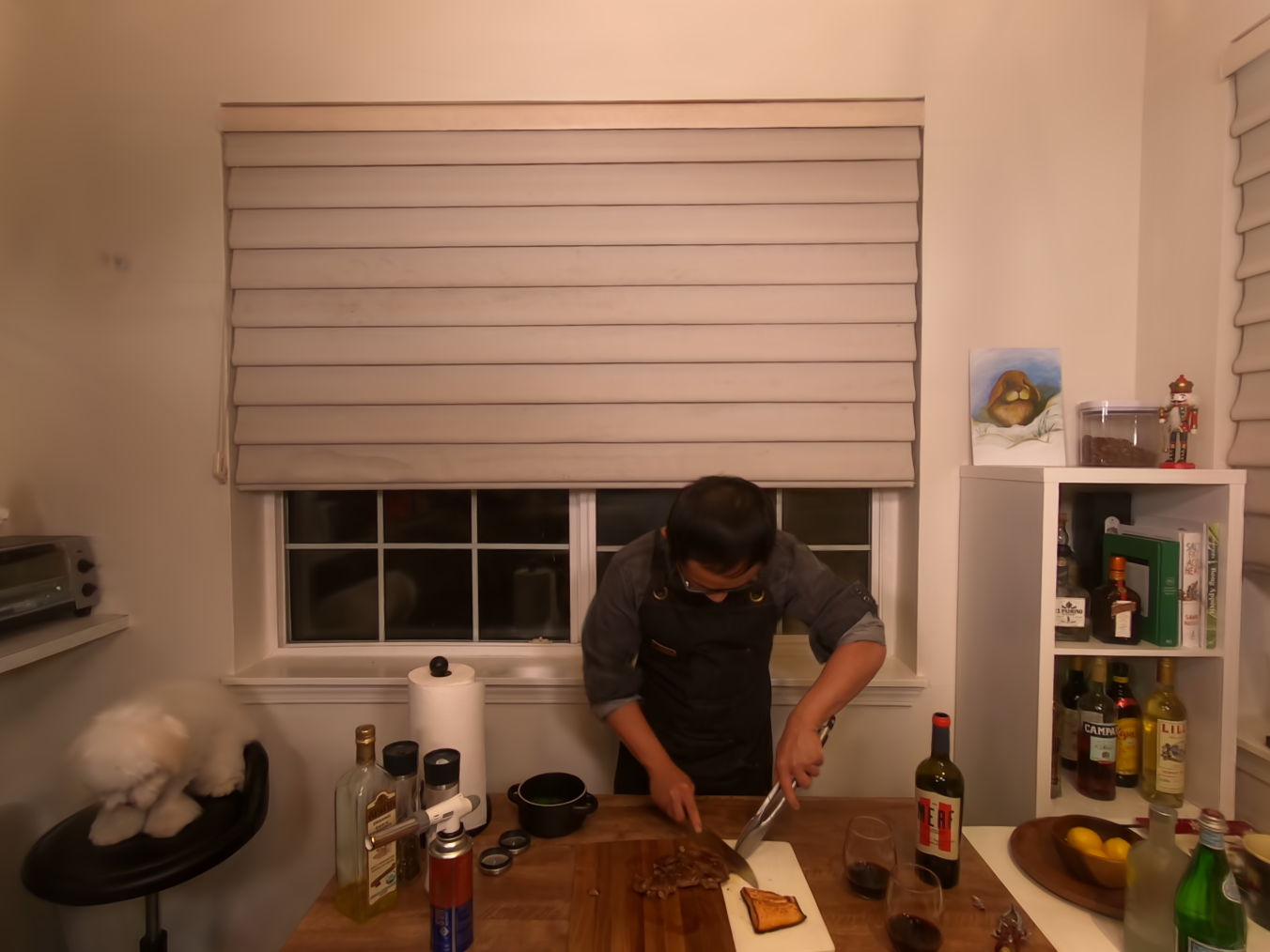}{3.5}{0.55}{0.14}{4}{0.92}{0.112}   &
    \videozoomed{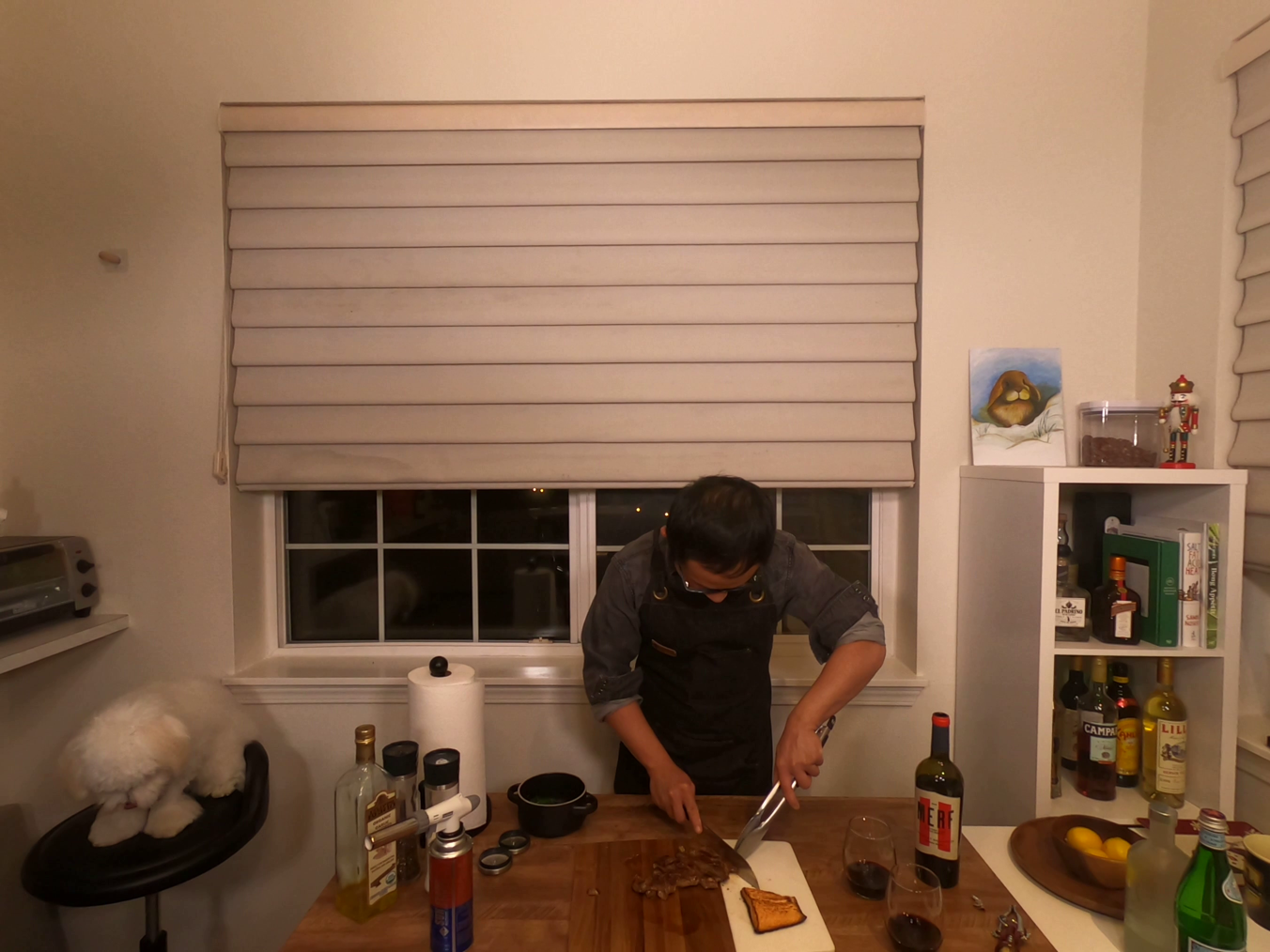}{3.5}{0.55}{0.14}{4}{0.92}{0.112} 
    \\ [-13.5mm]
    \multicolumn{1}{c}{\small HyperReel} &
    \multicolumn{1}{c}{\small MixVoxels} &
    \multicolumn{1}{c}{\small RealTime4DGS} &
    \multicolumn{1}{c}{\small Ours} &
    \multicolumn{1}{c}{\small Ground Truth}
    \end{tabular}
  
    \caption{\label{fig:plen} \textbf{Qualitative Comparison on Plenoptic Video Dataset}. Compared with prior work, our method recovers finer details of dynamic regions, \eg, the magnified human parts, and renders sharper static regions, \eg, the zoomed-in hook in row 3.}
\end{figure*}
\begin{table}[t]
\centering
\setlength{\tabcolsep}{1.5pt}
\caption{\textbf{Evaluation on Plenoptic Video Dataset}. We compare our method with both NeRF-based and Gaussian-based approaches. Our method significantly outperforms baselines on PSNR and inference speed. *: Only tested on the scene \textit{Flame Salmon}. **: Trained on 8 GPUs.~\textdagger: Results from paper.}
\resizebox{0.47\textwidth}{!}{
\begin{tabular}{cc|cccccc}
\specialrule{.15em}{.1em}{.1em}
ID&Method                               & PSNR$\uparrow$ & SSIM$\uparrow$ & LPIPS$\downarrow$ & Train$\downarrow$ & FPS$\uparrow$ \\ \hline
\textit{a}&DyNeRF~\cite{li2022neural}*\textdagger          &29.58 &- & 0.08 & 1344 h** &0.015 \\
\textit{b}&StreamRF~\cite{li2022streaming}      &28.16  &0.85  &0.31  &79 min  &8.50   \\ 
\textit{c}&HyperReel~\cite{attal2023hyperreel}  &30.36  &0.92  &0.17  &9 h  &2.00  \\ 
\textit{d}&NeRFPlayer~\cite{song2023nerfplayer}\textdagger &30.69  &-  &0.11  &6 h  &0.05   \\ 
\textit{e}&K-Planes~\cite{fridovich2023k}       &30.73  &0.93  &0.07  &190 min  &0.10   \\ 
\textit{f}&MixVoxels~\cite{wang2023mixed}       &30.85  &0.96  &0.21  &91 min  &16.70  \\ \hline

\textit{g}&Deformable4DGS~\cite{wu20234d}           &28.42  &0.92  &0.17  &72 min&39.93   \\ 
\textit{h}&RealTime4DGS~\cite{yang2023real}     &29.95  &0.92  &0.16  & 8 h & 72.80 \\ \hline

\textit{i}&Ours           & 31.62  & 0.94 & 0.14 & 60 min & 277.47  \\ 
\specialrule{.1em}{.05em}{.05em}
\end{tabular}
}
\label{table:Plenoptic}
\end{table}

\section{Experiments}
\label{sec:rationale}

\subsection{Datasets}

We evaluate our method on two commonly used datasets representative of various challenges in dynamic scene modeling. \textbf{Plenoptic Video Dataset}~\cite{li2022neural} contains real-world multi-view videos of 6 scenes. It includes abrupt motion as well as transparent and reflective materials. Following prior work~\cite{li2022neural}, we use resolution of 1352$\times$1014. 
\textbf{D-NeRF Dataset}~\cite{pumarola2021d} contains monocular videos for 8 synthetic scenes. We use resolution of 400$\times$400 following standard practice~\cite{pumarola2021d}.

\subsection{Implementation Details}

\paragraph{Initialization.} We uniformly sample 100,000 points in a 4D bounding box as Gaussian means. For Plenoptic dataset, we initialize with colored COLMAP~\cite{fisher2021colmap} reconstruction. 3D scales are set as the distance to the nearest neighbor. Rotors are initialized with $(1, 0, 0, 0, 0, 0, 0, 0)$ equivalent of static identity transformation.

\paragraph{Training.} Using Adam optimizer, we train for 20,000 steps with batch size 3 on D-NeRF dataset and 30,000 steps with batch size 2 on Plenoptic dataset. Densification gradient thresholds are $5e-5$ and $2e-4$ for D-NeRF and Plenoptic datasets, respectively. We set $\lambda_1=0.2, \lambda_2=0.01, \lambda_3=0.05$, and $K=8$, except that $\lambda_2=0$ for Plenoptic dataset since its videos contain a large number of transparent objects. Learning rates, densification, pruning, and opacity reset settings all follow~\cite{kerbl20233d}.

\paragraph{CUDA Acceleration.} We implemented the forward and backward functions of 4D rotors to 4D rotation matrices and 4D Gaussian slicing in our customized CUDA training framework. The duplication and pruning of the 4D Gaussians are also performed by CUDA, ensuring a low GPU memory usage.
\subsection{Baselines}
We compare with both NeRF-based and concurrent Gaussian-based methods on the two datasets. Most compared methods have released official codebase, in which case we run the code as is and report the obtained numbers for novel-view rendering quality, training time, and inference speed. Otherwise, we copy the results from their papers. All the numbers reported in the tables are benchmarked on an NVIDIA RTX 3090 GPU.

\subsection{Results}
\subsubsection{Evaluation on Plenoptic Video Dataset}
As detailed in Tab.~\ref{table:Plenoptic}, prior work struggles with trade-offs between rendering speed and quality, due to the slow volume rendering (\ita-\itf), time cost of querying neural network components (\itc, \itd, \itg), or the spatial-temporal entanglement (\ita, \itc, \itg).
Our method, however, exhibits significant advantages.
Foremost, it markedly outperforms existing work in rendering high-resolution videos (1352$\times$1014) at 277 FPS on an NVIDIA RTX 3090 GPU, over 10x faster than NeRF-based methods (\ita-\itf) and over 2x faster than Gaussian-based methods (\itg, \ith).
Moreover, our method achieves the highest PSNR of 31.62 (\vs the previous best 30.85) with a short average training time of 60 min.

As presented in Fig.~\ref{fig:plen}, our method promotes a clearer and more detailed reconstruction of dynamic regions over baselines. 
For all four scenes, the proposed approach reconstructs higher-quality human heads that move frequently and contain high-frequency details. 
As magnified in the first three scenes, baselines may produce blurry artifacts for the hand regions with fast motions.
In comparison, our method yields the sharpest renderings for the same regions.

\begin{figure}
	\centering
	\includegraphics[page=3,trim={2.0cm 20.3cm 11.5cm 2.7cm},clip]{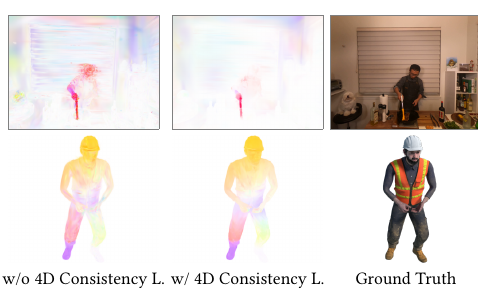}
\caption{\label{fig:flow} \textbf{Optical Flow Visualization.} Our method naturally derives a speed field. We compute 3D motions for all Gaussian points and visualize the rendered 2D optical flow. Adding 4D Consistency loss significantly reduces noises and improves the accuracy and smoothness of optical flow.}
\end{figure}

\new{Despite using a similar 4D Gaussian representation, we believe our improvements over~\cite{yang2023real} mainly come from the proposed regularization terms. We leverage the entropy loss to diminish floaters in challenging sparse-view reconstruction, evidenced in Fig.~\ref{fig:abl}, and introduce 4D Cosistency loss which demonstrates strong stabilization of the Gaussian motion distribution (Fig.~\ref{fig:flow})}.

\begin{table}[t]
\centering
\caption{\textbf{Evaluation on D-NeRF Dataset}. Our method outperforms NeRF-based and Gaussian-based baselines on both PSNR and rendering speed by a large margin. 
}
\setlength{\tabcolsep}{1.5pt}
\resizebox{0.47\textwidth}{!}{
\begin{tabular}{c|cccccc}
\specialrule{.15em}{.1em}{.1em}
Method                               & PSNR$\uparrow$ & SSIM$\uparrow$ & LPIPS$\downarrow$ & Train$\downarrow$ & FPS$\uparrow$  \\ \hline
D-NeRF~\cite{pumarola2021d}          &29.17  &0.95  &0.07  &24 h  &0.13 \\ 
TiNeuVox~\cite{fang2022fast}         &32.87 & 0.97 & 0.04 & 28 min &1.60 \\  
K-Planes~\cite{fridovich2023k}       &31.07  &0.97  &0.02  &54 min  &1.20   \\ \hline

Deformable3DGS~\cite{yang2023deformable} &39.31  &0.99  &0.01  &26 min  &85.45   \\ 
Deformable4DGS~\cite{wu20234d}           &32.99  &0.97  &0.05  &13 min  &104.00   \\ 
RealTime4DGS~\cite{yang2023real}    &32.71  &0.97  &0.03  &10 min  &289.07   \\ \hline

Ours                     &34.26  &0.97  & 0.03 & 5 min & 1257.63  \\ 
\specialrule{.1em}{.05em}{.05em}
\end{tabular}
}
\label{table:D-NeRF}
\end{table} 

\subsubsection{Evaluation on D-NeRF Dataset}
\begin{figure*}
	\centering
	\includegraphics[page=4,trim={2.0cm 7cm 2.0cm 2.8cm},clip]{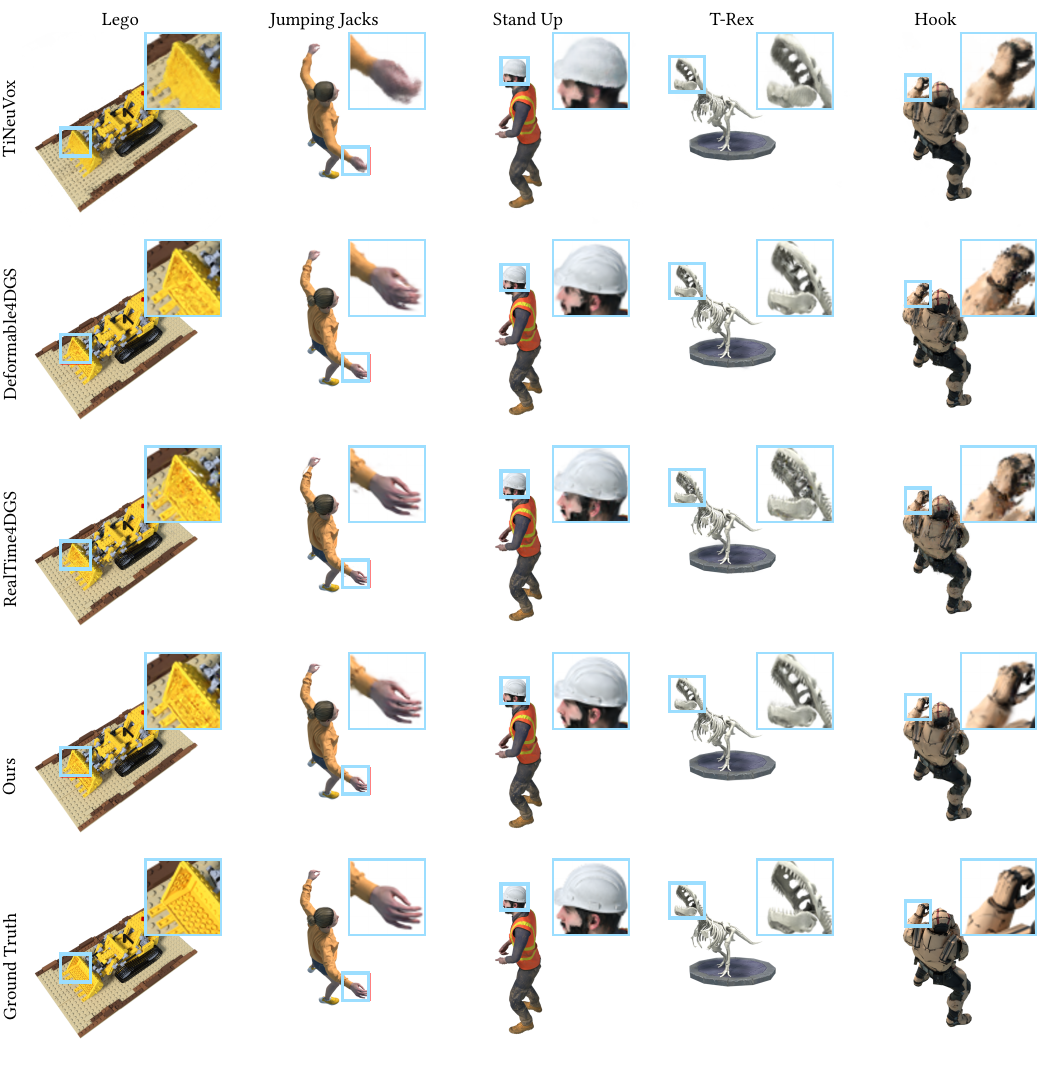}
	\caption{\label{fig:dnerf-white}  \textbf{Qualitative Comparison of Our Methods and Baselines on D-NeRF Dataset.} We compare with both NeRF-based (TiNeuVox~\cite{fang2022fast}) and Gaussian-based (Deformable4DGS~\cite{wu20234d} and RealTime4DGS~\cite{yang2023real}) methods. As revealed in the highlighted regions, our method is able to reconstruct scenes with less noises and more details,~\eg, the blade of the \textit{Lego} bulldozer, the cuffs and hands in \textit{Jumping Jacks} and \textit{Hook}, the teeth of \textit{T-Rex}, and the helmet and facial details in \textit{Stand Up}.}
\end{figure*}

Monocular video NVS is particularly challenging due to sparse input views. 
\new{As summarized in Tab.~\ref{table:D-NeRF},~\cite{yang2023deformable} achieves the highest PSNR since it directly tracks 3D Gaussian deformation, which perfectly aligns with D-NeRF dataset but may falter with sudden appearances or disappearances. Otherwise, our method achieves the best rendering quality among all the other methods}, at a rendering speed of 1258 FPS (8x faster than the previous best). Moreover, the training only takes around 5 min in our fast implementation. 
Fig.~\ref{fig:dnerf-white} showcases how this work surpasses baselines in reducing floaters and enhancing reconstruction. For instance, the blade of the \textit{Lego} bulldozer is now more defined. In \textit{Jumping Jacks}, our method generates fingers with crisper shapes and eliminates artifacts as observed in baseline results,~\eg, floaters beside the cuff in row 3. In \textit{Stand Up}, the patterns on the helmet and facial features are more pronounced in our results. The missing teeth details in baseline results are recovered in ours. Additionally, noises in \textit{Hook}'s hand are mitigated by the proposed method, resulting in clearer fingers.

\begin{table}
\centering
\caption{\textbf{Ablation on D-NeRF Dataset}. We validate three designs on rendering quality and number of points (in millions): (\textit{b}) Entropy (with cross-entropy loss), (\textit{c}) KNN (with 4D KNN Consistency loss), and (\textit{Full}) Batch (with batch training).}
\setlength{\tabcolsep}{3pt}
\resizebox{0.9\linewidth}{!}{
\begin{tabular}{cccc|ccc}
\specialrule{.15em}{.1em}{.1em}
ID&\multicolumn{3}{c|}{Ablation Items}&\multicolumn{3}{c}{D-NeRF}\\
& Entropy    &KNN       &Batch        & PSNR$\uparrow$ & SSIM$\uparrow$ & \#Point(M)$\downarrow$  \\ \hline
\textit{a}          &          &&            &31.53  &0.96  &$1.7e5$\\ 
\textit{b}  &\checkmark &         &            &31.50  &0.97  &$5.4e4$  \\ 
\textit{c} &\checkmark &\checkmark&            &31.91  &0.97  &$3.0e5$  \\  \hline
\textit{Full}  &\checkmark &\checkmark&\checkmark&33.06  &0.98  & $1.4e5$ \\ \specialrule{.1em}{.05em}{.05em}
\end{tabular}
}
\label{table:ablation}
\end{table}

\subsection{Ablation Studies}
Tab.~\ref{table:ablation} ablates the effectiveness of individual designs in our method on the challenging D-NeRF dataset.

\begin{figure*}
	\centering
	\includegraphics[page=5,trim={2.0cm 6.7cm 2.0cm 2.8cm},clip]{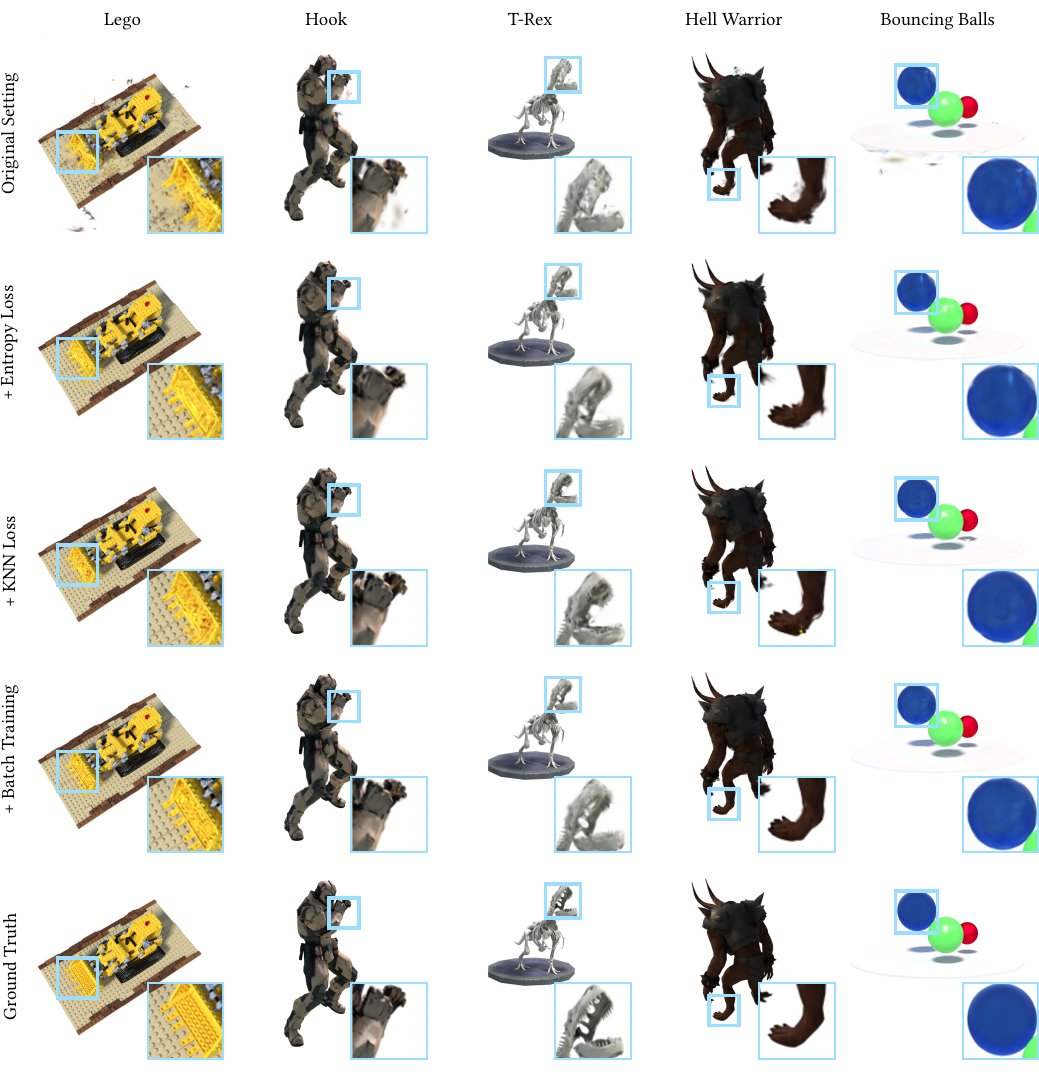}
	\caption{\label{fig:abl}  \textbf{Qualitative Comparison of Our Methods and Baselines on D-NeRF Dataset.} We compare with both NeRF-based (TiNeuVox~\cite{fang2022fast}) and Gaussian-based (Deformable4DGS~\cite{wu20234d} and RealTime4DGS~\cite{yang2023real}) methods. As revealed in the highlighted regions, our method is able to reconstruct scenes with less noises and more details,~\eg, the blade of the \textit{Lego} bulldozer, the cuffs and hands in \textit{Jumping Jacks} and \textit{Hook}, the teeth of \textit{T-Rex}, and the helmet and facial details in \textit{Stand Up}.}
\end{figure*}

\paragraph{Entropy Loss.} As shown in Tab.~\ref{table:ablation} (\itb), adding entropy loss significantly reduces the number of points by an order of magnitude while maintaining the overall rendering quality as measured in PSNR and SSIM. The effect of entropy loss is clearly revealed in Fig.~\ref{fig:abl}. For example, the floaters around the scene \textit{Lego}, \textit{Hook}, and \textit{Bouncing Balls} in row 1 have been completely removed in row 2. This demonstrates that the entropy loss helps impose strong constraints on the 4D Gaussian point distribution during optimization. 
However, we also find that it results in PSNR degradation in Plenoptic dataset. We believe this is because Plenoptic dataset provides dense views and contains a lot of transparent objects. Therefore, we recommend the addition of entropy loss for opaque surfaces and sparse views.

\paragraph{4D Consistency Loss.} 
Originally, the states of neighboring Gaussians in 4D space can change freely, which increases the difficulty of optimization and the redundancy of the model. However, the application of 4D Consistency loss enforces local consistency across both spatial and temporal dimensions. This is confirmed in Fig.~\ref{fig:abl}, Tab.~\ref{table:ablation} (\itc) and Fig.~\ref{fig:flow} where 4D Consistency loss helps recover consistent motions, add more details, and improve rendering quality.

\paragraph{Batch Training.} Batch training helps reduce the gradient noise and stabilize training. Tab.~\ref{table:ablation} (\textit{Full}) shows that batch training further improves the rendering quality over setting (\itc). For sparse view settings such as monocular videos, batch training also helps improve the geometry consistency by jointly optimizing over multiple views, as evidenced in Fig.~\ref{fig:abl},~\eg, feet in \textit{Hell Warrior}.

\section{Conclusion}
In this work, we propose 4D Gaussian Splatting, a novel approach that enables high-quality 4D dynamic scene modeling. Our method outperforms prior arts by a large margin and achieves an unprecedented 583 FPS rendering speed on an RTX 4090 GPU. Moreover, this is a unified framework for both 3D static and 4D dynamic reconstruction. 
While we have already achieved state-of-the-art reconstruction quality, we observe that, due to the increased dimensions, 4D Gaussians are hard to constrain and cause artifacts such as floaters and inconsistent motions. While entropy loss and 4D consistency loss help mitigate these issues, artifacts still exist. Future directions include exploiting 4D Gaussians for downstream tasks such as tracking and dynamic scene generation.

\begin{acks}
This work was supported in part by National Key R\&D Program of China 2022ZD0160801. We thank reviewers for their valuable feedback, Szymon Rusinkiewicz for helpful discussions, and CVRP Lab from National University of Singapore for computing support.
\end{acks}


\bibliographystyle{ACM-Reference-Format}
\bibliography{main}

\clearpage

\clearpage
\appendix

\section{Details of 4D Gaussian Splatting}

The 4D rotor $\rotorFour$ is constructed from a combination of 8 components based on a set of basis:
\begin{equation}
    \label{eq:rotor4} \rotorFour = \s + \bi_{01} \e_{01} + \bi_{02} \e_{02} + \bi_{03} \e_{03} + \bi_{12} \e_{12} + \bi_{13} \e_{13} + \bi_{23} \e_{23} + \p \e_{0123}, 
\end{equation}
where $\e_{ij}=\e_i \wedge \e_j$ represents the exterior product between 4D axis $\e_i (i\in\{0,1,2,3\})$ that defines the orthonormal basis in the 4D Euclidean space and corresponds to $x, y, z, t$ axes respectively. $\e_{0123}=\e_0 \wedge \e_1 \wedge\e_2 \wedge \e_3$ is the exterior product of all 4D axis $\e_i (i\in\{0,1,2,3\})$.

\subsection{4D Rotor Normalization}
To ensure that $\rotorFour$ represents a valid 4D rotation, it must be normalized with 
\begin{equation}
\label{eq:rotor4_norm} \rotorFour\rotorFour^\dagger=1,
\end{equation}
where $\rotorFour^\dagger$ is the conjugate of $\rotorFour$:
\begin{equation}
    \rotorFour^\dagger = \s - \bi_{01} \e_{01} - \bi_{02} \e_{02} - \bi_{03} \e_{03} - \bi_{12} \e_{12} - \bi_{13} \e_{13} - \bi_{23} \e_{23} + \p \e_{0123}.
\end{equation}

By integrating Eq.~\ref{eq:rotor4_norm}, we get
\begin{equation}
	\begin{array}{rl}
		\rotorFour\rotorFour^\dagger= 
		            & \left(-2\bi_{01}\bi_{23} + 2 \bi_{02} \bi_{13} - 2 \bi_{03} \bi_{12} + 2 \p
            \s\right) \mathbf{\e}_{0}\wedge \mathbf{\e}_{1}\wedge \mathbf{\e}_{2}\wedge \mathbf{\e}_{3} + \\
            & \left(\bi_{01}^2+\bi_{02}^2+\bi_{03}^2+\bi_{12}^2+\bi_{13}^2+\bi_{23}^2+\p^2+\s^2\right)                                                            \\
		=           & 1.
	\end{array}
\end{equation}

This leads to two conditions:
\begin{equation}
\left\{
\begin{array}{rl}
&\p \s-\bi_{01}\bi_{23} + \bi_{02} \bi_{13} - \bi_{03} \bi_{12}= 0, \\
&\bi_{01}^2+\bi_{02}^2+\bi_{03}^2+\bi_{12}^2+\bi_{13}^2+\bi_{23}^2+\p^2+s^2 = 1.
\end{array}
\right.
\end{equation}

We define:
\begin{equation}
    \varepsiloN = f(\rotorFour)  = \p \s - \bi_{01}\bi_{23} + \bi_{02} \bi_{13} - \bi_{03} \bi_{12} = \p \s - \DeltA,
\end{equation}
\begin{equation}
    \length^2 = \bi_{01}^2+\bi_{02}^2+\bi_{03}^2+\bi_{12}^2+\bi_{13}^2+\bi_{23}^2+\p^2+s^2.
\end{equation}

Regarding the first condition, our goal is to achieve \(\varepsiloN=f(\rotorFour)=0\). Near the zero point, when $f(\rotorFour)$ is conceptualized as a linear function of $\rotorFour$, the root can be approximated utilizing the first derivative. Thus, we assume 
\begin{equation}
    0 = f(\rotorFour+\deltA\nabla\varepsiloN), 
\end{equation}
where $\deltA$ is a small number, and $\nabla\varepsiloN$ is the gradient of $\varepsiloN=f(\rotorFour)$, which can be computed as:
\begin{equation}
    \left\{\begin{array}{rl}
        \frac{\partial \varepsiloN}{\partial \p}      & = \s       \\
        \frac{\partial \varepsiloN}{\partial \s}      & = \p       \\
        \frac{\partial \varepsiloN}{\partial \bi_{01}} & = -\bi_{23} \\
        \frac{\partial \varepsiloN}{\partial \bi_{23}} & = -\bi_{01} \\
        \frac{\partial \varepsiloN}{\partial \bi_{02}} & = \bi_{13}  \\
        \frac{\partial \varepsiloN}{\partial \bi_{13}} & = \bi_{02}  \\
        \frac{\partial \varepsiloN}{\partial \bi_{03}} & = -\bi_{12} \\
        \frac{\partial \varepsiloN}{\partial \bi_{12}} & = -\bi_{03} \\
    \end{array}\right.,
\end{equation}

resulting in
\begin{equation}
    f(\rotorFour+\deltA\nabla\varepsiloN) = (ps-\DeltA)(1+\deltA^2)+l^2\deltA.
\end{equation}

To compute \(\deltA\), it is evident that the solution's existence condition
\begin{equation}
    \length^4 \geqslant 4(\p\s-\DeltA)^2
\end{equation}
is inherently satisfied. Consequently, two solutions for \(\deltA\) are deduced:
\begin{equation}
    \deltA = \frac{-\length^2 \pm \sqrt{\length^4 - 4(\p\s-\DeltA)^2}}{2(\p\s-\DeltA)}.
\end{equation}
To determine the sign, let \(\length^2 = 1 + \chi, \p\s - \DeltA = \xi\). As \(\chi\rightarrow 0, \xi\rightarrow 0\), $\deltA$ must satisfy
\begin{equation}
	\deltA = \frac{\pm(1 + \chi - 2\xi^2) - 1 - \chi}{2\xi}\rightarrow 0,
\end{equation}
and the positive sign is taken, \(\deltA = -\xi\), which meets the requirement.
Therefore,
\begin{equation}
    \deltA = \frac{-\length^2 + \sqrt{\length^4 - 4(\p\s-\DeltA)^2}}{2(\p\s-\DeltA)},
\end{equation}
and applying \(\rotorFour := \rotorFour+\deltA\nabla\varepsiloN\) satisfies the first normalization condition.

Regarding the second condition, it suffices to calculate \(\length^2\) for \(\rotorFour+\deltA\nabla\varepsiloN\) post-update and then divide each component by \(\length\). As each component undergoes proportional scaling, the condition \(\p\s-\DeltA=0\) remains intact.

To summarize, within the 4D rotor normalization $\rotorNormFunc$, we first apply:
\begin{equation}
\rotorFour := \rotorFour+\deltA\nabla\varepsiloN, 
\end{equation}
where
\begin{equation}
\deltA = \dfrac{-\length^2 + \sqrt{\length^4 - 4\varepsiloN^2}}{2\varepsiloN},
\end{equation}
\begin{equation}
\varepsiloN = \p \s - \bi_{01}\bi_{23} - \bi_{02} \bi_{13} + \bi_{03} \bi_{12}.
\end{equation}
Then with the updated \(\rotorFour\), we calculate
\begin{equation}
\length^2 = \bi_{01}^2+\bi_{02}^2+\bi_{03}^2+\bi_{12}^2+\bi_{13}^2+\bi_{23}^2+\p^2+\s^2,
\end{equation}
and the final normalized coefficients are obtained as:
\begin{equation}
\rotorFour :=\rotorFour / \length.
\end{equation}
This results in a normalized 4D rotor $\rotorFour$ suitable for 4D rotation.

\subsection{4D Rotor to Rotation Matrix Transformation}
After normalization, we map a source 4D vector $\uSrcFourD$ to a target vector $\uTargetFourD$ via
\begin{equation}
\label{eq:rotor4-rotation} \uTargetFourD = \rotorFour\uSrcFourD\rotorFour^\dagger,
\end{equation}
where such mapping can also be written in  4D rotation matrix $\fourDRotation$ form
\begin{equation}
\uTargetFourD = \fourDRotation \uSrcFourD = \rotorConvertFunc(r)\uSrcFourD = 
\left(\begin{matrix}
			\fourDRotation^{00} & \fourDRotation^{01} & \fourDRotation^{02} & \fourDRotation^{03} \\
			\fourDRotation^{10} & \fourDRotation^{11} & \fourDRotation^{12} & \fourDRotation^{13} \\
			\fourDRotation^{20} & \fourDRotation^{21} & \fourDRotation^{22} & \fourDRotation^{23} \\
			\fourDRotation^{30} & \fourDRotation^{31} & \fourDRotation^{32} & \fourDRotation^{33} \\
      \end{matrix}\right) \uSrcFourD,
\end{equation}
where
\begin{align}
&\fourDRotation^{00} = -\bi_{01}^2 - \bi_{02}^2 - \bi_{03}^2 + \bi_{12}^2 + \bi_{13}^2 + \bi_{23}^2 - \p^2 +  \s^2, \\
&\fourDRotation^{01} = 2 \left(\bi_{01} \s - \bi_{02} \bi_{12} - \bi_{03} \bi_{13} + \bi_{23} \p \right), \\
&\fourDRotation^{02} = 2 \left( \bi_{01} \bi_{12} + \bi_{02} \s - \bi_{03} \bi_{23} - \bi_{13} \p \right),\\
&\fourDRotation^{03} = 2 \left( \bi_{01} \bi_{13} + \bi_{02} \bi_{23} + \bi_{03} \s + \bi_{12} \p \right),\\
&\fourDRotation^{10} = 2 \left( -\bi_{01} s - \bi_{02} \bi_{12} - \bi_{03} \bi_{13} - \bi_{23} \p \right),\\
&\fourDRotation^{11} = -\bi_{01}^2 + \bi_{02}^2 + \bi_{03}^2 - \bi_{12}^2 - \bi_{13}^2 + \bi_{23}^2 - \p^2 +  \s^2,\\
&\fourDRotation^{12} = 2 \left( -\bi_{01} \bi_{02} + \bi_{03} \p + \bi_{12} \s - \bi_{13} \bi_{23} \right),\\
&\fourDRotation^{13} = 2 \left( -\bi_{01} \bi_{03} - \bi_{02} \p + \bi_{12} \bi_{23} + \bi_{13} \s \right), \\
&\fourDRotation^{20} = 2 \left( \bi_{01} \bi_{12} - \bi_{02} \s - \bi_{03} \bi_{23} + \bi_{13} \p \right),\\
&\fourDRotation^{21} = 2 \left( -\bi_{01} \bi_{02} - \bi_{03} \p - \bi_{12} \s - \bi_{13} \bi_{23} \right),\\
&\fourDRotation^{22} = \bi_{01}^2 - \bi_{02}^2 + \bi_{03}^2 - \bi_{12}^2 + \bi_{13}^2 - \bi_{23}^2 - \p^2 + \s^2,\\
&\fourDRotation^{23} = 2 \left( \bi_{01} \p - \bi_{02} \bi_{03} - \bi_{12} \bi_{13} + \bi_{23} \s \right),\\
&\fourDRotation^{30} = 2 \left( \bi_{01} \bi_{13} + \bi_{02} \bi_{23} - \bi_{03} \s - \bi_{12} \p \right),\\
&\fourDRotation^{31} = 2 \left( -\bi_{01} \bi_{03} + \bi_{02} \p + \bi_{12} \bi_{23} - \bi_{13} \s \right),\\
&\fourDRotation^{32} = 2 \left( -\bi_{01} \p - \bi_{02} \bi_{03} - \bi_{12} \bi_{13} - \bi_{23} \s \right),\\
&\fourDRotation^{33} = \bi_{01}^2 + \bi_{02}^2 - \bi_{03}^2 + \bi_{12}^2 - \bi_{13}^2 - \bi_{23}^2 - \p^2 +  \s^2.
\end{align}


\subsection{Temporally-Slicing 4D Guassians}
In this section, we provide the details about slicing the 4D Gaussian to 3D over time $\timeT$. That is, we calculate the 3D center position and 3D covariance after being intercepted by the $\timeT$ plane.

\textbf{Calculation of the 3D Center Position and 3D Covariance.}
First, we have the 4D covariance matrix represented by $\fourDCov$ and the rotation $\fourDRotation$
\begin{equation}
\fourDCov = \fourDRotation \fourDScaling \fourDScaling^T \fourDRotation^T.
\end{equation}
Then we get
\begin{equation}
	\fourDCov^{-1} = \fourDRotation \fourDScaling^{-1}\left(\fourDScaling^{-1}\right)^T \fourDRotation^T = \left(\begin{matrix}
			c_{xx} & c_{xy} & c_{xz} & c_{xt} \\
			c_{xy} & c_{yy} & c_{yz} & c_{yt} \\
			c_{xz} & c_{yz} & c_{zz} & c_{zt} \\
			c_{xt} & c_{yt} & c_{zt} & c_{tt} \\
		\end{matrix}\right).
	\label{eq:cov_4d_inv}
\end{equation}

Since a 4D Gaussian can be expressed as 
\begin{equation}
\fourDGaussian(\mathbf{x}) = e^{-\frac{1}{2} \mathbf{x}'^T \fourDCov^{-1} \mathbf{x}'} = e^{-\frac{1}{2}\B},
\label{eq:4d_gau}
\end{equation}
where $\mathbf{x}' = \mathbf{x} - (\centerPos_{x}, \centerPos_{y}, \centerPos_{z}, \centerPos_{t})$.
Then we get
\begin{align}
\B = &x'^2c_{xx} + y'^2c_{yy} + z'^2c_{zz} + 2x'y'c_{xy} + 2x'z'c_{xz} \nonumber \\
&+ 2y'z'c_{yz} + 2x't'c_{xt} + 2y't'c_{yt} + 2z't'c_{zt} + t'^2c_{tt}.
\end{align}

To obtain the 3D center position sliced by the $t$ plane, we set
\begin{align}
		\B = & c_{xx}(x'+\alpha t')^2 + c_{yy}(y'+\beta t')^2 + c_{zz}(z'+\gamma t')^2+ \nonumber\\
  & 2c_{xy}(x'+\alpha t')(y'+\beta t') + 2c_{yz}(y'+\beta t')(z'+\gamma t')+ \nonumber\\
  & 2c_{xz}(x'+\alpha t')(z+\gamma t')+ \nonumber\\
		    & (c_{tt} - c_{xx}\alpha^2 - c_{yy}\beta^2 - c_{zz}\gamma^2 - 2c_{xy}\alpha\beta - 2c_{yz}\beta\gamma - 2c_{xz}\alpha\gamma) t'^2.                                     
	\label{eq:A_4D}
\end{align}
Then the equation group is obtained:
\begin{equation}
	\left\{\begin{array}{l}
		c_{xx}\alpha + c_{xy}\beta + c_{xz}\gamma = c_{xt} \\
		c_{xy}\alpha + c_{yy}\beta + c_{yz}\gamma = c_{yt}
        \\
		c_{xz}\alpha + c_{yz}\beta + c_{zz}\gamma = c_{zt} \\
	\end{array}\right..
	\label{eq:alpha_beta_gamma}
\end{equation}
After solving $\alpha$, $\beta$, and $\gamma$, the 3D center position at time $t$ is obtained
\begin{equation}
	\left\{\begin{array}{l}
		\mu_x(t) = \mu_{x} - \alpha (t-\mu_{t}) \\
		\mu_y(t) = \mu_{y} - \beta (t-\mu_{t})  \\
		\mu_z(t) = \mu_{z} - \gamma (t-\mu_{t}) \\
	\end{array}\right.,
\end{equation}
In addition, from Eq.~\ref{eq:A_4D}, the inverse of 3D covariance matrix $\threeDCovFromfourDCov^{-1}$ is:
\begin{equation}
\threeDCovFromfourDCov^{-1} = \left(\begin{matrix}
c_{xx} & c_{xy} & c_{xz} \\
c_{xy} & c_{yy} & c_{yz} \\
c_{xz} & c_{yz} & c_{zz} \\
\end{matrix}\right),
\label{eq:cov_3d_inv}
\end{equation}
Let
\begin{equation}
\lambda = c_{tt} - c_{xx}\alpha^2 - c_{yy}\beta^2 - c_{zz}\gamma^2 - 2c_{xy}\alpha\beta - 2c_{yz}\beta\gamma - 2c_{xz}\alpha\gamma,
\end{equation}
after adding $\centerPos_{x}, \centerPos_{y}, \centerPos_{z}, \centerPos_t$ back to $x', y', z', t'$, we get
\begin{equation}
\label{eq:projected3dgs}
	\threeDGaussianFromFourDG(\threeDPosition, \timeT) = \text{e}^{-\frac{1}{2}\lambdA (\timeT - \mu_t)^2}\text{e}^{-\frac{1}{2}\left[\threeDPosition - \centerPos(\timeT)\right]^T\threeDCovFromfourDCov^{-1}\left[\threeDPosition - \centerPos(\timeT)\right]}.
\end{equation}

\textbf{Avoiding Numerical Instability.}
Directly calculating $\threeDCovFromfourDCov$ from $\threeDCovFromfourDCov^{-1}$ according to Eq.~\ref{eq:cov_3d_inv} can induce numerical instability of matrix inverse. This issue predominantly arises when the scales of the 3D Gaussian exhibit substantial magnitude discrepancies, leading to significant errors in calculating $\threeDCovFromfourDCov$, and resulting in excessively large Gaussians. To circumvent this challenge, direct calculation of $\threeDCovFromfourDCov$ must be avoided.

Given that $\fourDCov$ and its inverse $\fourDCov^{-1}$ are both symmetric matrices, we set:
\begin{equation}
\fourDCov = \left(\begin{matrix}
        \bigU   & \V \\
        \V^T & \W \\
\end{matrix}\right)
\text{~and~}
\fourDCov^{-1} = \left(\begin{matrix}
        \A   & \M \\
        \M^T & \Z      \\
    \end{matrix}\right),
\end{equation}
where $\bigU$ and $\A$ are 3$\times$3 matrices.
By applying the inverse formula for symmetric block matrices, it follows that:
\begin{equation}
\fourDCov = \left(\fourDCov^{-1}\right)^{-1} = \left(\begin{matrix}
\A^{-1} + \dfrac{\A^{-1}\eta\eta^T\A^{-1}}{\h} & -\dfrac{\A^{-1}\eta}{\h} \\
-\dfrac{\eta^T\A^{-1}}{\h} & \dfrac{1}{\h} \\
\end{matrix}\right)
= \left(\begin{matrix}
\bigU & \V \\
\V^T & \W \\
\end{matrix}\right),
\end{equation}
where $\h = \Z - \eta^T\A^{-1}\eta$. 

Comparing Eq.~\ref{eq:cov_4d_inv} with Eq.~\ref{eq:cov_3d_inv}, we have $\threeDCovFromfourDCov = \A^{-1}$, and comparing the two expressions of $\fourDCov$, we get
\begin{equation}
\label{eq:sigma'}
\threeDCovFromfourDCov = \A^{-1} = \bigU - \dfrac{\V\V^T}{\W},
\end{equation}
thus effectively avoiding direct inversion of $A$ and leveraging the computationally feasible sub-blocks of $\fourDCov$ to compute $\threeDCovFromfourDCov$.

Additionally, we can use the above expression to simplify the calculation of $(\alpha, \beta, \gamma, \lambda)$. Let $\g=(\alpha, \beta, \gamma)^T$, then Eq. \ref{eq:alpha_beta_gamma} is reformulated as:
\begin{equation}
\A \g=\eta,
\end{equation}
and its solution is
\begin{equation}
\label{eq:g}
\g=\A^{-1}\eta = -\h \V = -\dfrac{\V}{\W}.
\end{equation}
Similarly, for $\lambda$:
\begin{equation}
\label{eq:lambda}
\begin{array}{cl}
\lambda & = \Z - \g^T \A \g \\
& = \Z - \eta^T(\A^{-1})^T\eta \\
& = \Z - \eta^T \A^{-1}\eta \\
& = \h + \eta^T \A^{-1}\eta - \eta^T \A^{-1}\eta \\
& = {\W}^{-1} \\
\end{array}.
\end{equation}
To summarize, the 3D Gaussian project from the 4D at time $\timeT$ is obtained as:  
\begin{equation}
\label{eq:projected3dgs}
	\threeDGaussianFromFourDG(\threeDPosition, \timeT) = \text{e}^{-\frac{1}{2}\lambdA (\timeT - \centerPos_t)^2}\text{e}^{-\frac{1}{2}\left[\threeDPosition - \centerPos(\timeT)\right]^T\threeDCovFromfourDCov^{-1}\left[\threeDPosition - \centerPos(\timeT)\right]},
\end{equation}
where
\begin{equation}
\begin{aligned}
\lambdA &= \W^{-1}, \\
\threeDCovFromfourDCov &= \A^{-1} = \bigU - \dfrac{\V \V^T}{\W}, \\ 
\centerPos(\timeT) &= (\centerPos_x, \centerPos_y, \centerPos_z)^T + (\timeT - \centerPos_t)\dfrac{\V}{\W} .
	\label{eq:center-motion}
\end{aligned}
\end{equation}

\section{Additional Experiments}

\subsection{Datasets Details}~\label{sec_supp:dataset}
\textbf{Plenoptic Video Dataset~\cite{li2022neural}.} This is a real-world dataset captured by a multi-view GoPro camera system. We evaluate the baselines on 6 scenes: \textit{Coffee Martini}, \textit{Flame Salmon}, \textit{Cook Spinach}, \textit{Cut Roasted Beef}, \textit{Flame Steak}, and \textit{Sear Steak}. Each scene includes from 17 to 20 views for training and one central view for evaluation. The size of the images is downsampled to 1352$\times$1014 for fair comparison. This dataset presents a variety of challenging elements, including the sudden appearance of flames, moving shadows, as well as translucent and reflective materials.

\textbf{D-NeRF Dataset~\cite{pumarola2021d}.} This is a synthetic dataset of monocular video that presents a significant challenge due to the single camera viewpoint available at each timestamp. This dataset contains 8 scenes: \textit{Hell Warrior}, \textit{Mutant}, \textit{Hook}, \textit{Bouncing Balls}, \textit{Lego}, \textit{T-Rex}, \textit{Stand Up}, and \textit{Jumping Jacks}. Each scene comprises 50 to 200 images for training, 10 or 20 images for validation, and 20 images for testing. Each image within this dataset is downsampled to a standard resolution of 400$\times$400 for training and evaluation following the previous work~\cite{pumarola2021d}.

\new{\textbf{HyperNeRF Dataset~\cite{park2021hypernerf}.} This real-world dataset is captured using two cameras for each scene. Each scene has an equal number of photos from the left and right cameras, with the number of photos from a single camera ranging from 163 to 512. We strictly follow the state-of-the-art method on this dataset Deformable4DGS~\cite{wu20234d} for all the experiment settings as follows. We evaluate on 4 scenes from the dataset: \textit{3D Printer}, \textit{Chicken}, \textit{Broom}, and \textit{Banana}. We take every other view from both left and right cameras as the test views, with the remaining views used as training views. The image resolution is 960$\times$540.

\textbf{Total-Recon Dataset~\cite{song2023totalrecon}.} This dataset comprises 11 RGBD sequences featuring 3 distinct cats, 1 dog, and 2 human subjects across 4 different indoor environments. These three scenes are subsampled at 10 FPS and range from 550 to 641 frames, resulting in more dynamic motions than other datasets used in this paper. The difficulties in reconstructing this dataset include very sparse perspectives and scenes with a wide range of variations. Following~\cite{song2023totalrecon}, we train on images from the left cameras and evaluate the novel view synthesis results on the held-out right videos. We use three out of the eleven scenes in this dataset: \textit{Cat1}, \textit{Dog1}, and \textit{Human1}.}
   \begin{table}[t]
\centering
\caption{\new{\textbf{Evaluation on HyperNeRF Dataset}. We conduct a comparision with the state-of-the-art method on HyperNeRF dataset~\cite{park2021hypernerf}. Our method surpasses the baseline in terms of PSNR for the majority of scenes and on average.}
}
\setlength{\tabcolsep}{1.5pt}
\resizebox{\linewidth}{!}{
\begin{tabular}{c|cccc|c}
\specialrule{.15em}{.1em}{.1em}
Method                               & 3D Printer & Broom& Chicken& Banana &Avg \\ \hline
Deformable4DGS~\cite{wu20234d}           &21.96  &21.94  &\textbf{28.55}  &27.44  &24.97   \\ 
Ours                     &\textbf{24.83 } &\textbf{22.43}  & 27.50 & \textbf{28.07} & \textbf{25.71} \\ 
\specialrule{.1em}{.05em}{.05em}
\end{tabular}
} 
\label{table:HyperNeRF}
\end{table}

\subsection{Additional Implementation Details}
In the spatial initialization for Plenoptic Video Dataset~\cite{li2022neural}, we define an axis-aligned bounding box sized according to the range of SfM points. For D-NeRF dataset, the box dimensions are set to $[-1.3, 1.3]^3$. Within these boxes, we randomly sample 100,000 points as the positions of the Gaussians. The time means of Gaussians are uniformly sampled from the entire time duration, \ie, $[0, 1]$ for D-NeRF dataset and $[0, 10]$ for Plenoptic Video Dataset. The initialized time scale is set to 0.1414 for D-NeRF dataset and 1.414 for Plenoptic dataset.

Following~\cite{kerbl20233d}, we employ the Adam optimizer with specific learning rates: $1.6e-4$ for positions, $1.6e-4$ for times, $5e-3$ for 3D scales and time scales, $1e-3$ for rotation, $2.5e-3$ for SH low degree and $1.25e-4$ for SH high degrees, and 0.05 for opacity.
We apply an exponential decay schedule to the learning rates for positions and times, starting from the initial rate and decaying to $1.6e-6$ for positions and times at step 30,000. 
The total optimization consists of 30,000 steps on Plenoptic dataset and 20,000 steps on D-NeRF dataset. Opacity is reset every 3,000 steps, while densification is executed at intervals of 100 steps, starting from 500 to 15,000 steps.
\subsection{\new{Additional Real-World Dataset Results}}
\new{We provide results on two additional real-world datasets. Dataset details are explained in Sec.~\ref{sec_supp:dataset}}.

\begin{figure*}
	\centering
\setlength{\tabcolsep}{2pt}
 \resizebox{\textwidth}{!}{
    \begin{tabular}{ccccccccc}
     {\includegraphics[width=0.18\textwidth]{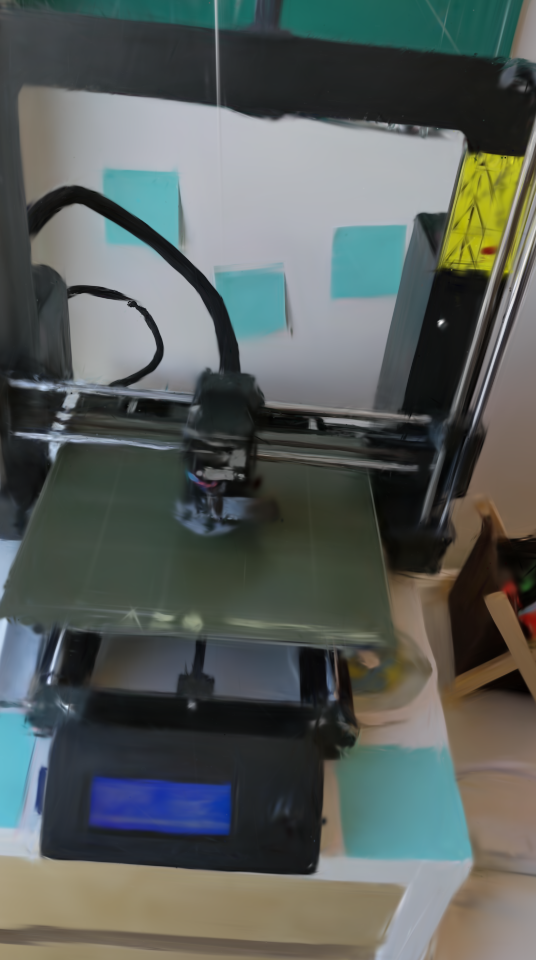}}&
     {\includegraphics[width=0.18\textwidth]{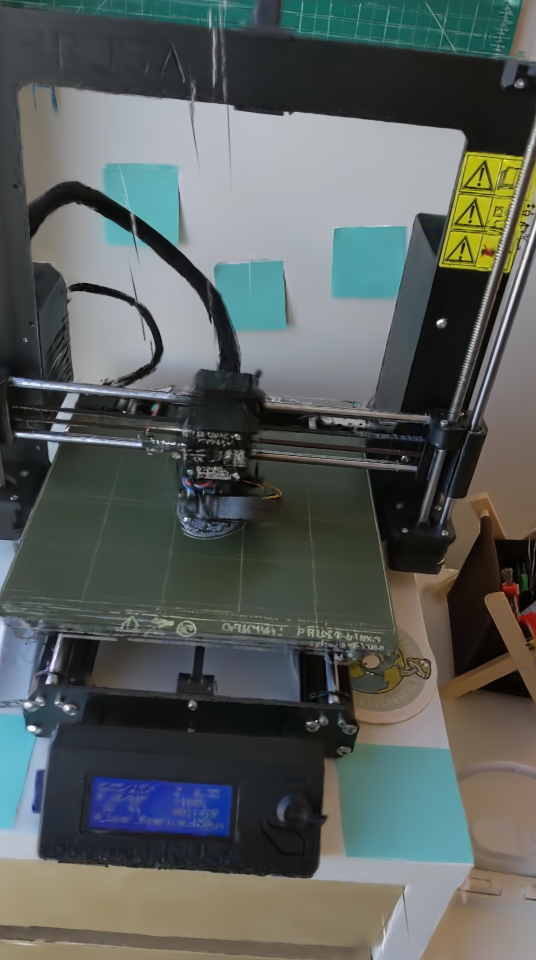}}&
     {\includegraphics[width=0.18\textwidth]{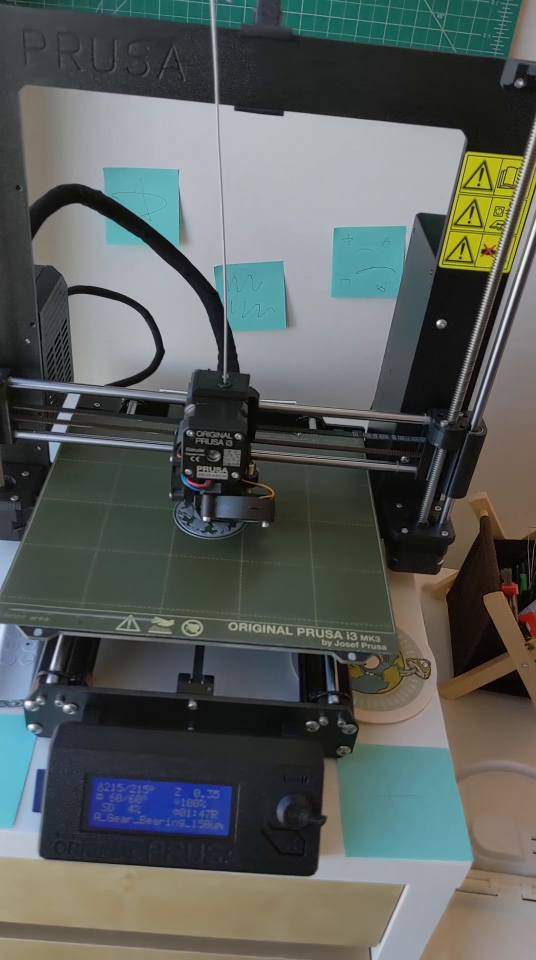}}& 
    \includegraphics[width=0.18\textwidth]{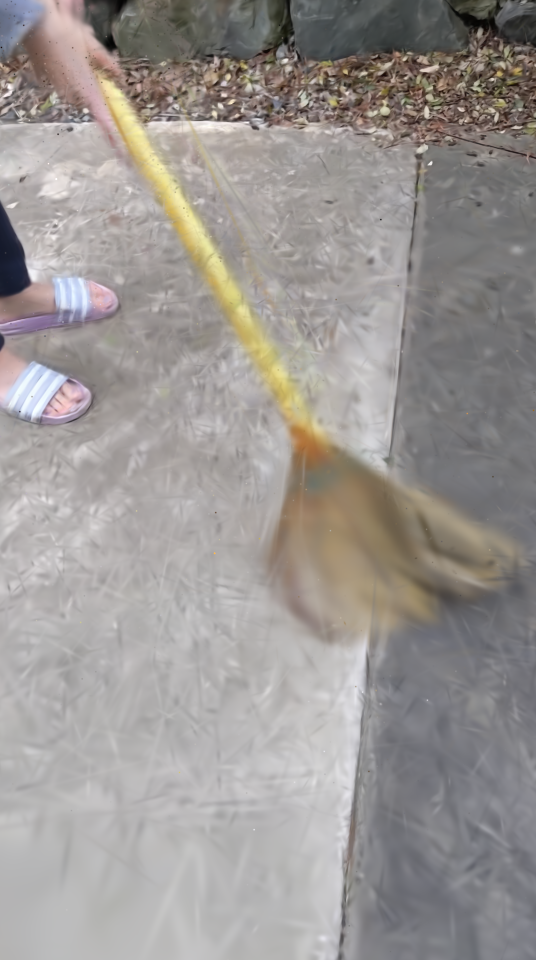}&
	\includegraphics[width=0.18\textwidth]{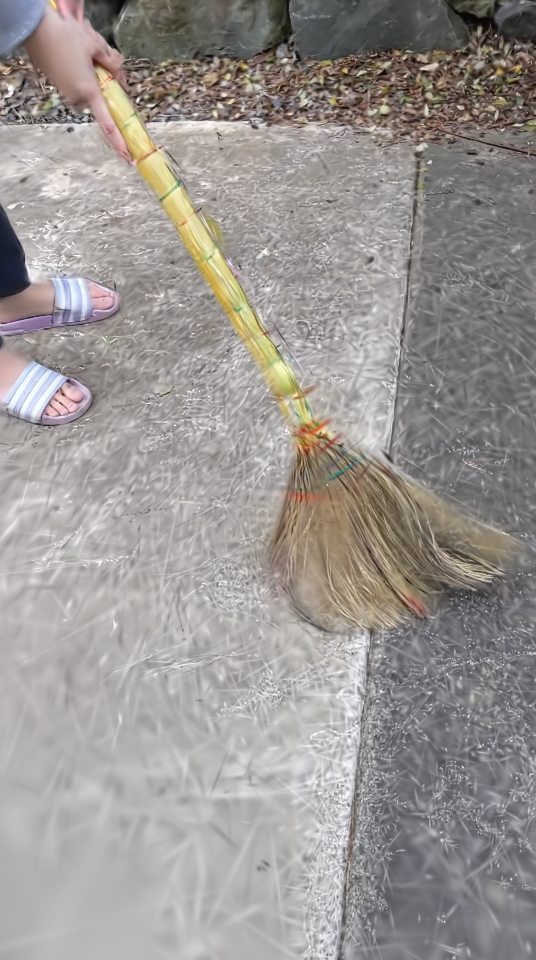}&
    \includegraphics[width=0.18\textwidth]{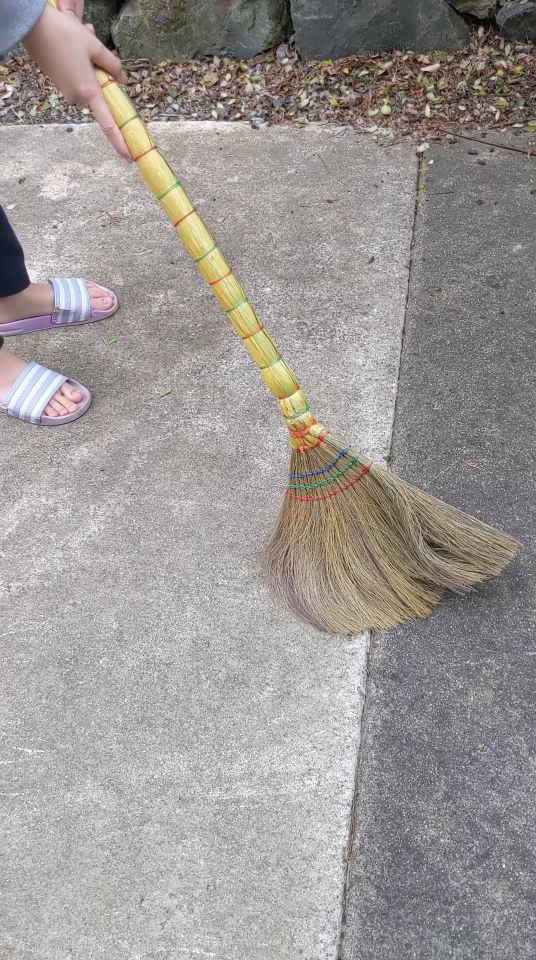}& 
    \includegraphics[width=0.18\textwidth]{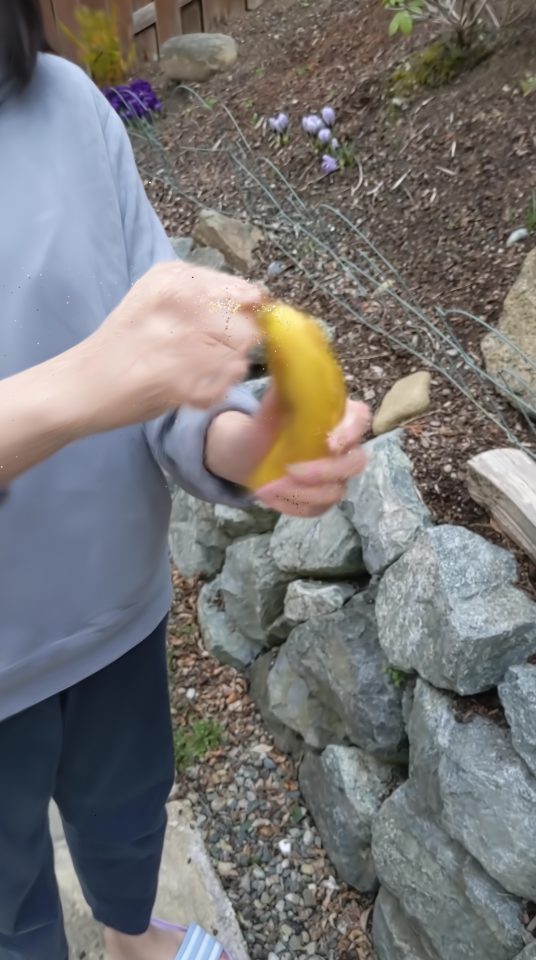}&
	\includegraphics[width=0.18\textwidth]{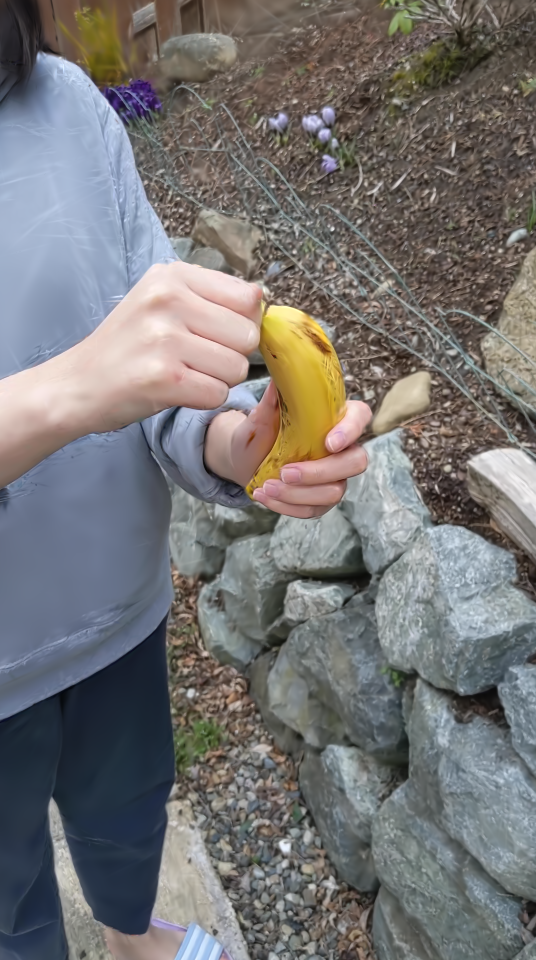}&
    \includegraphics[width=0.18\textwidth]{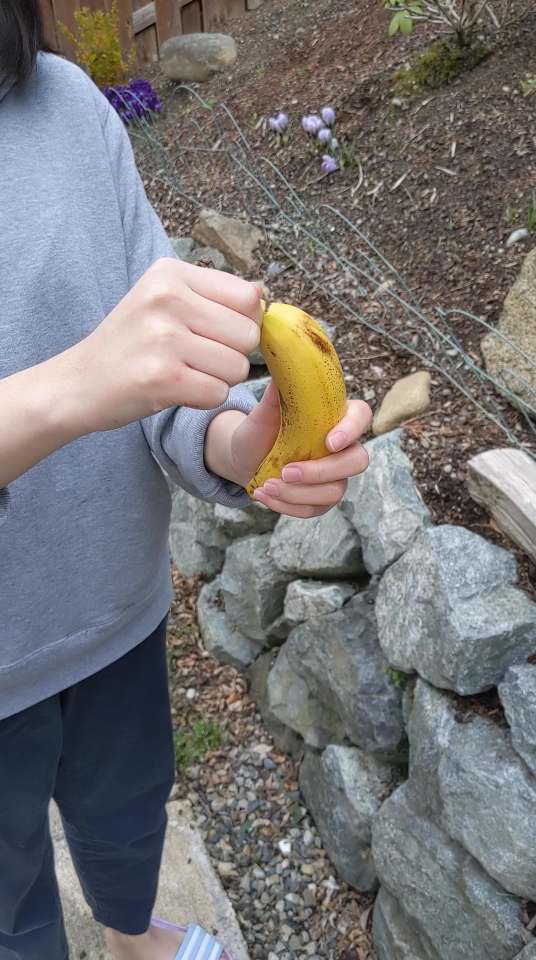}\\ 
    Deformable4DGS & Ours & Ground Truth&Deformable4DGS & Ours & Ground Truth&Deformable4DGS & Ours & Ground Truth
    \end{tabular}
    }
	\caption{\label{fig:HyperNeRF} \new{\textbf{Qualitative Comparison of Ours Methods and Deformable4DGS on HyperNeRF Dataset.} We compare our method with Deformable4DGS~\cite{wu20234d}, the state-of-the-art method on HyperNeRF dataset~\cite{park2021hypernerf}. This figure shows that our method reconstructs more details and reduces blurring. For example, the center part of the 3D printer, the entire broom, as well as the banana peel and hand.}}
\end{figure*}

\textbf{\new{HyperNeRF Dataset~\cite{park2021hypernerf}}} We run the official released code from the authors~\cite{wu20234d} to obtain the baseline results. As shown in Tab.~\ref{table:HyperNeRF}, our method exhibits superior rendering quality compared to the baseline method across most scenes and on average.
As presented in Fig.~\ref{fig:HyperNeRF}, our method promotes a more detailed reconstruction of static regions and reduces blurring over the baseline. For example, in the static regions, our approach better reconstructs the yellow sticker in \textit{3D Printer} and the intricate texture of the banana peel in \textit{Banana}. In the dynamic regions, our method notably enhances the clarity of textures on objects like the broom in \textit{Broom}, while effectively reducing hand motion blur in \textit{Banana}. These outcomes emphasize the precision of our method in capturing both static and dynamic visual details.

\begin{figure*}
	\centering
\setlength{\tabcolsep}{2pt}
 \resizebox{\textwidth}{!}{
    \begin{tabular}{cccc}
     {\includegraphics[width=0.25\textwidth]{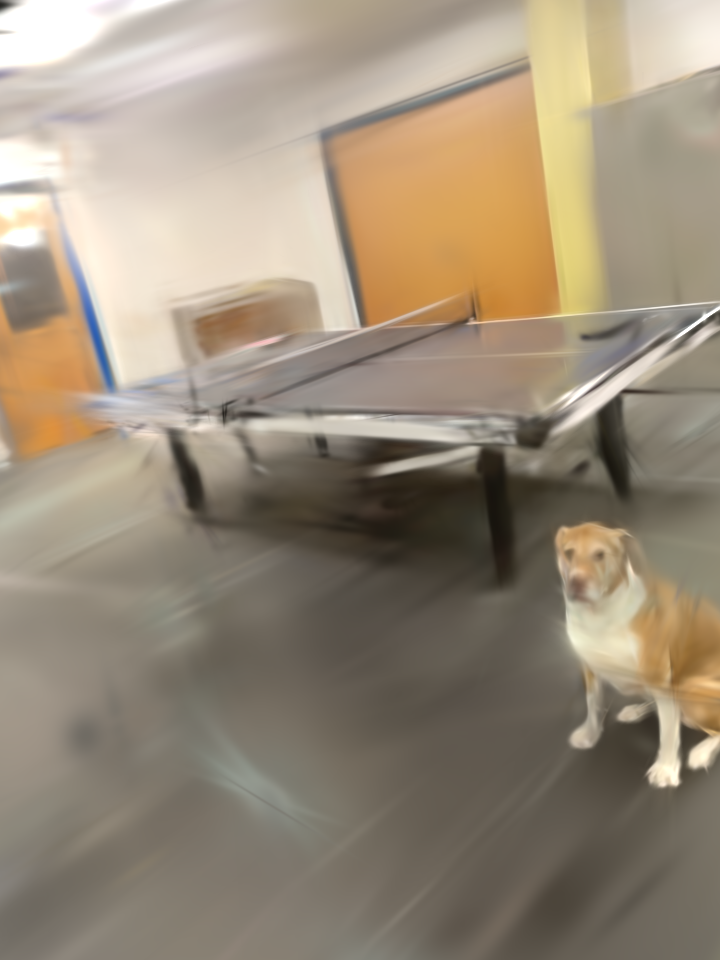}}&
     {\includegraphics[width=0.25\textwidth]{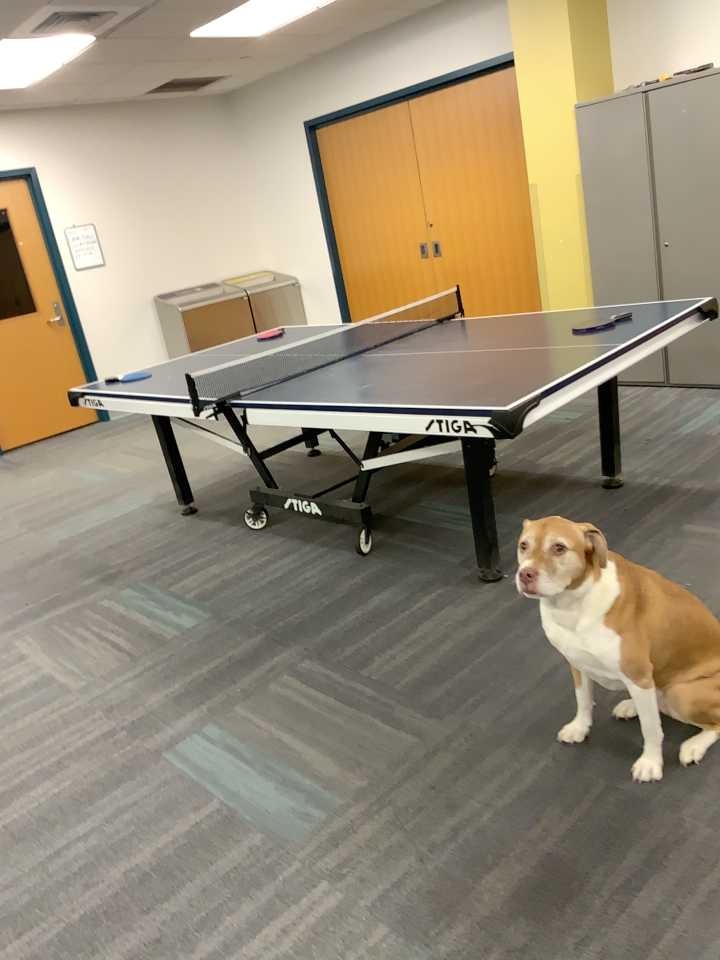}}&
     {\includegraphics[width=0.25\textwidth]{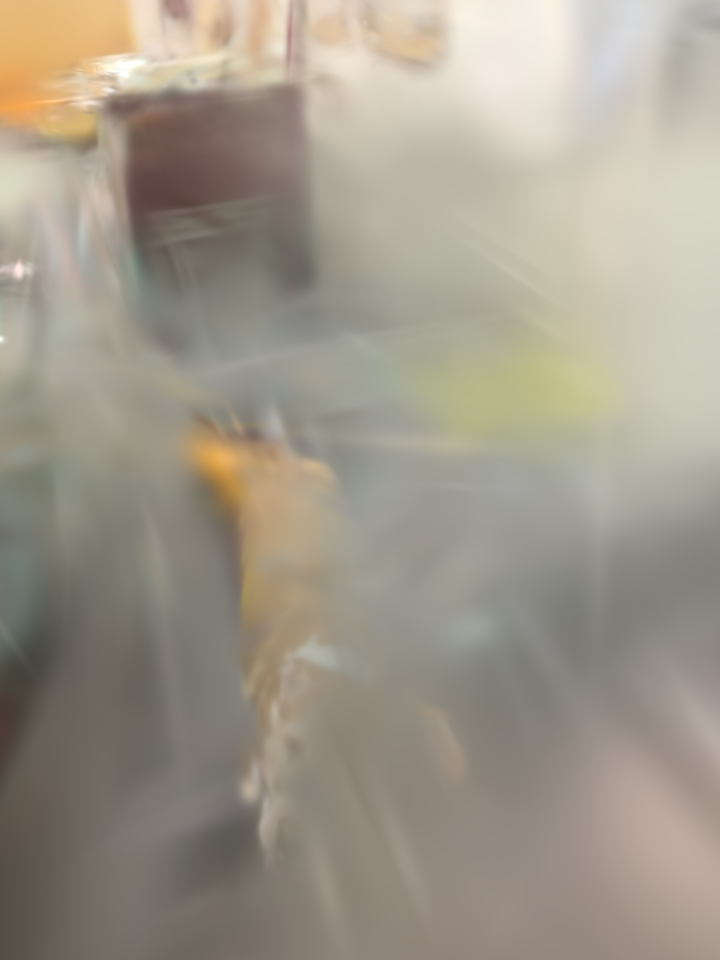}}& 
    \includegraphics[width=0.25\textwidth]{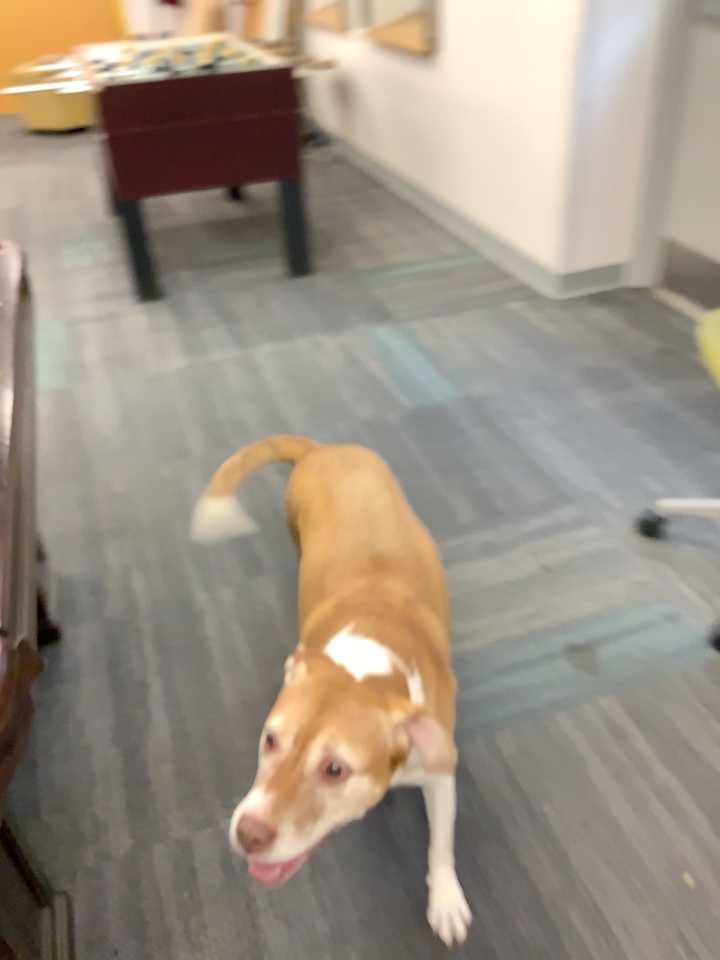}\\ 
    Ours & Ground Truth&Ours & Ground Truth
    \end{tabular}
    }
	\caption{\label{fig:totalrecon} \new{\textbf{Qualitative Results on Total-Recon Dataset.} We show the qualitative results of our method on the challenging Total-Recon dataset~\cite{park2021hypernerf}. In this scene, the dog runs or walks freely in a spacious room. Due to the very sparse training views, our NVS results become much worse than those on other datasets. For scene regions that have been captured by multiple frames (\eg, left example), the synthesized novel views are relatively clear. When the dog runs and the camera follows with rapidly changing camera poses, the rendering quality becomes severely affected. For example, the right example shows the renderings are very bad due to floaters caused by too sparse training views.}}
\end{figure*}

\begin{table}[t]
\centering
\caption{\new{\textbf{Evaluation on Total-Recon Dataset}. We conduct a comparision with the HyperNeRF~\cite{park2021hypernerf}, D$^2$NeRF~\cite{wu2022d} and Total-Recon~\cite{song2023totalrecon} on Total-Recon dataset~\cite{song2023totalrecon}. Our method achieves the highest PSNR among methods that only use supervision from RGB images (depth loss not used). Our method also achieves results comparable with the baselines with added depth supervision.}
}
\setlength{\tabcolsep}{1.5pt}
\resizebox{\linewidth}{!}{
\begin{tabular}{c|c|ccc|c}
\specialrule{.15em}{.1em}{.1em}
Depth &Method                               & Dog1 & Cat1& Human1 &Avg \\ \hline
&HyperNeRF~\cite{park2021hypernerf}           &12.84  &14.27  &11.94  &13.02   \\ 
w/o &D$^{2}$NeRF~\cite{wu2022d}      &13.37  &11.74  &11.88  &12.33   \\ 
&Ours          &\textbf{16.65}  &\textbf{14.91}  &\textbf{13.96}  &\textbf{15.17 }  \\ 
\hline
&HyperNeRF (+depth)~\cite{park2021hypernerf}           &16.86  &\textbf{16.95}  &13.25  &15.69   \\ 
w/ &D$^{2}$NeRF (+depth) ~\cite{wu2022d}         &13.44  &11.88  &12.14  &12.49   \\ 
&Total-Recon~\cite{song2023totalrecon}         &\textbf{17.60}  &15.77 &\textbf{18.39}  &\textbf{17.25}   \\ 
\specialrule{.1em}{.05em}{.05em}
\end{tabular}
}

\label{table:totalrecon}
\end{table}

\new{\textbf{Total-Recon Dataset~\cite{song2023totalrecon}}
We compare with baselines on the very challenging Total-Recon dataset. We use the first example in each category and report both qualitative and quantitative results. The PSNR numbers in Tab.~\ref{table:totalrecon} for all compared baselines are copied from the Tab.~5 in the Total-Recon paper~\cite{song2023totalrecon}. Due to extremely sparse views and fast large motions, methods that only use RGB supervision have very low rendering quality. Nevertheless, our method outperforms other RGB-only baselines by a large margin. In fact, our method is on par with the variants of the RGB-only baselines that are trained with additional depth supervision. The best-performing method Total-Recon~\cite{song2023totalrecon} uses mask, depth, and flow supervision in addition to RGB loss. This method also involves multi-stage training that requires first pretraining separately on foreground objects and backgrounds and a second finetuning stage. We believe by leveraging these various techniques and rich supervision, the proposed method can be largely improved.}

\subsection{Entropy Loss}

\new{Entropy loss has also been proposed in a concurrent work~\cite{guedon2023sugar}. While~\cite{guedon2023sugar} uses entropy loss for surface recovery, we leverage it to diminish floaters in challenging sparse-view reconstruction, evidenced in main paper Fig.~8. 

3DGS~\cite{kerbl20233d} utilizes Sigmoid function but the opacity is optimized from RGB only. We found that entropy loss further pushes Gaussian opacities towards 1, which densifies Gaussians around the surface and removes floaters. A stronger push towards extreme values will better facilitate surface-based rendering~\cite{guedon2023sugar} but will not improve PSNR. To investigate the effect of entropy loss on the vanilla 3DGS, we ran our method with the entropy loss on the garden scene as used in the main paper Fig.~4. We obtained a performance similar to that of running without entropy loss. We find that entropy loss is helpful in sparse-view reconstruction (\eg, in the D-NeRF dataset) as it condenses the surface. However, it may not improve rendering quality for dense-view reconstruction.}

\begin{table}[t]
\centering
\caption{\textbf{Ablation of {\methodshort} on D-NeRF Dataset}. We validate three designs on rendering quality by removing one component at each time from the full model: (\textit{b}) Entropy (with cross-entropy loss), (\textit{c}) KNN (with 4D KNN Consistency loss), and (\textit{Full}) Batch (with batch training).}
\setlength{\tabcolsep}{3pt}
\resizebox{0.65\linewidth}{!}{
\begin{tabular}{cccc|c}
\specialrule{.15em}{.1em}{.1em}
ID&\multicolumn{3}{c|}{Ablation Items}&\multicolumn{1}{c}{D-NeRF}\\
& Entropy    &KNN       &Batch        & PSNR$\uparrow$   \\ \hline
\textit{a}&          &\checkmark&\checkmark&32.11\\ 
\textit{b}  &\checkmark &         &\checkmark            &32.64  \\ 
\textit{c} &\checkmark &\checkmark&            &31.91  \\  \hline
\textit{Full}  &\checkmark &\checkmark&\checkmark&33.06 \\ \specialrule{.1em}{.05em}{.05em}
\end{tabular}
}
\label{table:ablation2}
\end{table}

\begin{figure*}
	\centering
    \includegraphics[width=0.7\textwidth,trim={0 0.1cm 0 0},clip]{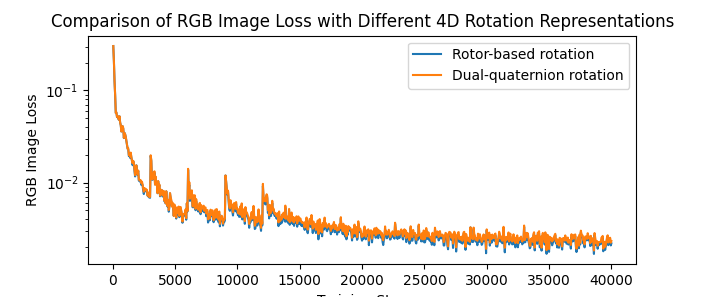}
	\caption{\label{fig:rotation} \new{\textbf{Comparison of RGB Image Loss with Different 4D Rotation Representations.} We compare the training loss curves using the proposed rotor-based 4D rotation and dual-quaternion 4D rotation. We run three times for each rotation and there is no significant difference.}}
\end{figure*}

\begin{center}
\begin{table*}[t]
\centering
\setlength{\tabcolsep}{2pt}
\caption{\textbf{Influence of Background Colors on D-NeRF Dataset.} We report the per-scene and average PSNR for each method. All methods use the default background colors as set in their official released code. It is observed that most scenes yield a higher PSNR on black backgrounds, especially when the foregrounds are darker. This phenomenon has also been observed by Deformable3DGS~\cite{yang2023deformable}. Our method outperforms baselines on both background colors.}
\label{table:D-NeRF-background}
\resizebox{0.99\textwidth}{!}{

\begin{tabular}{cc|cccccccc|c}
\specialrule{.15em}{.1em}{.1em}
Color&Method   &T-Rex&Jumping Jacks&Hell Warrior&Stand Up&Bouncing Balls&Mutant&Hook&Lego&Avg\\ \hline
&D-NeRF~\cite{pumarola2021d}        &31.45&32.56&24.70&33.63&38.87&21.41&28.95&21.76  &29.17  \\ 
&TiNeuVox~\cite{fang2022fast}       &32.78&34.81&28.20&35.92&40.56&33.73&31.85&25.13  &32.87 \\  
&K-Planes~\cite{fridovich2023k}   &31.44&32.53&25.38&34.26&39.71&33.88&28.61&22.73    &31.07    \\ 
White&Deformable4DGS~\cite{wu20234d}        &33.12&34.65&25.31&36.80&39.29&37.63&31.79&25.31   &32.99 \\
&Deformable3DGS~\cite{yang2023deformable}&40.14&38.32&32.51&42.65&43.97&42.20&36.40&25.55&37.72     \\ 
&RealTime4DGS~\cite{yang2023real} &31.22&31.29&24.44&37.89&35.75&37.69&30.93&24.85  &31.76 \\
&Ours &31.24&33.37&36.85&38.89&36.30&39.26  &33.33  &25.24&33.06   \\
\hline

&Deformable3DGS~\cite{yang2023deformable} &38.55&39.21&42.06&45.74&41.33&44.16&38.04&25.38 &39.31    \\ 
Black&RealTime4DGS~\cite{yang2023real} &29.82&30.44&34.67&39.11&32.85&38.74&31.77&24.29    &32.71   \\  
&Ours   &31.77&33.40&33.52  &40.79&34.74  &40.66&34.24&24.93  &34.26   \\ 
\specialrule{.1em}{.05em}{.05em}
\end{tabular}

}
\end{table*} 
\end{center}

\subsection{\new{Additional Ablation Studies}}
In Tab.~\ref{table:ablation2}, we present an additional ablation table where we remove one component from the final model at each time.

\textbf{Rotor-based rotation \vs dual-quaternion rotation.} We further compare the proposed rotor-based rotation and the dual quaternion rotation~\cite{yang2023real} about their effects on rendering performance. We modify our code for Gaussian initialization and optimization by replacing the rotor-based rotation with the dual-quaternion rotation following the officially released code~\cite{yang2023real}. All the other settings are kept the same as used in the main paper Tab.~2. We found no differences between the testing PSNRs from both rotation representations. In Fig.~\ref{fig:rotation}, we plot the training curves of the two experiments using different rotation representations on scene \textit{Lego}. This experimentally verifies that both representations are equally effective when representing 4D dynamic scenes. However, our proposed rotor-based representation unifies and supports both static and dynamic scene modeling as showcased in the main paper Fig.~4. Moreover, the rotor-based rotation representation is more interpretable and enables new spatial-temporal analysis. For example, the magnitude of the rotor’s temporal components indicates how still a Gaussian is. This may benefit downstream tasks such as tracking or static/dynamic content separation. 
\subsection{Influence of Background Colors on D-NeRF Dataset.}
D-NeRF dataset provides synthetic images without backgrounds. Consequently, previous baseline methods incorporate either a black or white background during training and evaluation. Specifically, Deformable3DGS~\cite{yang2023deformable} and RealTime4DGS~\cite{yang2023real} utilize a black background, while other methods opt for a white background. 

Our observations, as shown in Tab.~\ref{table:D-NeRF-background}, indicate that our method yields higher rendering quality (PSNR 34.26) when trained with a black background as opposed to a white one (PSNR 33.06). In particular, for scenes with brighter foregrounds (\textit{Jumping Jacks}, \textit{Bouncing Balls}, and \textit{Lego}), models trained using white backgrounds perform higher. These performance discrepancies between the two background colors have also been observed and reported in prior work~\cite{yang2023deformable}.

The results reported in Tab.~2 
for our method are based on experiments using black background. For all the other baselines, we follow their original settings. Note that our results on the white background outperform all previous methods that use white background. And similarly, our results on the black background outperform all previous methods that use black background in their original experiments.  

The results reported in ablation Tab.~3 
are conducted with white background.  For the purpose of visualization, we show images trained with white background in Fig.~6 
and Fig.~7 
which include our method and two reproduced baselines (Deformable3DGS~\cite{yang2023deformable} and RealTime4DGS~\cite{yang2023deformable}). Finally, for all experiments on Plenoptic Video Dataset~\cite{li2022neural}, we simply choose black background for our method and follow the original settings of all the baselines.

\end{document}